\documentclass[10pt,twocolumn,letterpaper]{article}

\usepackage[pagenumbers]{cvpr} % To force page numbers, e.g. for an arXiv version

\usepackage[dvipsnames]{xcolor}

\usepackage{algorithm,algpseudocode}
\usepackage{command}
\definecolor{cvprblue}{rgb}{0.21,0.49,0.74}
\usepackage[pagebackref,breaklinks,colorlinks,citecolor=cvprblue]{hyperref}

\def\paperID{3637} % *** Enter the Paper ID here
\def\confName{CVPR}
\def\confYear{2024}

\title{Generating Visually Realistic Adversarial Patch}

\author{Xiaosen Wang\\
Huawei Singular Security Lab\\
{\tt\small xiaosen@hust.edu.cn}
\and
Kunyu Wang\\
Chinese University of Hong Kong\\
{\tt\small kunyuwang@link.cuhk.edu.hk}
}

\begin{document}
\maketitle
\begin{abstract}
    Deep neural networks (DNNs) are vulnerable to various types of adversarial examples, bringing huge threats to security-critical applications. Among these, adversarial patches have drawn increasing attention due to their good applicability to fool DNNs in the physical world. However, existing works often generate patches with meaningless noise or patterns, making it conspicuous to humans. To address this issue, we explore how to generate visually realistic adversarial patches to fool DNNs. Firstly, we analyze that a high-quality adversarial patch should be \textbf{realistic}, \textbf{position irrelevant}, and \textbf{printable} to be deployed in the physical world. Based on this analysis, we propose an effective attack called \name, to generate visually realistic adversarial patches. Specifically, \name constrains the patch in the neighborhood of a real image to ensure the visual reality, optimizes the patch at the poorest position for position irrelevance, and adopts Total Variance loss as well as gamma transformation to make the generated patch printable without losing information. Empirical evaluations on the ImageNet dataset demonstrate that the proposed \name exhibits outstanding attack performance in the digital world. Moreover, the generated adversarial patches can be disguised as the scrawl or logo in the physical world to fool the deep models without being detected, bringing significant threats to DNNs-enabled applications.
\end{abstract}
\section{Introduction}
\label{sec:intro}
% Deep neural networks (DNNs) have achieved unprecedented success in many tasks, \eg, image classification~\cite{alex2012ImageNet,he2016resnet,huang2017densely}, object detection~\cite{ren2015faster,joseph2016you}, segmentation~\cite{olaf2015unet,jonathan2015fully}, face recognition~\cite{florian2015facenet,wang2018cosface}, \etc. However, recent works found that DNNs are vulnerable to numerous types of adversarial attacks, such as adversarial examples~\cite{szegedy2014intriguing,goodfellow2015FGSM,dong2018boosting,wang2021enhancing}, adversarial patches~\cite{tom2017adversarial,Karmon2018lavan,liu2019perceptual}, backdoor attacks~\cite{chen2017targeted,li2022backdoor}, which have attracted great attention for security-critical applications. Compared with adversarial examples and backdoor attacks, adversarial patches can be deployed to attack real-world applications, such as face recognition~\cite{sharif2016accessorize,xiao2021improving,wei2022adversarial}, autonomous driving~\cite{eykholt2018robust,nesti2022evaluating}, making it become one of the most popular adversarial attacks. 
With the prosperous development of Deep Neural Networks (DNNs), DNNs have achieved excellent performance in many tasks, \eg, image classification~\cite{alex2012ImageNet,he2016resnet}, object detection~\cite{ren2015faster,joseph2016you}, segmentation~\cite{olaf2015unet,jonathan2015fully}, \etc. However, \citet{szegedy2014intriguing} found that DNNs are vulnerable to adversarial examples, \ie, the maliciously crafted inputs that are indistinguishable from the correctly classified images but can induce misclassification on the target model. Such vulnerability poses significant threats when applying DNNs to security-critical applications, which also attracts broad attention to the security of DNNs~\cite{goodfellow2015FGSM,dong2018boosting,wang2021enhancing,wang2023structure,yang2022robust,yu2022texthacker}.

As the research of adversarial examples progresses, various types of adversarial attacks have been proposed, such as adversarial patches~\cite{tom2017adversarial,Karmon2018lavan,liu2019perceptual}, backdoor attacks~\cite{chen2017targeted,li2022backdoor}, weight-level attacks~\cite{breier2018practical,rakin2020tbt}. Compared with adversarial examples, backdoor attacks, and weight-level attacks, adversarial patches can be printed to attack real-world applications, such as face recognition~\cite{sharif2016accessorize,xiao2021improving,wei2022adversarial}, autonomous driving~\cite{eykholt2018robust,nesti2022evaluating}, making it become one of the most popular adversarial attacks~\cite{sharma2022adversarial}. Hence, it is essential to conduct an in-depth analysis of the intrinsic properties of adversarial patches.

% numerous types of adversarial attacks, such as adversarial examples~\cite{szegedy2014intriguing,goodfellow2015FGSM,dong2018boosting,wang2021enhancing}, adversarial patches~\cite{tom2017adversarial,Karmon2018lavan,liu2019perceptual}, backdoor attacks~\cite{chen2017targeted,li2022backdoor}, which have attracted great attention for security-critical applications. Compared with adversarial examples and backdoor attacks, adversarial patches can be deployed to attack real-world applications, such as face recognition~\cite{sharif2016accessorize,xiao2021improving,wei2022adversarial}, autonomous driving~\cite{eykholt2018robust,nesti2022evaluating}, making it become one of the most popular adversarial attacks. 

\begin{figure}
    \centering
    \begin{minipage}{0.15\textwidth}
        \includegraphics[width=\linewidth]{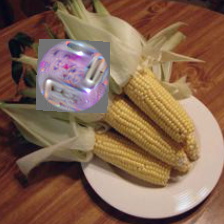}\\
        \vspace{-0.3cm}
        \includegraphics[width=\linewidth]{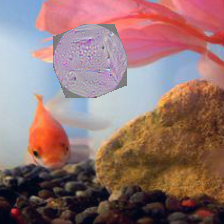}\\
        \vspace{-0.3cm}
        \includegraphics[width=\linewidth]{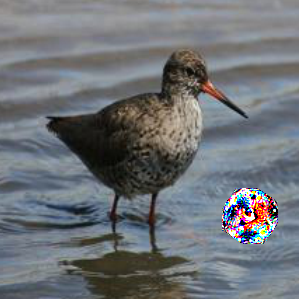}
        %\centering
        \caption*{GoogleAp}
    \end{minipage}%
    \hspace{0.1em}
    \begin{minipage}{0.15\textwidth}
        \includegraphics[width=\linewidth]{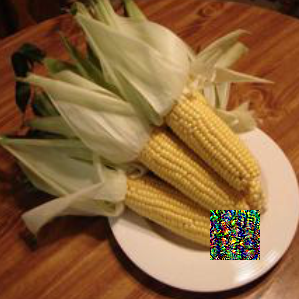}\\
        \vspace{-0.3cm}
        \includegraphics[width=\linewidth]{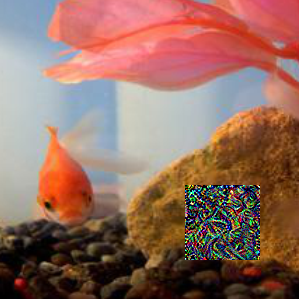}\\
        \vspace{-0.3cm}
        \includegraphics[width=\linewidth]{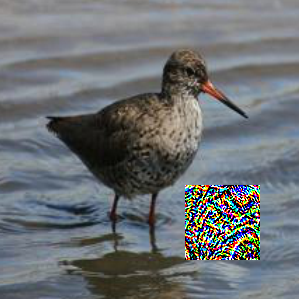}
       % \centering
        \caption*{LaVAN}
    \end{minipage}%
    \hspace{0.1em}
    \begin{minipage}{0.15\textwidth}
        \includegraphics[width=\linewidth]{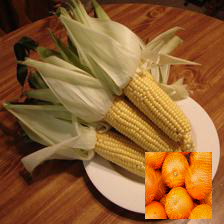}\\
        \vspace{-0.3cm}
        \includegraphics[width=\linewidth]{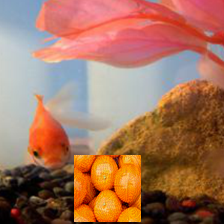}\\
        \vspace{-0.3cm}
        \includegraphics[width=\linewidth]{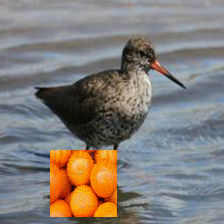}
        \caption*{\name}
    \end{minipage}%
    \caption{The adversarial patches \wrt the same input image generated by GoogleAp, LAVAN and our purposed \name, in which all the images are from ImageNet dataset.}
    \label{fig:realpatch}
\end{figure}

Numerous works~\cite{sharif2016accessorize,tom2017adversarial,eykholt2018robust,liu2019perceptual} have been proposed to enhance the attack ability of adversarial patches. As shown in Fig.~\ref{fig:realpatch}, however, we find that most adversarial patches are not as realistic as the real scrawls~\cite{tom2017adversarial,Karmon2018lavan}, making it easy to be detected when deployed in the physical world. Given the common existence of scrawls and logos, we can disguise the adversarial patches as scrawls to make them inconspicuous in the physical world. Thus, we argue that adversarial patches should be visually realistic instead of strange patterns to humans. To this end, we explore how to effectively and efficiently generate visually realistic adversarial patches.

In this work, we first analyze that a high-quality adversarial patch in the physical world should be 1) \textit{realistic}: it would be similar to the real scrawls or logos; 2) \textit{position irrelevant}: it would be easy to paste the patch to the real scenes; and 3) \textit{printable}: it can be pasted to common scenes. Based on this analysis, we propose an effective visually realistic adversarial patch generating algorithm, called \name. In particular, \name constrains the adversarial patches in the $\epsilon$-neighborhood of a real image to guarantee the visual reality; iteratively maximizes the loss at the position with the minimum loss in a local neighborhood of the current patch to ensure position irrelevant; smooths the patches using Total Variation loss and gamma transformation to make it printable.
Our main contributions are summarized as follows:
\begin{itemize}[leftmargin=*,noitemsep,topsep=2pt]
    \item We analyze that a high-quality adversarial patch should be \textit{realistic}, \textit{position irrelevant} and \textit{printable} to be deployed in the physical world.
    \item To the best of our knowledge, it is the first work that aims to generate realistic adversarial patches using any natural images for the image classification task.
    \item We propose a novel attack called \name to craft visually realistic adversarial patches by constraining the adversarial patches in the neighborhood of a real image. Despite the common belief that adversarial examples are not printable, the adversarial patches crafted by \name can be printed to attack DNNs in the physical world.
    \item Empirical evaluations on ImageNet dataset show its high effectiveness in the digital world and ability to naturally attack real-world applications without being detected.
\end{itemize}

% However, we find that ...

\section{Related Work}

% Since Szegedy~\etal~\cite{szegedy2014intriguing} identified the vulnerability of DNNs to adversarial examples, numerous works~\cite{goodfellow2015FGSM,madry2018pgd,athalye2018obfuscated,dong2018boosting,brendel2018decision,wang2021admix,wang2022triangle} have been proposed to craft more powerful adversarial examples in various setting. Different from adversarial examples that adds imperceptible perturbation to benign samples, adversarial patches~\cite{tom2017adversarial} paste an image patch to a benign image, leading to incorrect prediction. Compared with adversarial examples, adversarial patches can be printed and added to the existing images, making it popular for attacks in physical world.

In this section, we provide a brief overview of the existing works about adversarial examples in the digital world and adversarial patches in the physical world.

% \subsection{Adversarial Examples}
\textbf{Adversarial Examples in the Digital World.}
\citet{szegedy2014intriguing} first identified the vulnerability of DNNs to adversarial examples in the digital world, which are indistinguishable from benign ones by adding tiny perturbations but lead to incorrect predictions. Recently, numerous works have been proposed to generate more powerful adversarial examples, which mainly fall into two categories: 1) \textit{White-box setting}: the attacker~\cite{moosavi2016deepfool,kurakin2017adversarial,tramer2018ensemble,wang2019atgan,croce2020reliable} can access any information of target model, \eg, outputs, architecture, (hyper-)parameters, gradient, \etc. 2) \textit{Black-box setting}: the attacker only allows access to limited information, which can be further split into three categories: a) \textit{Score-based attacks}~\cite{ilyas2018black,guo2019simple,li2019nattack,cheng2019improving} utilize the prediction probability to craft adversarial examples. b) \textit{Decision-based attacks}~\cite{brendel2018decision,cheng2019query,li2020qeba,wang2022triangle} can only access the prediction label for attack. c) \textit{Transfer-based attacks}~\cite{dong2018boosting,xie2019improving,wang2021admix,wang2021boosting} adopt the adversarial examples generated on the surrogate model to attack the target model. Adversarial examples perform very well in digital world~\cite{athalye2018obfuscated,wang2023rethinkinga,wang2023boosting}, which could attack online commercial APIs taking digital images as input~\cite{li2020qeba,wang2022triangle}, but the imperceptible perturbation makes it hard to be deployed in the physical world~\cite{tom2017adversarial}.

\textbf{Adversarial Patches in the Physical World.} Different from adversarial examples that add imperceptible perturbation to the image, adversarial patches paste a small but noticeable patch to the benign image to fool the target model, which can be printed in the physical world. \citet{tom2017adversarial} first proposed GoogleAp to generate a universal and printable adversarial patch to fool the classifier by maximizing the loss of a patch initialized by a real image patch on several images simultaneously. Currently, adversarial patches have become one of the primary forms of physical attacks~\cite{sharma2022adversarial}. LaVAN~\cite{Karmon2018lavan} shows that a small localized adversarial noise is enough to fool the classifier. PatchAttack~\cite{yang2020patchattack} adopts reinforcement learning to optimize the adversarial texture patch from a designed texture dictionary. Several works adopt generative adversarial network (GAN)~\cite{goodfellow2014generative} to craft adversarial patches. PS-GAN~\cite{liu2019perceptual} adopts a perceptual-sensitive generative adversarial
network to improve visual fidelity. PEPG~\cite{yakura2020generate} finetunes a pretrained GAN to generate virtual insects as adversarial patches to fool the classifier. GDPA~\cite{li2021generative} employs a generator to generate dynamic/static patch patterns and locations for the given input image. Adversarial patches are also widely adopted in other domains, such as object detection~\cite{huang2020universal,xu2020adversarial,hu2021naturalistic,hu2022adversarial}, face recognition~\cite{sharif2016accessorize,xiao2021improving,komkov2021advhat,wei2022adversarial} and autonomous driving~\cite{eykholt2018robust,duan2020adversarial,zolfi2021translucent,nesti2022evaluating}. To the best of our knowledge, however, except for a few adversarial patches using the real sticker to fool face recognition~\cite{wei2022adversarial} or object detector~\cite{hu2021naturalistic}, most adversarial patches are not \textit{visually realistic} so that they are significantly different from real scrawls, making them easy to be detected in the physical world. 

In this work, we argue that a high-quality adversarial patch should be \textit{visually realistic} to make it inconspicuous in the physical world. Thus, we propose a novel adversarial attack, called \name, to efficiently and effectively generate adversarial patches for the image classification task.

\section{Methodology} 
Here we first introduce preliminaries, and analyze the crucial properties of high-quality adversarial patches. Then we detail our \name. 
\subsection{Preliminaries}
Given a classifier $f$ with parameters $\theta$ and benign image $\mathbf{x} \in \mathbb{R}^{H\times W\times C}$ labeled $\mathbf{y}\in \mathbb{R}^N$, adversarial patch is defined as:

% Given a target classifier $f$ with parameters $\theta$ and a benign image $\mathbf{x} \in \mathbb{R}^{H\times W\times C}$ labeled $\mathbf{y}\in \mathbb{R}^N$, we can define the adversarial patch as follows:
\begin{definition}{(Adversarial Patch).}
    Adversarial patch is an additive patch $\bar{\mathbf{x}}\in \mathbb{R}^{H\times W\times C}$ with a location mask $\mathbf{m}\in \{0,1\}^{H\times W\times C}$ that satisfies:
    \begin{equation*}
        f(\tilde{\mathbf{x}};\theta)\neq f(\mathbf{x};\theta)=y, \ \mathrm{ where } \ \tilde{\mathbf{x}} = (1-\mathbf{m})\odot \mathbf{x} + \mathbf{m}\odot \bar{\mathbf{x}},
    \end{equation*}
    where $\odot$ is element-wise multiplication.
\end{definition}

Suppose $h$ and $w$ are the height and width and $(i,j)$ is the index of the upper left corner of the region of $1s$ in the location mask $\mathbf{m}$, $\delta \in \mathbb{R}^{h\times w\times C}$ is the corresponding additive patch of $\bar{\mathbf{x}}$, we can denote the crafted adversarial patches as:
\begin{equation}
    \tilde{\mathbf{x}} = \mathbf{x} +_{i,j} \delta,
    \label{eq:paste_patch}
\end{equation}
where $+_{i,j}$ indicates pasting the patch $\delta$ to $\mathbf{x}$ at the position $(i,j)$. We will use such notation in the rest of the paper without ambiguity. To generate adversarial patch, we can regard the attack as an optimization problem that searches an image patch $\delta$ with the size $(w,h,C)$ as well as the position $(i,j)$ that maximizes the loss function $J$ of the target classifier:
\begin{equation}
    (\delta, i, j) = \argmax_{\substack{\delta \in \mathbb{R}^{w\times h \times C}, \\ 0\le i \le H-h, \ 0\le j \le W-w}} J(\tilde{\mathbf{x}}, y; \theta),
    \label{eq:optimization}
\end{equation}
where $\tilde{\mathbf{x}}$ is calculated by Eq.~\eqref{eq:paste_patch}.

\subsection{Crucial Properties of Adversarial Patch}
\label{sec:properties}
\citet{tom2017adversarial} first proposed adversarial patch, which can be printed and added to any scene to deceive classifiers. Due to its good applicability to attack the DNNs in the physical world, it has gained increasing interests~\cite{eykholt2018robust,liu2019perceptual,xu2020adversarial,huang2020universal,xiao2021improving}. However, existing adversarial patches, even those initialized by real images~\cite{tom2017adversarial}, are often unrealistic to human perception, making them easily detectable in the physical world. Also, there are no consistent properties of adversarial patches among the existing works, for instance, whether it should be irrelated to single image~\cite{tom2017adversarial,Karmon2018lavan}, whether it should lead the DNNs to target prediction\cite{liu2019perceptual,yang2020patchattack}. Here we first analyze what crucial properties a high-quality adversarial patch should have to be deployed in the physical world:
\begin{itemize}[leftmargin=*,noitemsep,topsep=2pt]
    \item \textit{Realistic}: As noticed by previous works~\cite{liu2019perceptual}, there are often meaningful scrawls, logos or patterns on the objects in the physical world. To make the adversarial patches inconspicuous, they should be visually realistic so that humans mistakenly take them as such real scrawls.
    \item \textit{Position irrelevant}: Since we need to add the patches to the scenes manually, it is hard to place the patches in a precise position. Hence, a high-quality adversarial patch should be related to its position so that it can be easy to be deployed in the physical world.
    \item \textit{Printable}: Adversarial patches should work effectively in the digital world as well as the physical world. Thus, it must be printable to be pasted into physical objects.
\end{itemize}

We argue that the above properties are necessary to make adversarial patches inconspicuous in the physical world. Other significant properties are beyond our discussion. For instance, the patches are universal for various scenes~\cite{huang2020universal} or the patches cover objects with arbitrary shapes~\cite{hu2022adversarial}. In this work, we show that adding perturbation to any image patches can craft realistic adversarial patches and design a novel attack in the next section.
% as described in Sec.~\ref{sec:vrap}.

\subsection{Crafting Visually Realistic Adversarial Patch}
\label{sec:vrap}
%\NOTE{In this section, you should detailedly introduce our proposed method.}

Based on the above analysis, visual reality is of utmost importance for a high-quality adversarial patch. PS-GAN~\cite{liu2019perceptual} employs a perceptual-sensitive generative adversarial network to enhance the visual fidelity for traffic sign recognition tasks, but the crafted adversarial patches are still not as realistic as real scrawls. \citet{wei2022adversarial} manipulate the actual sticker's position and rotation angle on the objects for face recognition. However, precise positioning and angle adjustments can pose deployment challenges in the physical world. \citet{hu2021naturalistic} sample the optimal image patch from a pretrained generative adversarial network for a high-quality adversarial patch against an object detector. However, these patches tend to be large, almost covering half of a T-shirt, which may not be ideal for practical applications. In this work, we first explore that: 

\textit{How can we guarantee the visual reality of adversarial patches while maintaining high attack performance?} 

Note that adversarial examples are visually realistic as they add tiny perturbations to the benign samples to make them imperceptible~\cite{szegedy2014intriguing}. This inspires us that constraining the image patches in the $\epsilon$-neighborhood of benign samples can ensure the visual reality. Thus, we first investigate whether pasting such image patches to benign images on a fixed position can fool the target model:
% , which can be formulated as follows:
\begin{equation}
    \delta = \argmax_{\delta \in \mathbb{B}_\epsilon(\bar{\mathbf{x}})} J(\mathbf{x}+_{i,j}\delta,y;\theta),
    \label{eq:delta_optimization}
\end{equation}
where $(i,j)$ is a fixed position, $\bar{\mathbf{x}}$ is the benign image patch with the size of $(w,h,C)$ and $ \mathbb{B}_\epsilon(\bar{\mathbf{x}})=\{\mathbf{x'}:\|\mathbf{x'}-\bar{\mathbf{x}}\|\le\epsilon\}$. Inspired by PGD optimization procedure~\cite{madry2018pgd}, we solve Eq.~\eqref{eq:delta_optimization} iteratively as follows:
\begin{equation*}
    \delta_{t+1} = \Pi_{\mathbb{B}_\epsilon(\bar{\mathbf{x}})}[\delta_t + \alpha\cdot \operatorname{sign}(\nabla_{\delta_t} J(\mathbf{x}+_{i,j} \delta_t, y; \theta))],
\end{equation*}
where $\delta_0=\bar{\mathbf{x}}$, $\nabla_{\delta} J(\mathbf{x}+_{i,j} \delta, y; \theta)$ is the gradient of loss function \wrt $\delta$, $\Pi_{\mathbb{B}_\epsilon(\bar{\mathbf{x}})}(\cdot)$ projects the input into $\mathbb{B}_\epsilon(\bar{\mathbf{x}})$ and $\operatorname{sign}(\cdot)$ denotes the sign function. 

We conduct the above attack in a randomly fixed position $(i,j)$ on VGG-16 model in the digital world with different perturbation budget $\epsilon$. As shown in the Appendix, we can successfully generate visually realistic adversarial patches in the neighborhood of the benign image patches with high attack performance. However, it is well-known that the imperceptible adversarial perturbations are too hard for printing to be deployed in the physical world. This brings us another issue:

\textit{How can we make such visually realistic adversarial patches printable?}

To address this issue, we analyze what information will be lost when printing the image. 1) The range of colors that printers can reproduce is a subset of RGB space, \ie, the printer introduces a small deviation to the brightness (RGB colors in the digital world). Thus, adversarial patches should not be sensitive to such deviations. To mitigate this effect, we use gamma correction to ensure that the poorest patch in the neighborhood of the current patch can fool the target model:
\begin{equation*}
    \operatorname{G}(\delta, \lambda, \gamma) = \lambda \cdot \delta^\gamma,
\end{equation*}
where $\lambda$ and $\gamma$ are the learnable parameters in this work. 2) Due to the limited range of colors, the contrast of image would decrease so that the printed image might lose some local details. Therefore, the colors in the generated adversarial patches should change gradually like natural images. To maintain the smoothness of perturbation, we add the total variation loss as a regularizer:
\begin{equation*}
    \mathcal{L}_{TV}(\delta) = \sum_{i,j} |\delta_{i,j+1} - \delta_{i,j}| + |\delta_{i+1,j} - \delta_{i,j}|.
\end{equation*}
To generate the printable adversarial patches, Eq.~\eqref{eq:delta_optimization} is rewritten as:
\begin{equation*}
    \delta = \argmax_{\delta\in \mathbb{B}_\epsilon(\bar{\mathbf{x}})} \min_{\lambda, \gamma} J(\mathbf{x}+_{i,j}\operatorname{G}(\delta,\lambda,\gamma),y;\theta) + \mathcal{L}_{TV}(\delta).
\end{equation*}

\begin{figure}
    \centering
    \includegraphics[width=\linewidth]{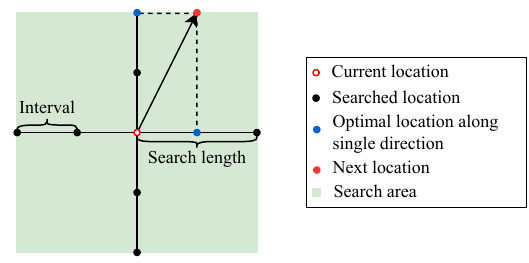}
    % \vspace{-0.5em}
    \caption{An illustration of local search for the next position.}
    \label{fig:position}
    \vspace{-0.5em}
\end{figure}

\begin{figure*}
    \begin{minipage}{.23\textwidth}
        \begin{subfigure}{\textwidth}
        \includegraphics[width=\linewidth]{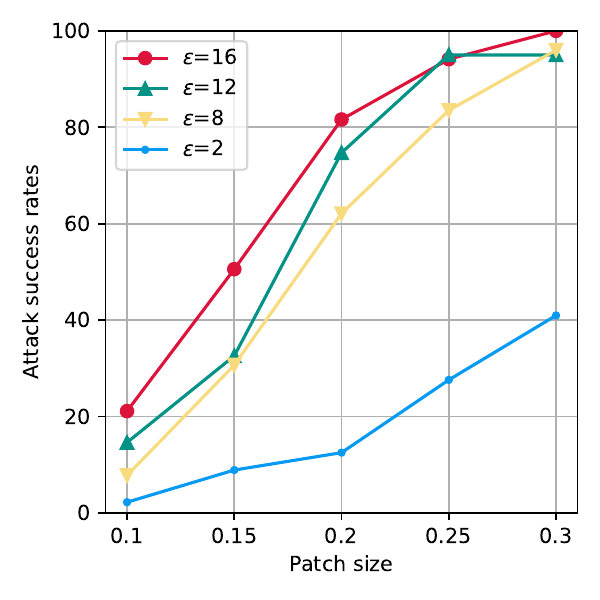}
        \caption{VGG-16}
        \end{subfigure}
    \end{minipage}%
    \hspace{1em}
    \begin{minipage}{.23\textwidth}
        \begin{subfigure}{\textwidth}
        \includegraphics[width=\linewidth]{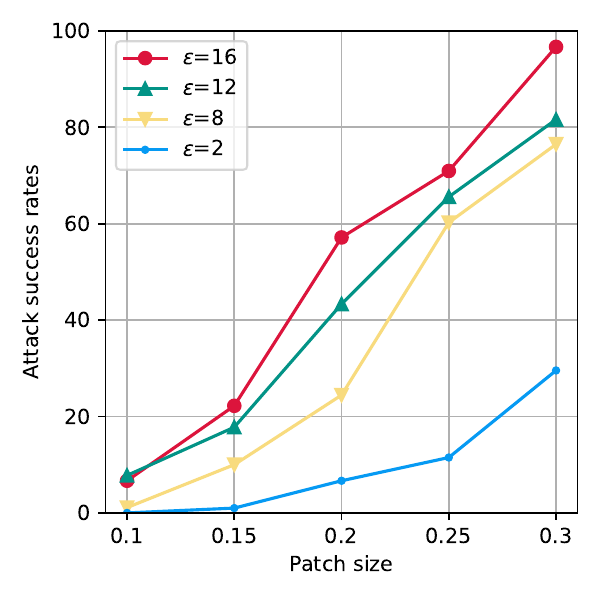}
        \caption{ResNet-18}
        \end{subfigure}
        %\hspace{1em}
    \end{minipage}
    \hspace{1em}
    \begin{minipage}{.23\textwidth} 
        \begin{subfigure}{\textwidth}
        \includegraphics[width=\linewidth]{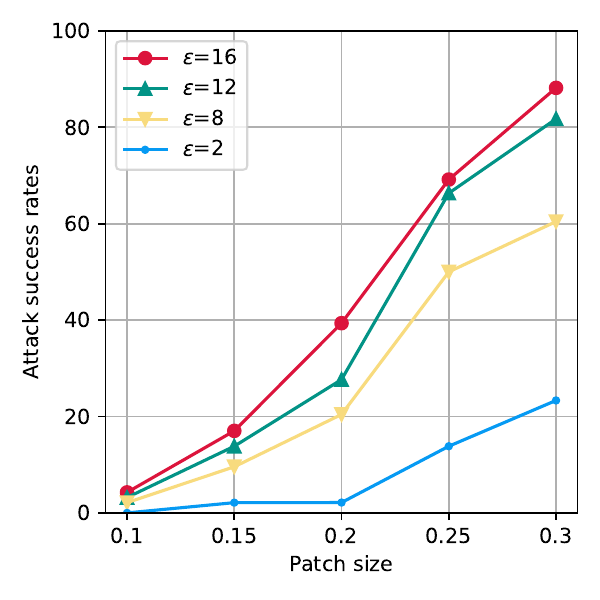}
        \centering
        \caption{DenseNet-121}
        \end{subfigure}
        
    \end{minipage}
    \hspace{1em}
    \begin{minipage}{.23\textwidth} 
        \begin{subfigure}{\textwidth}
        \includegraphics[width=\linewidth]{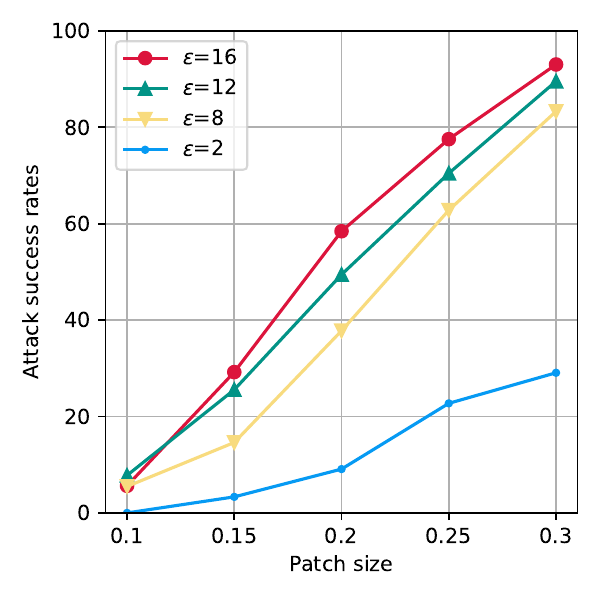}
        \caption{MobileNet-v2}
        \end{subfigure}
    \end{minipage}
    % \hspace{1em}
    % \begin{minipage}{.17\textwidth} 
    %     \includegraphics[width=\linewidth]{figs/pdf/attack_ability/attackvit.pdf}
    %     \centering
    %     \text{Vit-b-16}
    % \end{minipage}
    % \vspace{-1em}
    \caption{Attack success rates (\%) of \name on four models with various perturbation budgets and patch sizes.}
    \label{fig:asr}
    % \vspace{-1em}
\end{figure*}
\begin{figure*}
    \begin{minipage}{0.23\textwidth}
        \begin{subfigure}{\textwidth}
        \includegraphics[width=\linewidth]{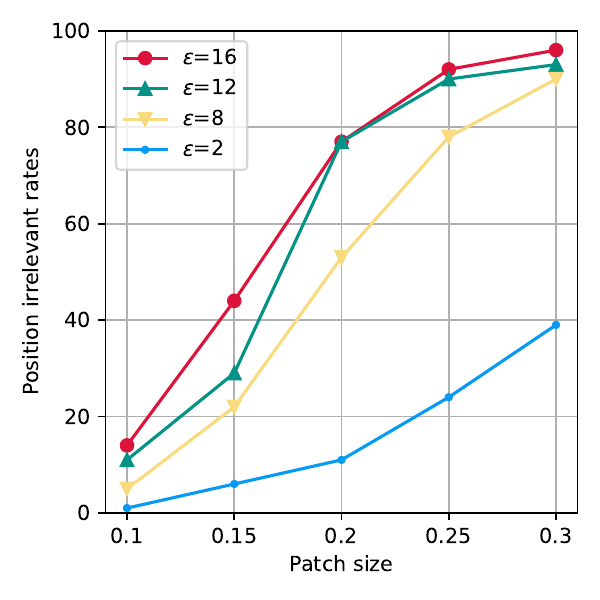}
        \caption{VGG-16}
        \end{subfigure}
    \end{minipage}%
    \hspace{1em}
    \begin{minipage}{0.23\textwidth}
        \begin{subfigure}{\textwidth}
        \includegraphics[width=\linewidth]{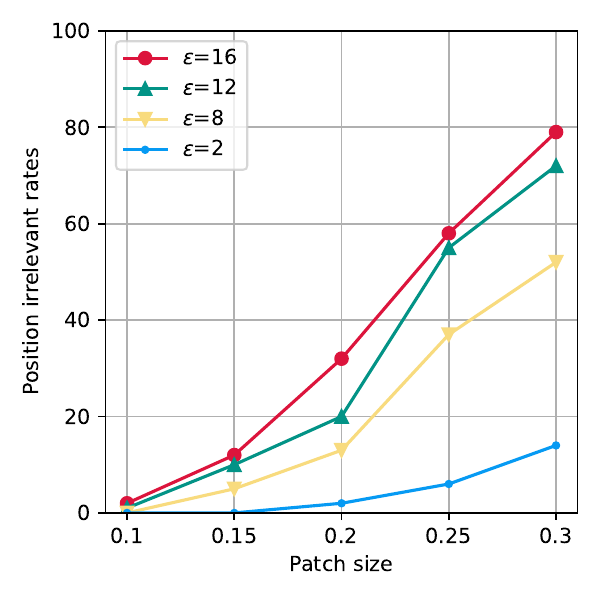}
        \caption{ResNet-18}
        \end{subfigure}
    \end{minipage}%
    \hspace{1em}
    \begin{minipage}{0.23\textwidth}
        \begin{subfigure}{\textwidth}
        \includegraphics[width=\linewidth]{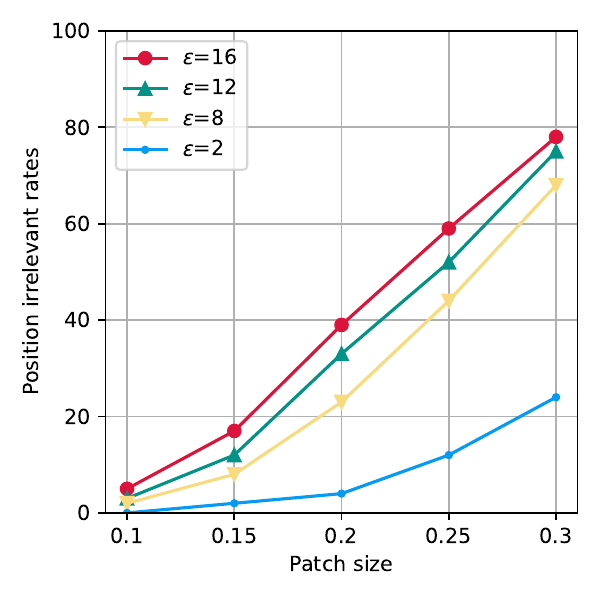}
        \caption{DenseNet-121}
        \end{subfigure}
    \end{minipage}%
    \hspace{1em}
    \begin{minipage}{0.23\textwidth}
        \begin{subfigure}{\textwidth}
        \includegraphics[width=\linewidth]{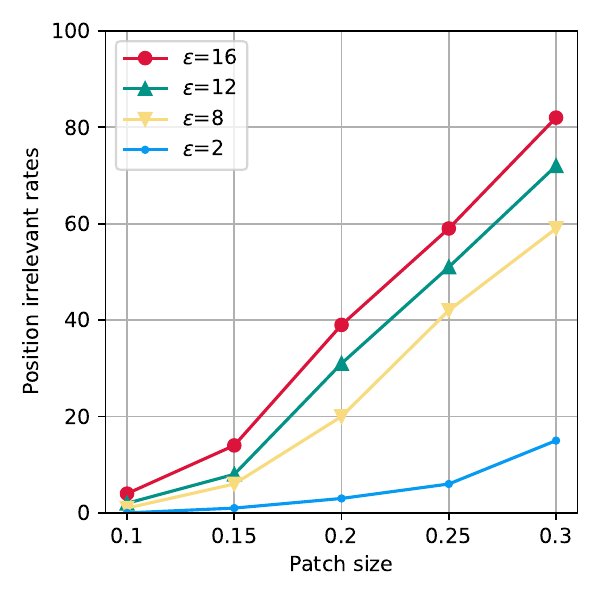}
        \caption{MobileNet-v2}
        \end{subfigure}
    \end{minipage}
    % \hspace{1em}
    % \begin{minipage}{0.17\textwidth}
    %     \includegraphics[width=\linewidth]{figs/pdf/ir/irvit.pdf}
    %     \centering
    %     \text{Vit-b-16}
    % \end{minipage}
     % \vspace{-0.5em}
    \caption{Position irrelevant rates (\%) of \name on four models with various perturbation budgets and patch sizes.}
    \label{fig:pir}
    \vspace{-1em}
\end{figure*}
The position irrelevance means that the generated adversarial patch pasted to various positions of the given image can successfully fool the target model. However, directly optimizing the position is an integer programming problem, which is NP-complete. With numerous potential positions in the given image, we adopt local search to find the poorest position around the current position, \ie, the position $(i^*,j^*)$ in the neighborhood of position $(i,j)$ that minimizes the loss:
\begin{equation}
    (i^*,j^*) = \argmin_{\substack{i-k\le i' \le i+k, \\ j-k\le j' \le j+k}} J(\mathbf{x}+_{i'j'}\delta,y;\theta),
    \label{eq:position}
\end{equation}
where $k$ is the search range. Since each query needs forward propagation, the grid search still takes a large number of queries, resulting in a huge computation cost. As shown in Fig.~\ref{fig:position}, we search the position along x-axis and y-axis to find the optimal position $(i^*,j)$ and $(i,j^*)$. Then we approximately combine the best positions in these directions as the optimal position $(i^*, j^*)$. Since shifting the patch within a few pixels does not bring obvious distinction, we search the positions with a stepsize $\tau$ to further boost the efficiency.
\begin{algorithm}[tb]
    \algnewcommand\algorithmicinput{\textbf{Input:}}
    \algnewcommand\Input{\item[\algorithmicinput]}
    \algnewcommand\algorithmicoutput{\textbf{Output:}}
    \algnewcommand\Output{\item[\algorithmicoutput]}

    \caption{VRAP}
    \label{alg:VRAP}
	\begin{algorithmic}[1]
		\Input A classifier $f$ with the parameters $\theta$ and loss function $J$; a raw example $x$ with the ground-truth label $y$, and the initial real image patch $\bar{x}$;
		the perturbation budget $\epsilon$; the number of iteration $T$; search range $k$ and stepsize $\tau$
        \Output An adversarial patch $\delta$
        % where $\|x^{adv}-x\|_\infty \leq \epsilon$.
		\State $\alpha = \epsilon/T$, $\beta = \epsilon/T$, $g_0 = 0$, $\delta_0=\hat{x}$, $\lambda=1$, $\gamma = 1$
		\State Randomly sample the position $(i,j)$
		\For{$t = 0 \rightarrow T-1$}
		    \State update position $(i,j)$ by optimizing Eq.~\eqref{eq:position} using local search
		    \State Calculate the gradient w.r.t. $\delta$:
		    \begin{equation*}
		        g_{t} = \nabla_{\delta}[J(\mathbf{x}+_{i,j}\operatorname{G}(\delta_{t},\lambda,\gamma), y; \theta) + L_{TV}(\delta_t)]
		    \end{equation*}
		    \State Update the adversarial patch $\delta_t$:
		    \begin{equation*}
		        \delta_{t+1} = \Pi_{\mathbb{B}_\epsilon(\bar{x})}[\delta_t + \alpha \cdot \operatorname{sign}(g_{t})]
		    \end{equation*}
		    \State Calculate the gradient w.r.t. $\lambda$ and $\gamma$:
		    \begin{equation*}
		        g_{\lambda}, \ g_{\gamma} = \nabla_{\lambda, \gamma}[J(\mathbf{x}+_{i,j}\operatorname{G}(\delta_{t+1},\lambda,\gamma), y; \theta) + L_{TV}(\delta_{t+1})]
		    \end{equation*}
		    \State Update $\lambda$ and $\gamma$ using gradient descent:
		    \begin{equation*}
		        \lambda = \lambda - \beta \cdot g_{\lambda}, \quad 
		        \gamma = \gamma - \beta \cdot g_{\gamma}
		    \end{equation*}
		\EndFor
        \State \Return $\delta_T$
	\end{algorithmic} 
\end{algorithm} 
In summary, the objective function can be formulated as:
\begin{equation*}
    \delta = \argmax_{\delta\in \mathbb{B}_\epsilon(\bar{\mathbf{x}})} \min_{\substack{\lambda, \gamma, \\ 0\le i \le H-h, \\ 0\le j \le W-w}} J(\mathbf{x}+_{i,j}G(\delta,\lambda,\gamma),y;\theta) + \mathcal{L}_{TV}(\delta),
\end{equation*}
in which $\bar{\mathbf{x}}$ is any benign image with the size of $(h,w,C)$. The algorithm of the proposed Visually Realistic Adversarial Patch attack (\name) is summarized in Algorithm~\ref{alg:VRAP}.

\section{Experiments}
In this section, we evaluate the visually realistic adversarial patches in the digital as well as physical world, and explore why these realistic patches can effectively attack DNNs.
% In this section, we provide experimental evidence to evaluate the effectiveness of our purposed method. We first clarify the experimental setting, and illustrate the attacking ability of our methods based on several models with different perturbation rate and patch size. Secondly, we evaluate our attacking ability in terms of position irrelevance. Furthermore, we test our purposed method in real world. The experiments results demonstrate the crafted patch using our method is printable, position irrelevant, and realistic. Finally, we present ablation study on hyperparameter search len and search range.

\subsection{Experimental Setting}
We summarize our experimental setting as follows:

\textbf{Dataset.} We use 1,000 images pertaining to 1,000 classes from ILSVRC 2012 validation set~\cite{Russa2015imagenet}, which are almost correctly classified by the chosen deep models.
% and we randomly choose one of the image each time as the initial patch.

\textbf{Models.} We utilize several popular deep models with different architectures for image classification, including VGG-16~\cite{Simonyan2015VGG}, ResNet-18~\cite{he2016resnet}, DenseNet-121~\cite{huang2017densely}, MobileNet-v2~\cite{Mark2018mobile}. All the classifiers are trained on ImageNet and achieve a classification accuracy of 90.4\%, 93.4\%, 90.2\%, and 95.3\% on our evaluation dataset. We also study several adversarial patch defense models, including DW~\cite{hayes2018on}, and LGS~\cite{naseer2019local}, PatchGuard~\cite{xiang2021patchguard}, CBN~\cite{zhang2020clipped}, and several methods against adversarial examples, \ie, Random Smoothing~\cite{Cohen2019RS} and FastAT~\cite{wong2020fast}.%, and certified defense method \ie De-Randomized Smoothing\cite{levine2020derandomized}.

\textbf{Evaluation Metrics.} We mainly evaluate the attack success rate of \name, the portion of images with the adversarial patches can mislead the target model but the images with original patches can be correctly classified. Given an image set $\mathcal{X}$, we can formulate the attack success rate (ASR) as:
\begin{equation*}
    \operatorname{ASR}(\mathcal{X}) = \frac{1}{|\mathcal{X}|} \sum_{x\in \mathcal{X}} \mathbb{I}(f(x+_{i,j} \delta)\neq y) \cdot \mathbb{I}(f(x+_{i,j} \delta_0) = y),
\end{equation*}
where $\mathbb{I}(\cdot)$ is the indicator function, $\delta$ is the adversarial patch and $\delta_0$ is the original patch. To evaluate the position irrelevant performance, we also define the position irrelevant rates (PIR) for a give image $x$ as:
\begin{equation*}
    \operatorname{PIR}(x) = \frac{\sum_{\substack{0\le \tau \cdot i\le H-h \\ 0\le \tau \cdot j\le W-w}}\mathbb{I}(f(x+_{\tau \cdot i, \tau \cdot j}\delta)\neq y))}{\lfloor(H-h)/\tau\rfloor \lfloor(W-w)/\tau\rfloor},
\end{equation*}
where $\tau$ is the step size, $(H, W)$ and $(h,w)$ are the size of the input image $x$ and the adversarial patch, respectively.

\textbf{Hyper-parameters.} We adopt the search range of $10$ with stepsize of $5$ to conduct the main evaluations.

\subsection{Evaluation on Attacking Ability}
\label{sec:exp:attack}
% In this section, we evaluate the attack success rate on several models (e.g. VGG, ResNet, DenseNet, MobileNet, etc.) using various epsilon (e.g. epsilon=2,4,6,8,10 ...16). We can use a figure to illustrate it.

To validate the effectiveness of \name, we evaluate the attack success rates and position irrelevant rates of adversarial patches with various patch sizes and perturbation budgets on four widely adopted deep models. We also visualize the adversarial patches in the Appendix.
The patch size is the relative size of the patch compared with the input image.
% , while the perturbation budget is the size of $\epsilon$-neighboorhood defined in Eq.~\eqref{eq:delta_optimization}. 
% Finally, we provide several adversarial patches generated by \name to verify their visual reality.

\textbf{Attack success rates.} The attack success rates on four deep models are summarized in Fig.~\ref{fig:asr}. In general, increasing the patch size can consistently improve the attack success rates on all four models, since the patch occupies a larger proportion in the original image. Besides, the perturbation budget constrains the search space for the adversarial patch. As expected, we can observe that the larger perturbation budget results in better attack success rates with the same patch size. In particular, when the patch size is $0.3$ with the perturbation budget $\epsilon=16$, the attack success rates are larger than $90.0\%$ on four models. Besides, on VGG-16 with the patch size of $0.3$, \name can achieve $100.0\%$ attack success rate with $\epsilon=16$ and more than $95\%$ when the perturbation budget $\epsilon \ge 8$. Such high and stable attack performance on the four deep models shows the remarkable effectiveness of \name to generate visually realistic adversarial patches.

\textbf{Position irrelevant rates.} Since we need to paste the adversarial patch to real-world objects manually, it is hard to place them in a precise position, making it crucial for the patch to be position irrelevant. To verify such property of adversarial patches crafted by \name, we calculate the position irrelevant rates on four models and summarize the results in Fig.~\ref{fig:pir}. Note that we adopt a brute force algorithm to calculate the position irrelevant rates, in which a $224\times 224 \times 3$ image and an adversarial patch with the patch size of $0.3$ take thousands of predictions using the interval $\tau=5$. Due to such colossal computation cost, we adopt 100 images for evaluation. Similar to the attack success rates, a larger patch size or perturbation budget also boosts the position irrelevant rates. In most cases, \name can achieve the position irrelevant rate of more than $40.0\%$ on all four models. Specifically, with the patch size of $0.3$ and perturbation budget $\epsilon=16$, \name achieves the position irrelevant rate near $100.0\%$ on VGG-16 and around $80.0\%$ on the other three models. Such superior performance makes \name easy to be deployed in the physical world without careful placement.
\begin{figure}[htb]
    \begin{minipage}{0.155\textwidth}
        \begin{subfigure}{\textwidth}
        \includegraphics[width=\linewidth]{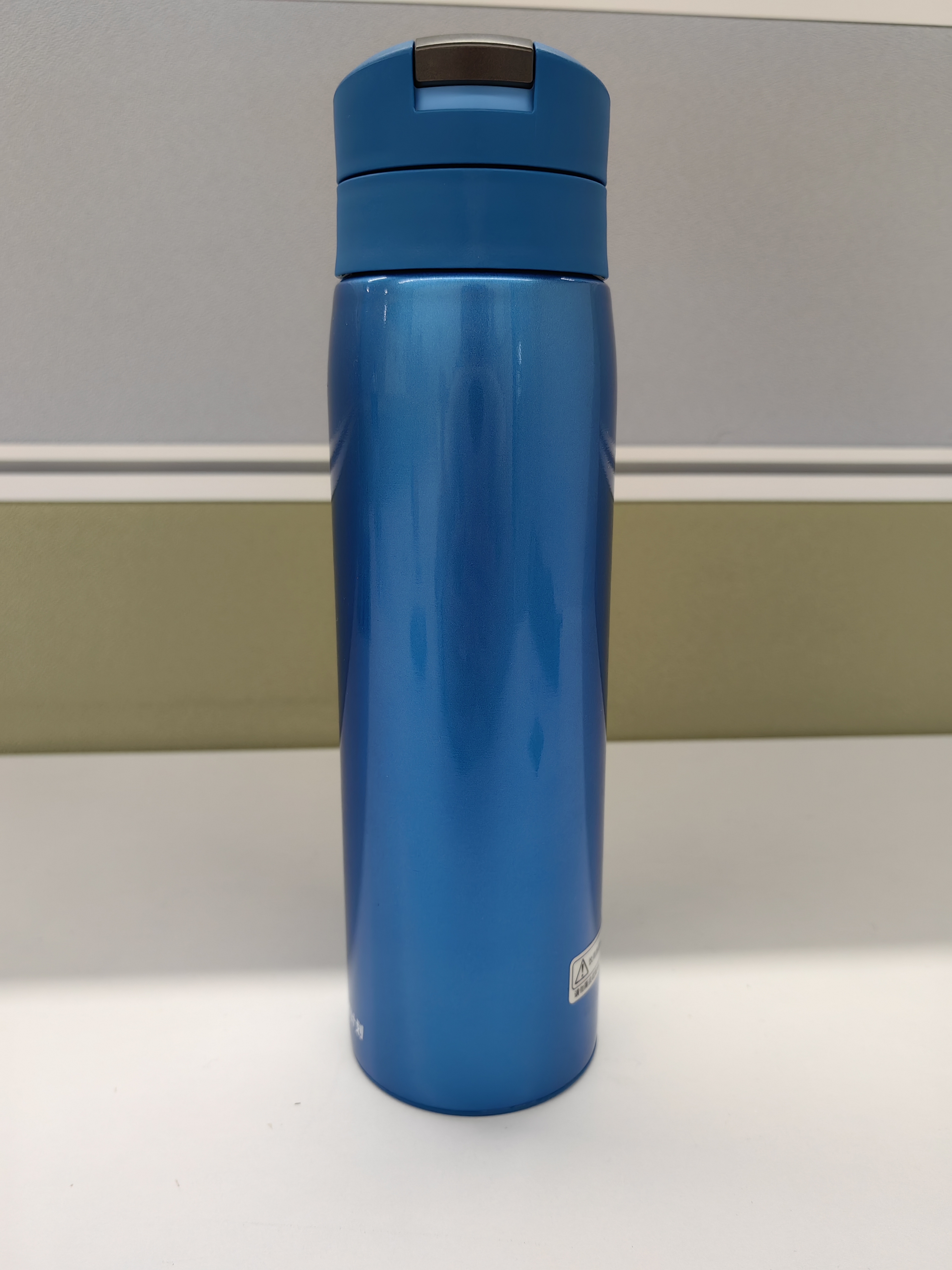}\\
        \vspace{-0.3cm}
        \includegraphics[width=\linewidth]{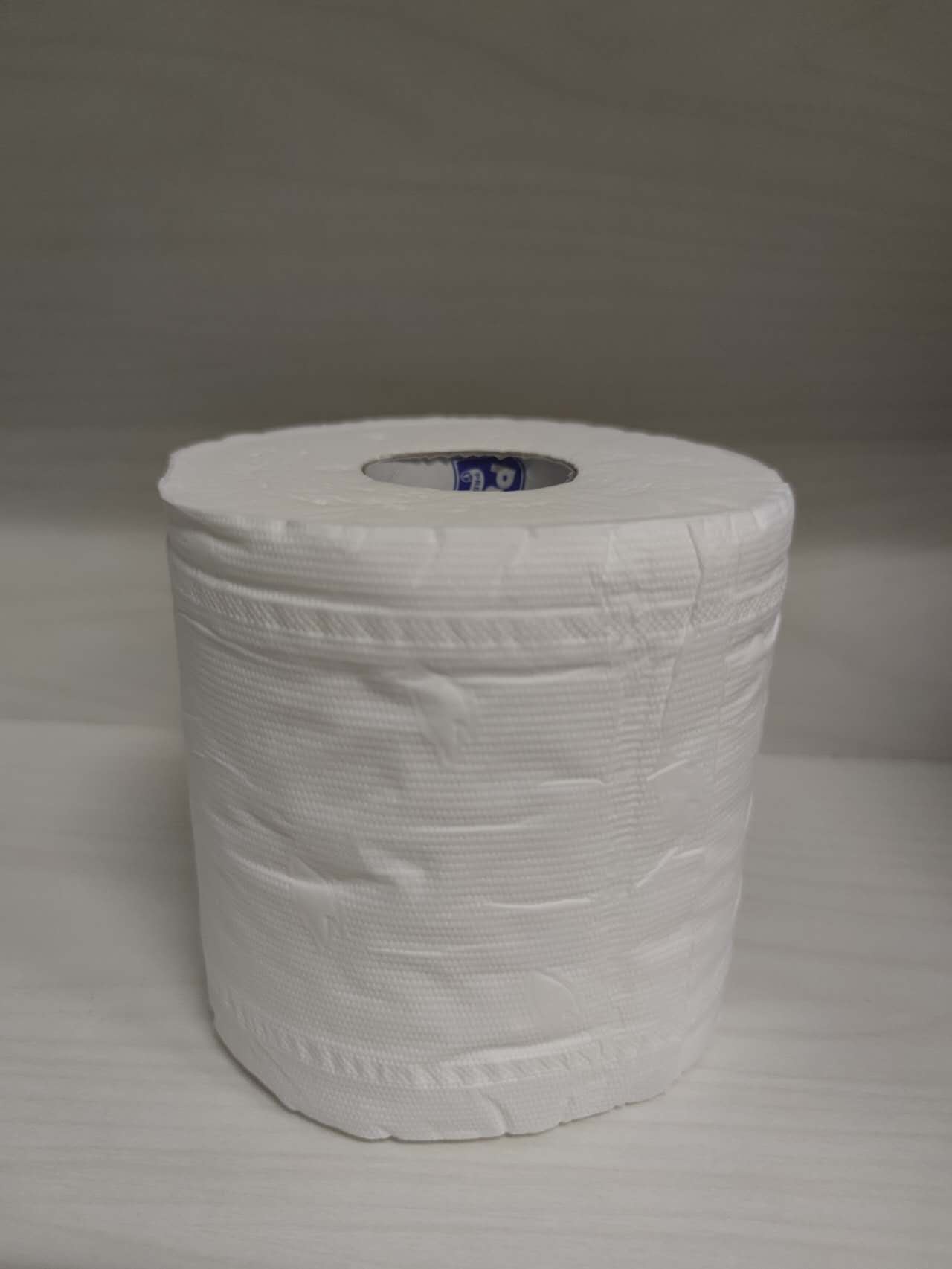}\\
        \vspace{-0.3cm}
        \includegraphics[width=\linewidth]{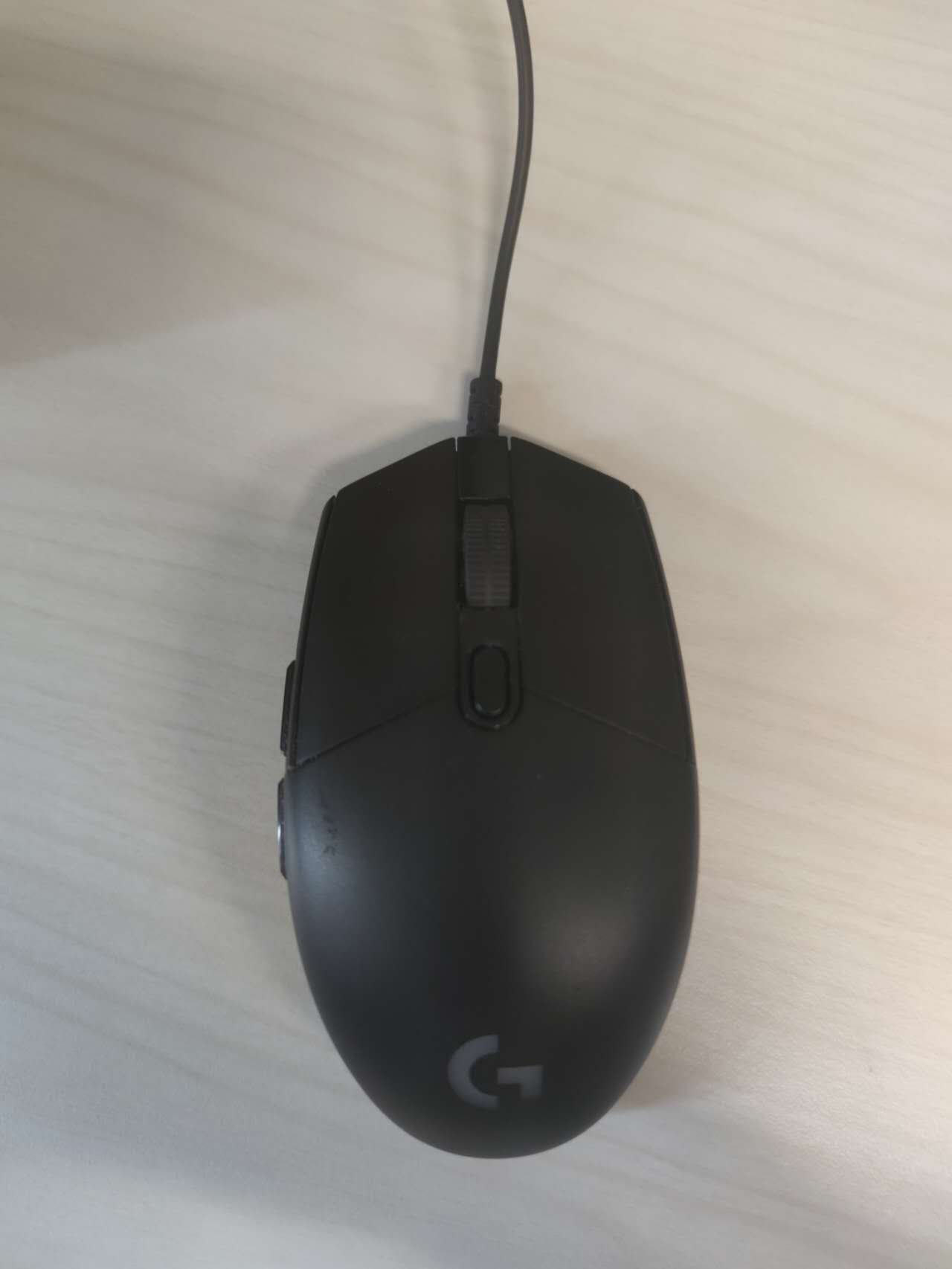}
        \caption{Raw}
        \label{fig:physical_world:raw}
        \end{subfigure}
    \end{minipage}%
    \hspace{0.1em}
    \begin{minipage}{0.155\textwidth}
        \begin{subfigure}{\textwidth}
        \includegraphics[width=\linewidth]{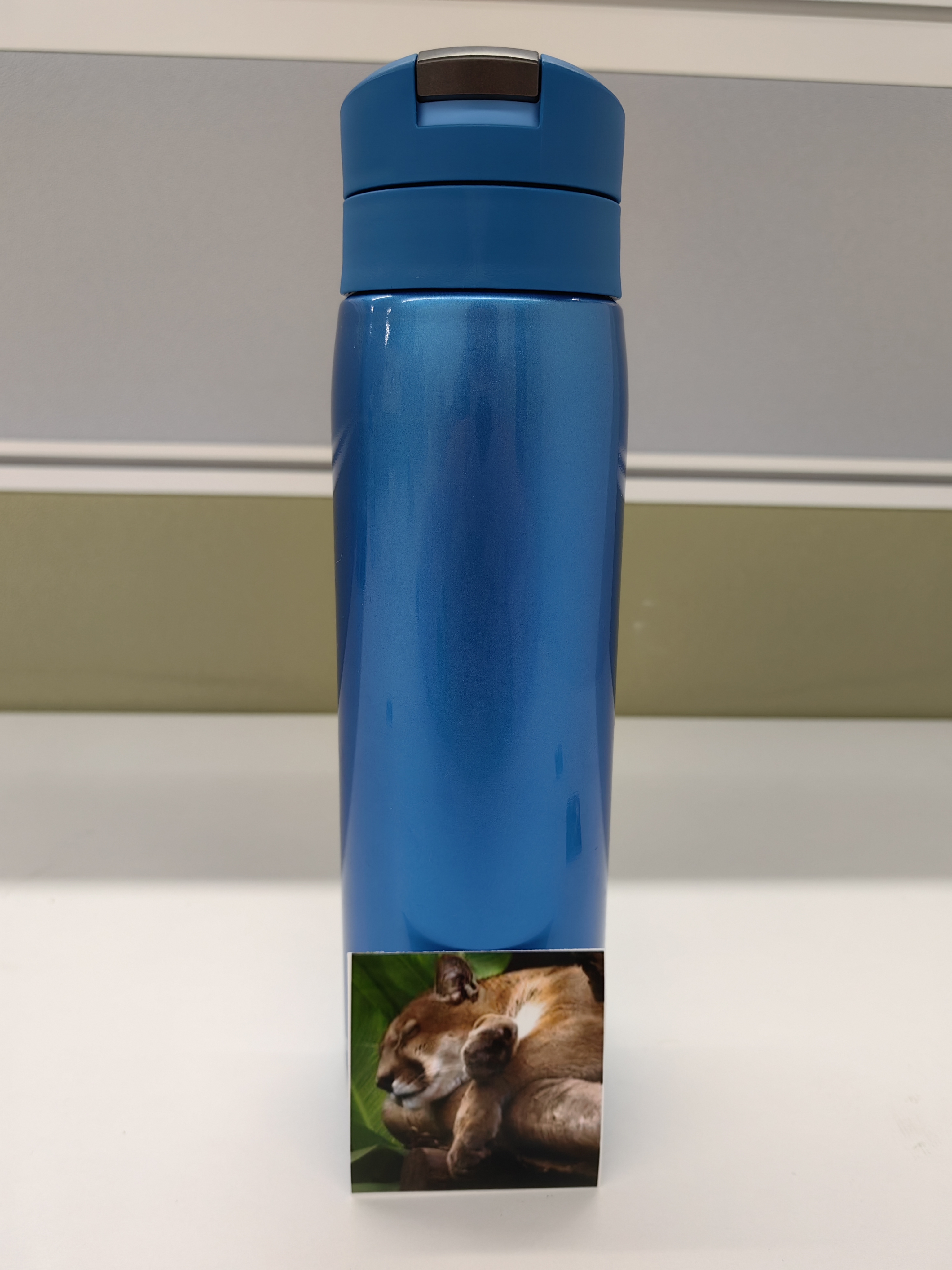}\\
        \vspace{-0.3cm}
        \includegraphics[width=\linewidth]{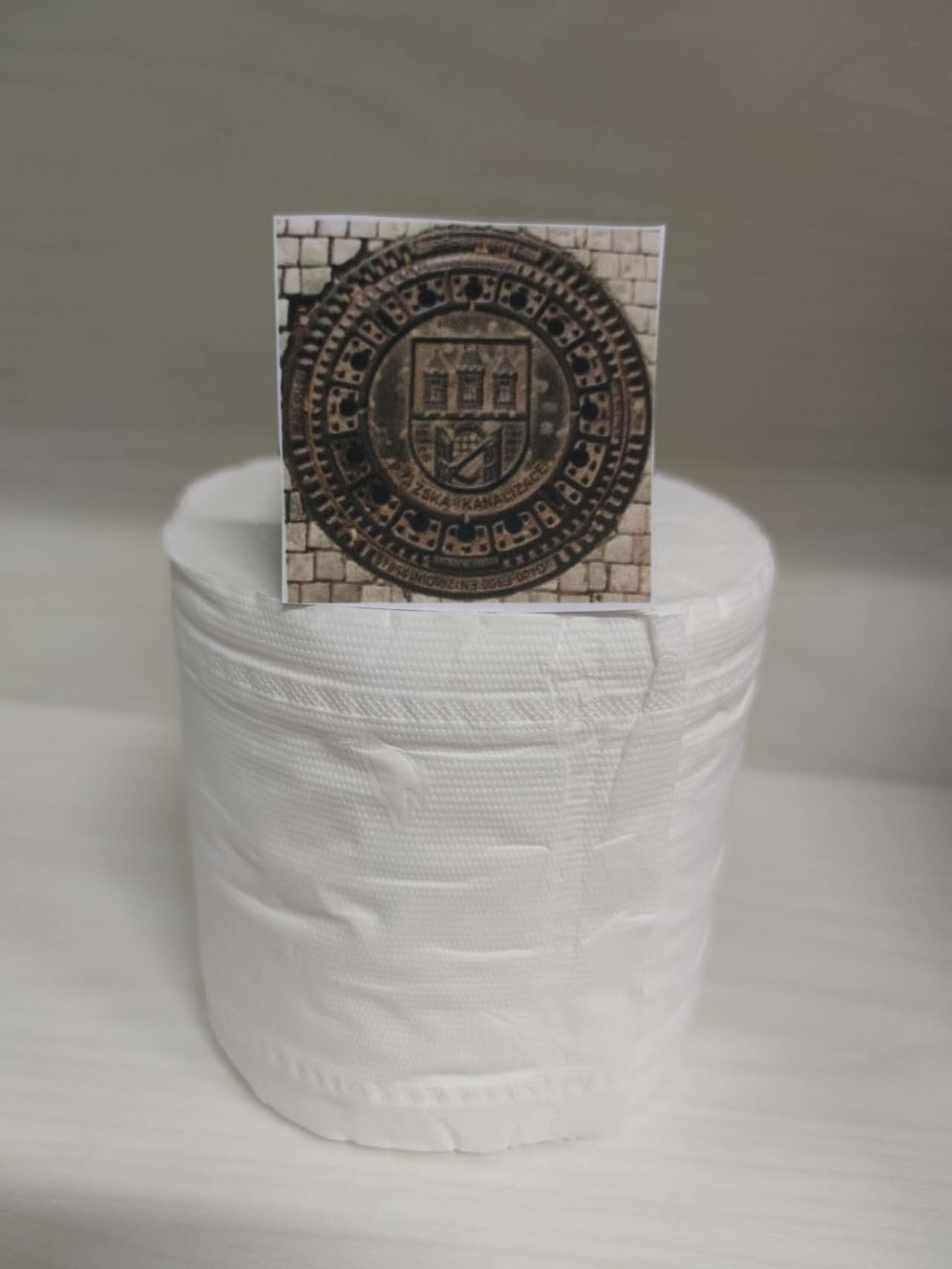}\\
        \vspace{-0.3cm}
        \includegraphics[width=\linewidth]{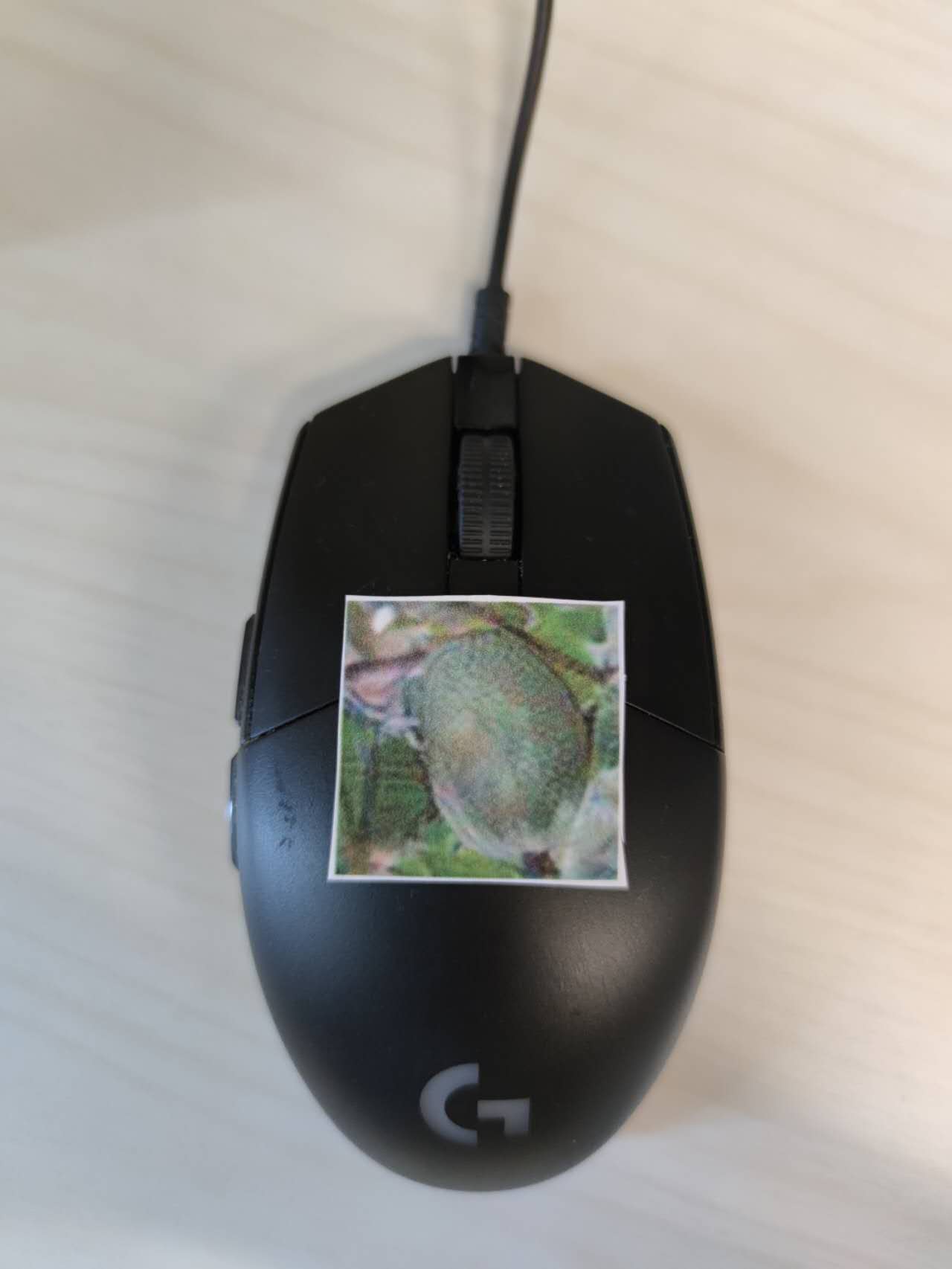}
        \caption{Original}
        \label{fig:physical_world:original}
        \end{subfigure}
    \end{minipage}%
    \hspace{0.1em}
    \begin{minipage}{0.155\textwidth}
        \begin{subfigure}{\textwidth}
            \includegraphics[width=\linewidth]{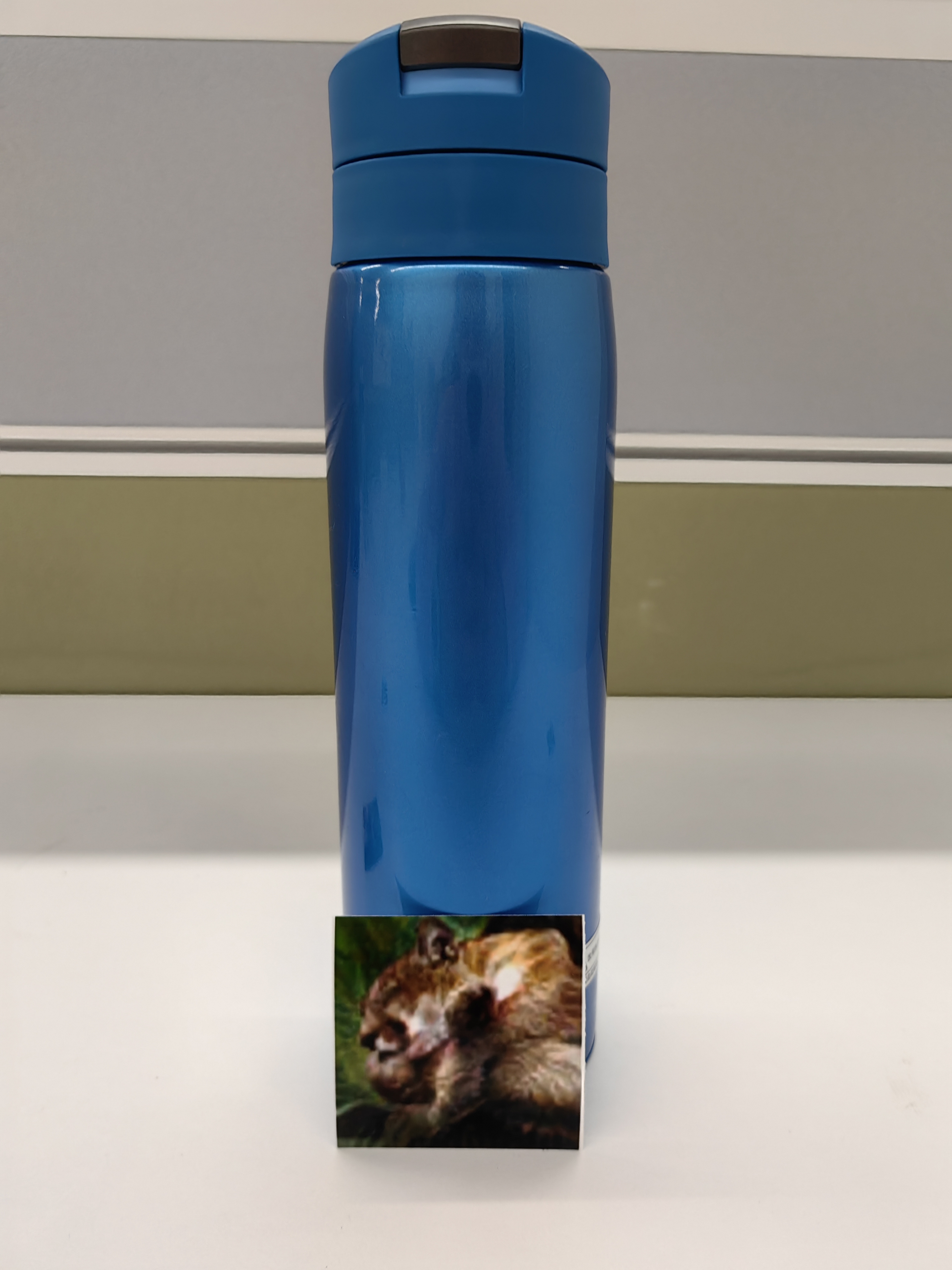}\\
            \vspace{-0.3cm}
            \includegraphics[width=\linewidth]{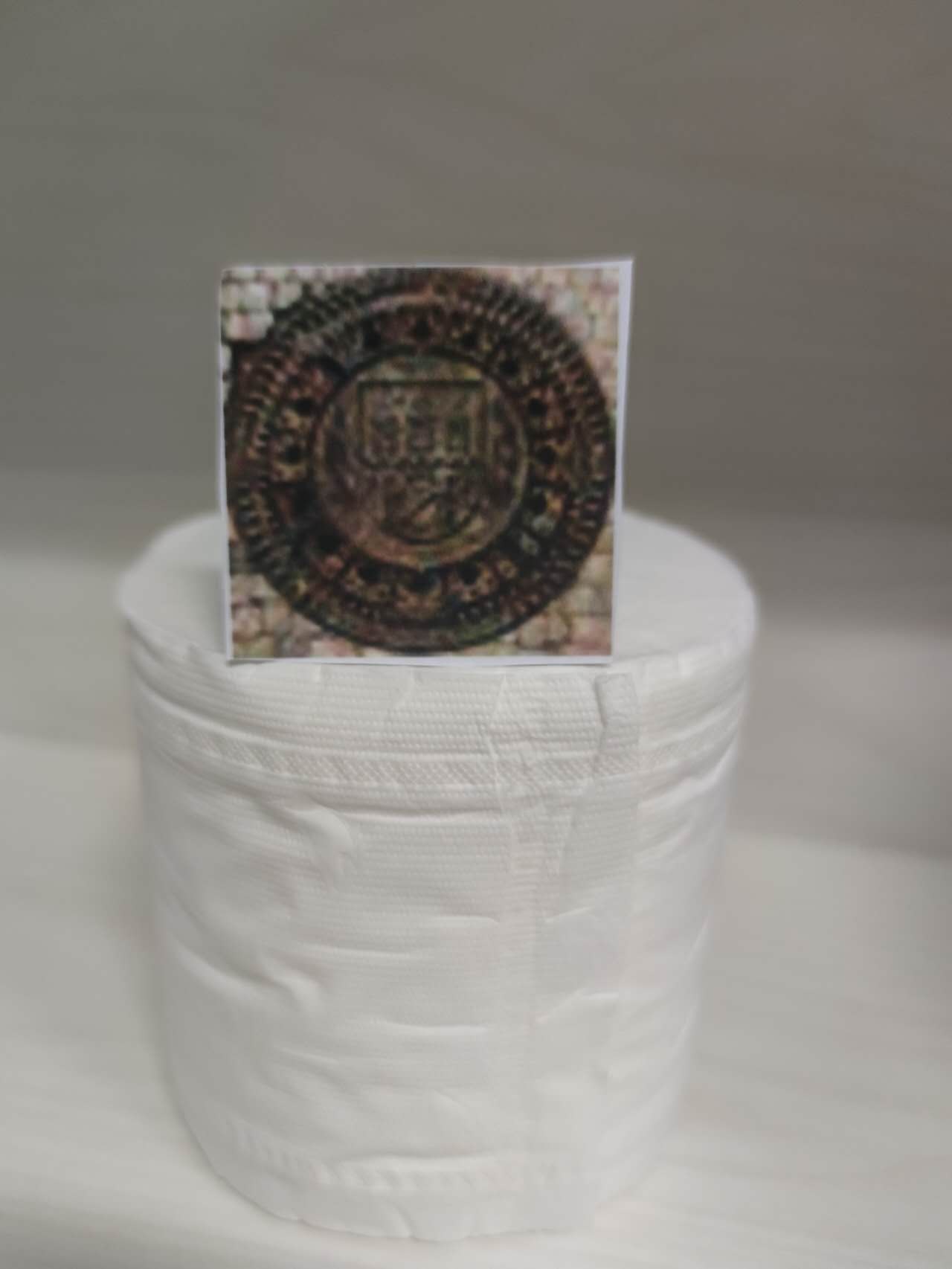}\\
            \vspace{-0.3cm}
            \includegraphics[width=\linewidth]{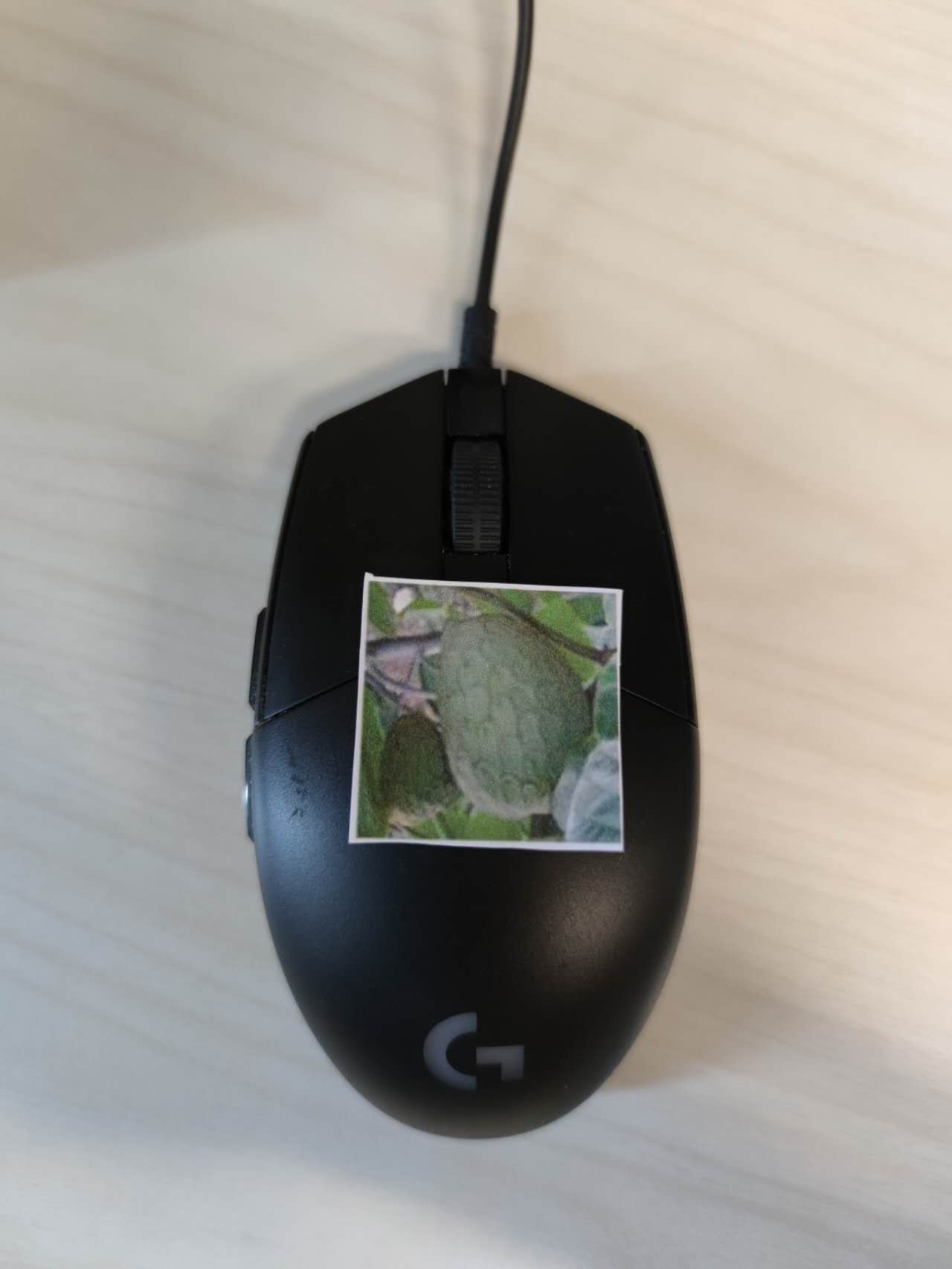}
            \caption{Adversarial}
            \label{fig:physical_world:adversarial}
        \end{subfigure}
    \end{minipage}
    \caption{The raw images, images with original image patches, and adversarial patches in the physical world. See more samples in Appendix.}
    \label{fig:physical_world}
    %\vspace{-1.5em}
\end{figure}
\begin{table*}[tb]
    \centering

    \begin{tabular}{cccccccc}
        \toprule
         Metric &LGS & DW & PatchGuard & CBN &DRS & Avg.\\
         \midrule
         ASR (\%)& 94.9 & 92.2& 80.8 & 65.6 &33.3 & 73.4\\
         PIR (\%)& 89.8& 90.1&73.5&62.2 &23.5 & 61.8\\
         \bottomrule
    \end{tabular}
    \caption{Attack success rates (\%) and position irrelevant rates (\%) of five defense methods.}
    \label{tab:defense}
    % \vspace{1.5em}
\end{table*}

\subsection{Evaluation on Defense Method}
Here we evaluate the effectiveness of \name against several adversarial patch defense methods including DW, LGS, PatchGuard, and CBN. For DW and LGS, we craft the adversarial patch of size 0.3 and $\epsilon = 16$ with VGG-16. For PatchGuard and CBN, we follow their experiment setting in their papers and craft the adversaries on BagNet-17 to evaluate our proposed methods. To further validate the effectiveness of \name, we also consider two adversarial defense methods, \ie, RS and FastAT.

The attack performance is summarized in Tab~\ref{tab:defense}. \name can pass adversarial patch defense methods with a high average attack success rate of 83.4\%, and a high position irrelevant rate of 78.9\%. It validates that our \name can bypass adversarial patch mechanisms. For certified and adversarial training methods, \name exhibitS weaker performance, the average attack success rate and position irrelevant rate against several defense methods is  65.2\% and 58.9\% respectively. It further validates the effectiveness of \name.

\subsection{Evaluation in Physical World}

% \textbf{Visualization of the adversarial patches.} To intuitively perceive the reality of adversarial patches, we visualize the raw patch, adversarial patch with the perturbation budget $\epsilon=8$ and $\epsilon=16$ in Fig.~\ref{fig:patch_viusalization}. Even when we directly face adversarial patches, it is hard for humans to distinguish the adversarial patches with a glance, let alone when adding them to the target image in real-world scenes. For instance, we show several images with the adversarial patches generated by GoogleAp, LaVan, and our \name in Fig.~\ref{fig:realpatch}. We can observe that the adversarial patches generated by \name are visually realistic to humans, which can disguise as scrawls in the real world. Thanks to their reality, the adversarial patches with the size of $0.3$ and perturbation budget of $16$ are not very large when added to the image. On the contrary, the adversarial patches generated by previous works (\eg, GoogleAp and LaVan) do not contain any meaningful information, making them easy to be detected by humans, although GoogleAp also initializes the adversarial patch with an actual image. Thus, we do not take these works as our baselines when evaluating the attack performance.

\begin{figure}
    \begin{minipage}{0.155\textwidth}
    \begin{subfigure}{\textwidth}
        \includegraphics[width=\linewidth]{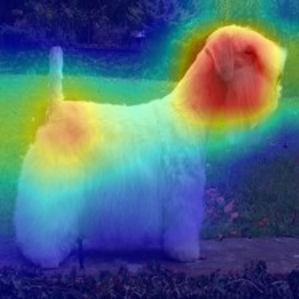}\\
        \vspace{-0.3cm}
        \includegraphics[width=\linewidth]{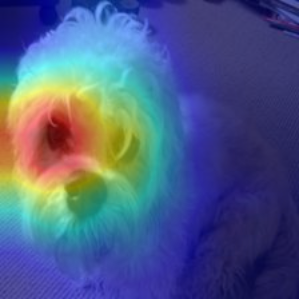}\\
        \vspace{-0.3cm}
        \includegraphics[width=\linewidth]{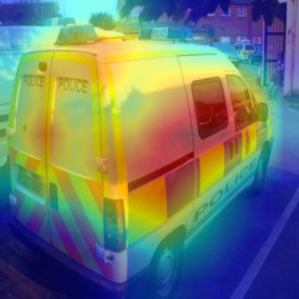}
        %\centering
        \caption*{Raw}
    \end{subfigure}
    \end{minipage}%
    \hspace{0.1em}
    \begin{minipage}{0.155\textwidth}
    \begin{subfigure}{\textwidth}
        \includegraphics[width=\linewidth]{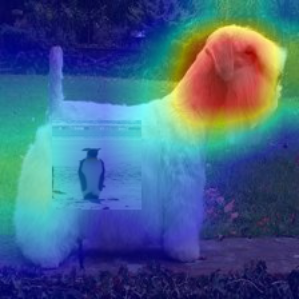}\\
        \vspace{-0.3cm}
        \includegraphics[width=\linewidth]{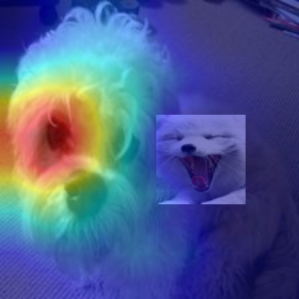}\\
        \vspace{-0.3cm}
        \includegraphics[width=\linewidth]{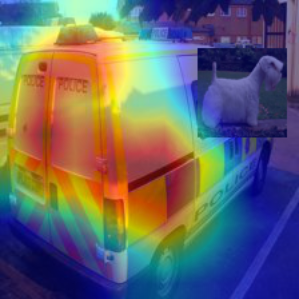}
        %\centering
        \caption*{Original patch}
    \end{subfigure}
    \end{minipage}%
    \hspace{0.1em}
    \begin{minipage}{0.155\textwidth}
        \begin{subfigure}{\textwidth}
        \includegraphics[width=\linewidth]{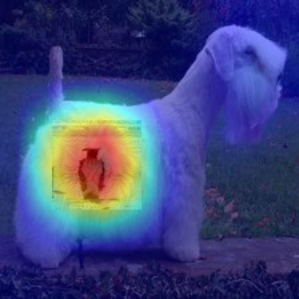}\\
        \vspace{-0.3cm}
        \includegraphics[width=\linewidth]{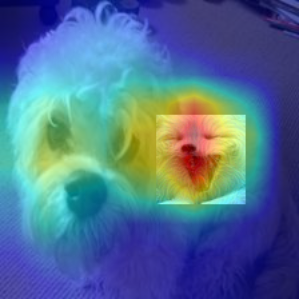}\\
        \vspace{-0.3cm}
       \includegraphics[width=\linewidth]{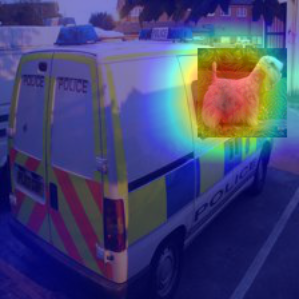}
        %\centering
        \caption*{Adversarial patch}
        \end{subfigure}
    \end{minipage}
    % \vspace{-0.4em}
    \caption{Attention heatmaps of raw image, the image with original patch and adversarial patches on ResNet-18. See more samples in the Appendix.}
    \label{fig:attention_heatmaps}
    \vspace{-1em}
\end{figure}

\label{sec:exp:physical_world}
The main advantage of adversarial patches is to be deployed in the physical world. \name makes it easier to place the adversarial patches and harder to be detected. Here we adopt ResNet-18 as the target model to conduct the evaluation in the physical world. We first take a photo of the real-world object, which can be correctly classified. Then we generate an adversarial patch using \name and print it by HITI digital photo printer P910L. Finally, we put the printed adversarial patch around the object and take a photo for recognition.

As shown in Fig.~\ref{fig:physical_world:raw}, we choose three everyday things in life, namely water bottle, toilet paper and mouse, which can be correctly classified. After placing the original image patch on the corresponding object illustrated in Fig.~\ref{fig:physical_world:original}, the model can still make the correct prediction, supporting that the deep models are robust to such a natural image patch. From Fig.~\ref{fig:physical_world:adversarial}, we can observe that there is no visual difference between the original patch and the adversarial patch in the physical world, which guarantees the visual reality of adversarial patches. When we put the adversarial patches in the same position of the original patch in Fig.~\ref{fig:physical_world:original}, the images with adversarial patches can successfully fool the victim model. Due to the reality of adversarial patches, they can easily disguise as logos or scrawls on the objects, making it hard to distinguish these adversarial patches and natural ones. Such superior visual reality and ease of deployment in various scenes support that \name can be widely used to attack real-world applications without being detected. The reality also makes the attack imperceptible, which reveals a new and significant threat to commercial applications. 

\begin{figure*}
    \begin{minipage}{0.23\textwidth}
    \begin{subfigure}{\textwidth}
        \includegraphics[width=\linewidth]{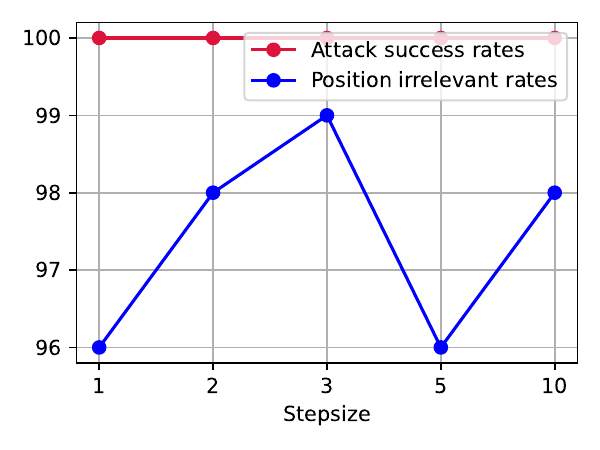}
        \caption{VGG-16}
    \end{subfigure}
    \end{minipage}%
    \hspace{1em}
    \begin{minipage}{0.23\textwidth}
    \begin{subfigure}{\textwidth}
        \includegraphics[width=\linewidth]{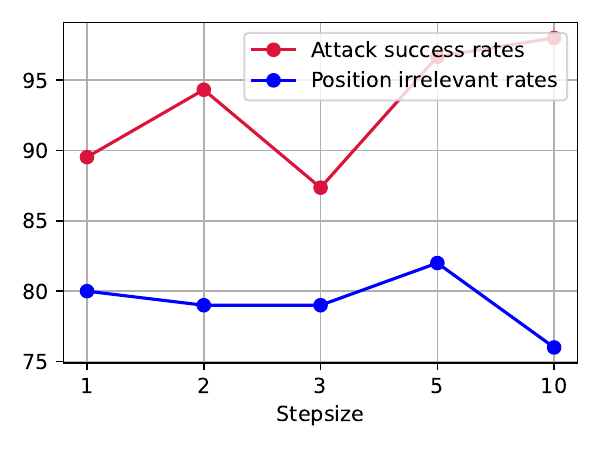}
        %\centering
        \caption{ResNet-18}
    \end{subfigure}
    \end{minipage}%
    \hspace{1em}
    \begin{minipage}{0.23\textwidth}
        \begin{subfigure}{\textwidth}
        \includegraphics[width=\linewidth]{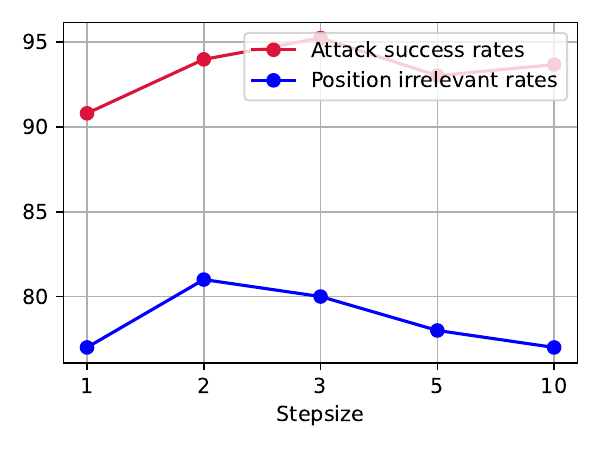}
        \caption{MobileNet-2}
        \end{subfigure}
    \end{minipage}%
    \hspace{1em}
    \begin{minipage}{0.23\textwidth}
        \begin{subfigure}{\textwidth}
        \includegraphics[width=\linewidth]{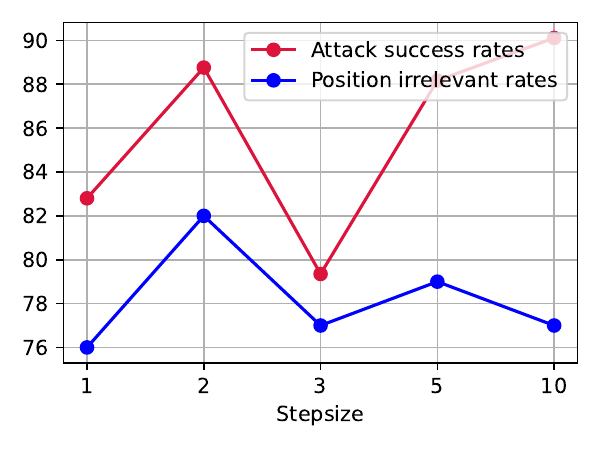}
        \caption{DenseNet-121}
        \end{subfigure}
    \end{minipage}
    %\vspace{-1em}
    \caption{Attack successful rates (\%) and position irrelevant rates (\%) of various models using \name on various stepsizes with the default search range $k=
    10$.}
    \label{fig:abl_interval}
   % \vspace{-1em}
\end{figure*}

% \begin{figure*}
%     \begin{subfigure}{0.23\textwidth}
%         \includegraphics[width=\linewidth]{figs/pdf/ablation/q/ab_q1.pdf}
%         \caption{VGG-16}
%     \end{subfigure}%
%     \begin{subfigure}{0.23\textwidth}
%         \includegraphics[width=\linewidth]{figs/pdf/ablation/q/ab_q2.pdf}
%         \caption{ResNet-18}
%     \end{subfigure}%
%     \begin{subfigure}{0.23\textwidth}
%         \includegraphics[width=\linewidth]{figs/pdf/ablation/q/ab_q3.pdf}
%         \caption{MobileNet-v2}
%     \end{subfigure}%
%     \begin{subfigure}{0.23\textwidth}
%         \includegraphics[width=\linewidth]{figs/pdf/ablation/q/ab_q4.pdf}
%         \caption{DenseNet}
%     \end{subfigure}
%     \label{fig:abl_length}
%     \caption{The attack successful rate (\%) and position irrelevant rate (\%) of various models using \name on various search length.}
% \end{figure*}

\begin{figure*}
    \begin{minipage}{0.23\textwidth}
    \begin{subfigure}{\textwidth}
        \includegraphics[width=\linewidth]{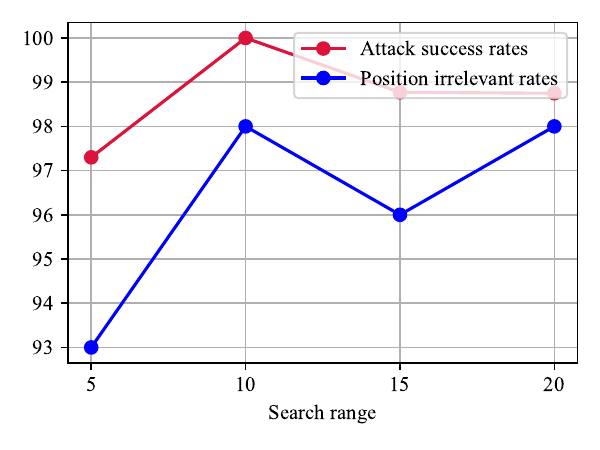}
        \caption{VGG-16}
    \end{subfigure}
    \end{minipage}%
    \hspace{1em}
    \begin{minipage}{0.23\textwidth}
        \begin{subfigure}{\textwidth}
        \includegraphics[width=\linewidth]{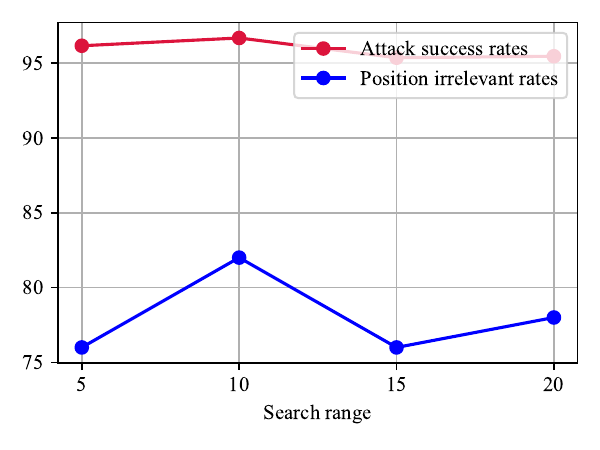}
        \caption{ResNet-18}
        \end{subfigure}
    \end{minipage}%
    \hspace{1em}
    \begin{minipage}{0.23\textwidth}
    \begin{subfigure}{\textwidth}
        \includegraphics[width=\linewidth]{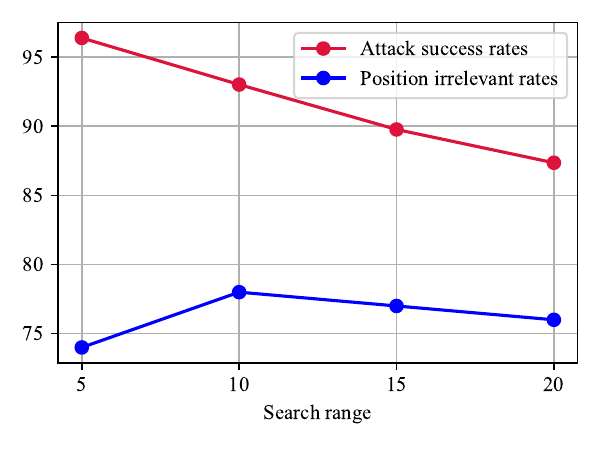}
        \caption{MobileNet-v2}
    \end{subfigure}
    \end{minipage}%
    \hspace{1em}
    \begin{minipage}{0.23\textwidth}
    \begin{subfigure}{\textwidth}
        \includegraphics[width=\linewidth]{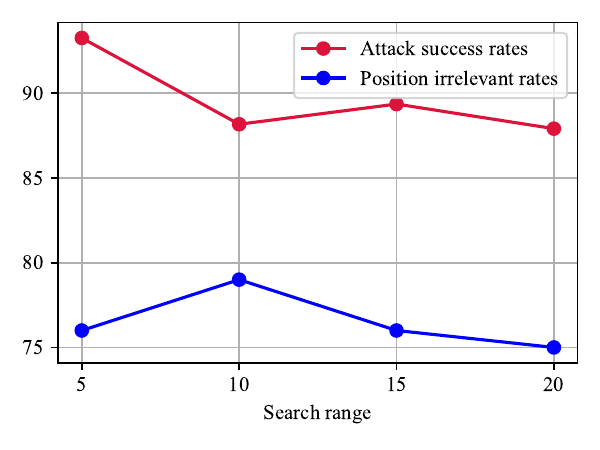}
        \caption{DenseNet-121}
    \end{subfigure}
    \end{minipage}
    \caption{Attack successful rates (\%) and position irrelevant rates (\%) of various models using \name on various search ranges with the default stepsize $\tau = 5$.}
    \label{fig:abl_range}
   \vspace{1em}
\end{figure*}

\subsection{Discussion}

Based on the above evaluations, we can conclude that \name can generate visually realistic adversarial patches in both digital and physical worlds to fool the deep models, which are hard for humans to perceive. As the first attack that focuses on generating visually realistic adversarial patches, we empirically investigate why such visually realistic patches can successfully fool the victim models. Since the deep models are not explainable, we adopt Grad-CAM, a widely adopted approach to calculate the attention heatmaps on the input image to highlight the significant features for recognition and interpret the model behavior on these images.

The attention heatmaps on the raw image, the image with the original patch and the adversarial patch at a random position are reported in Fig.~\ref{fig:attention_heatmaps}. As we can see, adding the original patch to the image does not change the attention heatmaps and cannot mislead the deep models, showing the stability and robustness of the model on these naturally transformed images. However, the attention heatmaps of the images with the adversarial patches are mainly located in the part of the added patch, which is significantly different from that on the original images. Hence, different from adversarial perturbation that makes the target model predict incorrectly, the perturbation in our adversarial patch misleads the victim model to focus on the wrong area. Without extracting the correct feature of the input image, the deep model cannot make the correct prediction. Interestingly, we also find that the prediction on these images is not consistent with the ground-truth label of the original patch, though the attention heatmaps are mainly on the patch. The reasons might be twofold: 1) The perturbation on the patch might also mislead the victim model, which is similar to adversarial perturbation. 2) The patch only occupies a small area of the image so that its surrounding image makes it hard for the deep model to recognize the patch. We will conduct more exploration and theoretical analysis to under why the visually realistic adversarial patches mislead the deep models in our future work.

\subsection{Parameter Studies}
We conduct a series of experiments to study the impact of two hyper-parameters to find the poorest position for optimizing the perturbation, namely the search interval and search range. 

\textbf{On the stepsize $\tau$.} The stepsize $\tau$ helps \name ignore several positions along the axis to boost the efficiency. We first investigate whether a large stepsize degrades the attack performance without checking each possible position. Specifically, we implement \name with various stepsizes on all four models and report the attack success rates and position irrelevant rates in Fig.~\ref{fig:abl_interval}. As we can see, when we adopt different stepsizes, the attack success rates and position irrelevant rates fluctuate in an acceptable range on all four models. This indicates that a large stepsize does not significantly decrease the attack performance but accelerates the search process. To balance the attack efficiency and effectiveness, we simply set $\tau=5$ in our experiments.

\textbf{On the search range $k$.} The search range $k$ constrains the size of the neighborhood to find a better position using local search. \name cannot find a suitable position with a small search range $k$, since it cannot effectively explore the neighborhood. With a large neighborhood, the approximation introduces a larger error when estimating the best position, which will also degrade the performance. To find a suitable search range $k$, we adopt various $k$ to conduct experiments on all four models. As shown in Fig.~\ref{fig:abl_range}, when $k=5$, \name achieves the lowest position irrelevant rates. When we increase $k$, the position irrelevant rates increase first but decrease slightly after $k>10$. Besides, the attack success rates are not significantly affected by various search ranges on these models. Hence, we adopt $k=10$ in our experiments.

In summary, the stepsize $\tau$ accelerates the local search without decreasing the attack performance, while a suitable search range results in better position irrelevant rates. In our experiments, we adopt $\tau=5$ and $k=10$ for better efficiency and effectiveness.
% In this section, we conduct ablation studies and parameter studies to investigate the effectiveness of our algorithm.

\section{Conclusion}

In this work, we first analyze and conclude that a high-quality adversarial patch should be \textit{realistic}, \textit{position irrelevant}, and \textit{printable} to be effectively deployed in the physical world. Based on this analysis, we propose the first visually realistic adversarial patch generating algorithm, denoted as \name. Specifically, \name iteratively optimizes the perturbation based on the real patch using a Total Variation loss and gamma transformation in the searched position with the minimum loss. Extensive evaluations on ImageNet dataset demonstrate that \name exhibits superior attack success rates and position irrelevant rates on various deep models in the digital world. Since the generated patches are realistic, position irrelevant, and printable, it is easy to deploy them in the physical world to fool deep models which are hard for humans to detect, raising a significant threat to DNNs-enabled applications. We hope our work can provide insights to generate these visually realistic patches for real-world attack, and draw more attention to mitigating the threat of such adversarial patches. 

{
    \small
    \bibliographystyle{ieeenat_fullname}
    \bibliography{main}

\begin{thebibliography}{67}
\providecommand{\natexlab}[1]{#1}
\providecommand{\url}[1]{\texttt{#1}}
\expandafter\ifx\csname urlstyle\endcsname\relax
  \providecommand{\doi}[1]{doi: #1}\else
  \providecommand{\doi}{doi: \begingroup \urlstyle{rm}\Url}\fi

\bibitem[Athalye et~al.(2018)Athalye, Carlini, and Wagner]{athalye2018obfuscated}
Anish Athalye, Nicholas Carlini, and David Wagner.
\newblock {Obfuscated Gradients Give a False Sense of Security: Circumventing Defenses to Adversarial Examples}.
\newblock pages 274--283, 2018.

\bibitem[Breier et~al.(2018)Breier, Hou, Jap, Ma, Bhasin, and Liu]{breier2018practical}
Jakub Breier, Xiaolu Hou, Dirmanto Jap, Lei Ma, Shivam Bhasin, and Yang Liu.
\newblock {DeepLaser: Practical Fault Attack on Deep Neural Networks}.
\newblock In \emph{Proceedings of the ACM SIGSAC Conference on Computer and Communications Security}, pages 2204--2206, 2018.

\bibitem[Brendel et~al.(2018)Brendel, Rauber, and Bethge]{brendel2018decision}
Wieland Brendel, Jonas Rauber, and Matthias Bethge.
\newblock {Decision-Based Adversarial Attacks: Reliable Attacks Against Black-Box Machine Learning Models}.
\newblock \emph{International Conference on Learning Representations}, 2018.

\bibitem[Brown et~al.(2017)Brown, Mané, Roy, Abadi, and Gilmer]{tom2017adversarial}
Tom~B. Brown, Dandelion Mané, Aurko Roy, Martín Abadi, and Justin Gilmer.
\newblock {Adversarial Patch}.
\newblock In \emph{Neural Information Processing Systems (Workshop)}, 2017.

\bibitem[Chen et~al.(2017)Chen, Liu, Li, Lu, and Song]{chen2017targeted}
Xinyun Chen, Chang Liu, Bo Li, Kimberly Lu, and Dawn Song.
\newblock {Targeted Backdoor Attacks on Deep Learning Systems Using Data Poisoning}.
\newblock In \emph{arXiv preprint arXiv:1712.05526}, 2017.

\bibitem[Cheng et~al.(2019{\natexlab{a}})Cheng, Le, Chen, Zhang, Yi, and Hsieh]{cheng2019query}
Minhao Cheng, Thong Le, Pin{-}Yu Chen, Huan Zhang, Jinfeng Yi, and Cho{-}Jui Hsieh.
\newblock {Query-Efficient Hard-label Black-box Attack: An Optimization-based Approach}.
\newblock In \emph{International Conference on Learning Representations}, 2019{\natexlab{a}}.

\bibitem[Cheng et~al.(2019{\natexlab{b}})Cheng, Dong, Pang, Su, and Zhu]{cheng2019improving}
Shuyu Cheng, Yinpeng Dong, Tianyu Pang, Hang Su, and Jun Zhu.
\newblock Improving black-box adversarial attacks with a transfer-based prior.
\newblock In \emph{Advances in Neural Information Processing Systems}, pages 10932--10942, 2019{\natexlab{b}}.

\bibitem[Cohen et~al.(2019)Cohen, Rosenfeld, and Kolter]{Cohen2019RS}
Jeremy~M Cohen, Elan Rosenfeld, and J~Zico Kolter.
\newblock {Certified adversarial robustness via randomized smoothing}.
\newblock In \emph{International Conference on Machine Learning}, 2019.

\bibitem[Croce and Hein(2020)]{croce2020reliable}
Francesco Croce and Matthias Hein.
\newblock {Reliable Evaluation of Adversarial Robustness with an Ensemble of Diverse Parameter-free Attacks}.
\newblock In \emph{International conference on machine learning}, pages 2206--2216, 2020.

\bibitem[Dong et~al.(2018)Dong, Liao, Pang, Su, Zhu, Hu, and Li]{dong2018boosting}
Yinpeng Dong, Fangzhou Liao, Tianyu Pang, Hang Su, Jun Zhu, Xiaolin Hu, and Jianguo Li.
\newblock {Boosting Adversarial Attacks with Momentum}.
\newblock In \emph{Proceedings of the IEEE Conference on Computer Vision and Pattern Recognition}, pages 9185--9193, 2018.

\bibitem[Duan et~al.(2020)Duan, Ma, Wang, Bailey, Qin, and Yang]{duan2020adversarial}
Ranjie Duan, Xingjun Ma, Yisen Wang, James Bailey, A~Kai Qin, and Yun Yang.
\newblock {Adversarial Camouflage: Hiding Physical-World Attacks With Natural Styles}.
\newblock In \emph{Proceedings of the IEEE/CVF conference on computer vision and pattern recognition}, pages 1000--1008, 2020.

\bibitem[Eykholt et~al.(2018)Eykholt, Evtimov, Fernandes, Li, Rahmati, Xiao, Prakash, Kohno, and Song]{eykholt2018robust}
Kevin Eykholt, Ivan Evtimov, Earlence Fernandes, Bo Li, Amir Rahmati, Chaowei Xiao, Atul Prakash, Tadayoshi Kohno, and Dawn Song.
\newblock {Robust Physical-World Attacks on Deep Learning Visual Classification}.
\newblock In \emph{Proceedings of the IEEE Conference on Computer Vision and Pattern Recognition}, pages 1625--1634, 2018.

\bibitem[Goodfellow et~al.(2014)Goodfellow, Pouget-Abadie, Mirza, Xu, Warde-Farley, Ozair, Courville, and Bengio]{goodfellow2014generative}
Ian Goodfellow, Jean Pouget-Abadie, Mehdi Mirza, Bing Xu, David Warde-Farley, Sherjil Ozair, Aaron Courville, and Yoshua Bengio.
\newblock {Generative Adversarial Nets}.
\newblock In \emph{Advances in Neural Information Processing Systems}, pages 2672--2680, 2014.

\bibitem[Goodfellow et~al.(2015)Goodfellow, Shlens, and Szegedy]{goodfellow2015FGSM}
Ian~J Goodfellow, Jonathon Shlens, and Christian Szegedy.
\newblock {Explaining and harnessing adversarial examples}.
\newblock In \emph{International Conference on Learning Representations}, 2015.

\bibitem[Guo et~al.(2019)Guo, Gardner, You, Wilson, and Weinberger]{guo2019simple}
Chuan Guo, Jacob Gardner, Yurong You, Andrew~Gordon Wilson, and Kilian Weinberger.
\newblock {Simple Black-box Adversarial Attacks}.
\newblock In \emph{International Conference on Machine Learning}, pages 2484--2493, 2019.

\bibitem[Hayes(2018)]{hayes2018on}
Jamie Hayes.
\newblock {On Visible Adversarial Perturbations \& Digital Watermarking}.
\newblock In \emph{Proceedings of the IEEE Conference on Computer Vision and Pattern Recognition (Workshop)}, pages 1597--1604, 2018.

\bibitem[He et~al.(2016)He, Zhang, Ren, and Sun]{he2016resnet}
Kaiming He, Xiangyu Zhang, Shaoqing Ren, and Jian Sun.
\newblock {Deep Residual Learning for Image Recognition}.
\newblock In \emph{Proceedings of the IEEE Conference on Computer Vision and Pattern Recognition}, pages 770--778, 2016.

\bibitem[Hu et~al.(2021)Hu, Kung, Tan, Chen, Hua, and Cheng]{hu2021naturalistic}
Yu-Chih-Tuan Hu, Bo-Han Kung, Daniel~Stanley Tan, Jun-Cheng Chen, Kai-Lung Hua, and Wen-Huang Cheng.
\newblock {Naturalistic Physical Adversarial Patch for Object Detectors}.
\newblock In \emph{Proceedings of the IEEE/CVF International Conference on Computer Vision}, pages 7848--7857, 2021.

\bibitem[Hu et~al.(2022)Hu, Huang, Zhu, Sun, Zhang, and Hu]{hu2022adversarial}
Zhanhao Hu, Siyuan Huang, Xiaopei Zhu, Fuchun Sun, Bo Zhang, and Xiaolin Hu.
\newblock {Adversarial Texture for Fooling Person Detectors in the Physical World}.
\newblock In \emph{Proceedings of the IEEE/CVF Conference on Computer Vision and Pattern Recognition}, pages 13307--13316, 2022.

\bibitem[Huang et~al.(2017)Huang, Liu, Van Der~Maaten, and Weinberger]{huang2017densely}
Gao Huang, Zhuang Liu, Laurens Van Der~Maaten, and Kilian~Q Weinberger.
\newblock {Densely connected convolutional networks}.
\newblock In \emph{Proceedings of the IEEE Conference on Computer Vision and Pattern Recognition}, pages 4700--4708, 2017.

\bibitem[Huang et~al.(2020)Huang, Gao, Zhou, Xie, Yuille, Zou, and Liu]{huang2020universal}
Lifeng Huang, Chengying Gao, Yuyin Zhou, Cihang Xie, Alan~L Yuille, Changqing Zou, and Ning Liu.
\newblock {Universal Physical Camouflage Attacks on Object Detectors}.
\newblock In \emph{Proceedings of the IEEE Conference on Computer Vision and Pattern Recognition}, pages 720--729, 2020.

\bibitem[Ilyas et~al.(2018)Ilyas, Engstrom, Athalye, and Lin]{ilyas2018black}
Andrew Ilyas, Logan Engstrom, Anish Athalye, and Jessy Lin.
\newblock {Black-box Adversarial Attacks with Limited Queries and Information}.
\newblock In \emph{International Conference on Machine Learning}, pages 2142--2151, 2018.

\bibitem[Karen and Andrew(2015)]{Simonyan2015VGG}
Simonyan Karen and Zisserman Andrew.
\newblock {Very Deep Convolutional Networks for Large-Scale Image Recognition}.
\newblock In \emph{International Conference on Learning Representations}, 2015.

\bibitem[Karmon et~al.(2018)Karmon, Zoran, and Goldberg]{Karmon2018lavan}
Danny Karmon, Daniel Zoran, and Yoav Goldberg.
\newblock {LaVAN: Localized and Visible Adversarial Noise}.
\newblock In \emph{International Conference on Machine Learning}, 2018.

\bibitem[Komkov and Petiushko(2021)]{komkov2021advhat}
Stepan Komkov and Aleksandr Petiushko.
\newblock {AdvHat: Real-world adversarial attack on ArcFace Face ID system}.
\newblock In \emph{International Conference on Pattern Recognition}, pages 819--826, 2021.

\bibitem[Krizhevsky et~al.(2012)Krizhevsky, Sutskever, and Hinton]{alex2012ImageNet}
Alex Krizhevsky, Ilya Sutskever, and Geoffrey~E. Hinton.
\newblock {ImageNet Classification with Deep Convolutional Neural Networks}.
\newblock In \emph{Neural Information Processing Systems}, pages 1106--1114, 2012.

\bibitem[Kurakin et~al.(2017)Kurakin, Goodfellow, and Bengio]{kurakin2017adversarial}
Alexey Kurakin, Ian Goodfellow, and Samy Bengio.
\newblock {Adversarial Examples in the Physical World}.
\newblock In \emph{International Conference on Learning Representations (Workshop)}, 2017.

\bibitem[Li et~al.(2020)Li, Xu, Zhang, Yang, and Li]{li2020qeba}
Huichen Li, Xiaojun Xu, Xiaolu Zhang, Shuang Yang, and Bo Li.
\newblock {QEBA: Query-Efficient Boundary-Based Blackbox Attack}.
\newblock In \emph{Proceedings of the IEEE Conference on Computer Vision and Pattern Recognition}, pages 1218--1227, 2020.

\bibitem[Li and Ji(2021)]{li2021generative}
Xiang Li and Shihao Ji.
\newblock {Generative Dynamic Patch Attack}.
\newblock 2021.

\bibitem[Li et~al.(2019)Li, Li, Wang, Zhang, and Gong]{li2019nattack}
Yandong Li, Lijun Li, Liqiang Wang, Tong Zhang, and Boqing Gong.
\newblock {NATTACK: Learning the Distributions of Adversarial Examples for an Improved Black-Box Attack on Deep Neural Networks}.
\newblock In \emph{International Conference on Machine Learning}, pages 3866--3876, 2019.

\bibitem[Li et~al.(2022)Li, Jiang, Li, and Xia]{li2022backdoor}
Yiming Li, Yong Jiang, Zhifeng Li, and Shutao Xia.
\newblock {Backdoor Learning: A Survey}.
\newblock In \emph{IEEE Transactions on Neural Networks and Learning Systems}, 2022.

\bibitem[Liu et~al.(2019)Liu, Liu, Fan, Ma, Zhang, Xie, and Tao]{liu2019perceptual}
Aishan Liu, Xianglong Liu, Jiaxin Fan, Yuqing Ma, Anlan Zhang, Huiyuan Xie, and Dacheng Tao.
\newblock {Perceptual-Sensitive GAN for Generating Adversarial Patches}.
\newblock In \emph{Proceedings of the AAAI Conference on Artificial Intelligence}, 2019.

\bibitem[Long et~al.(2015)Long, Shelhamer, and Darrell]{jonathan2015fully}
Jonathan Long, Evan Shelhamer, and Trevor Darrell.
\newblock {Fully Convolutional Networks for Semantic Segmentation}.
\newblock In \emph{Proceedings of the IEEE Conference on Computer Vision and Pattern Recognition}, pages 3431--3440, 2015.

\bibitem[Madry et~al.(2018)Madry, Makelov, Schmidt, Tsipras, and Vladu]{madry2018pgd}
Aleksander Madry, Aleksandar Makelov, Ludwig Schmidt, Dimitris Tsipras, and Adrian Vladu.
\newblock {Towards Deep Learning Models Resistant to Adversarial Attacks}.
\newblock In \emph{International Conference on Learning Representations}, 2018.

\bibitem[Mark et~al.(2018)Mark, Andrew, Menglong, Andrey, and Liang-Chieh]{Mark2018mobile}
Sandler Mark, Howard Andrew, Zhu Menglong, Zhmoginov Andrey, and Chen Liang-Chieh.
\newblock {Mobilenetv2: Inverted residuals and linear bottlenecks}.
\newblock In \emph{Proceedings of the IEEE Conference on Computer Vision and Pattern Recognition}, page 4510–4520, 2018.

\bibitem[Moosavi-Dezfooli et~al.(2016)Moosavi-Dezfooli, Fawzi, and Frossard]{moosavi2016deepfool}
Seyed-Mohsen Moosavi-Dezfooli, Alhussein Fawzi, and Pascal Frossard.
\newblock {Deepfool: A Simple and Accurate Method to Fool Deep Neural Networks}.
\newblock In \emph{Proceedings of the IEEE/CVF Conference on Computer Vision and Pattern Recognition}, pages 2574--2582, 2016.

\bibitem[Naseer et~al.(2019)Naseer, Khan, and Porikli]{naseer2019local}
Muzammal Naseer, Salman~H. Khan, and Fatih Porikli.
\newblock {Local Gradients Smoothing: Defense against localized adversarial attacks}.
\newblock In \emph{WACV. IEEE}, pages 1300--1307, 2019.

\bibitem[Nesti et~al.(2022)Nesti, Rossolini, Nair, Biondi, and Buttazzo]{nesti2022evaluating}
Federico Nesti, Giulio Rossolini, Saasha Nair, Alessandro Biondi, and Giorgio Buttazzo.
\newblock {Evaluating the Robustness of Semantic Segmentation for Autonomous Driving against Real-World Adversarial Patch Attacks}.
\newblock In \emph{Proceedings of the IEEE/CVF Winter Conference on Applications of Computer Vision}, pages 2280--2289, 2022.

\bibitem[Olga et~al.(2015)Olga, Jia, Hao, Jonathan, Sanjeev, Sean, hUANG Zhiheng, Andrej, Adiya, and Bernstein~Michael]{Russa2015imagenet}
Russakovsky Olga, Deng Jia, Su Hao, Krause Jonathan, Satheesh Sanjeev, Ma Sean, hUANG Zhiheng, Karpathy Andrej, Khosla Adiya, and et~al Bernstein~Michael.
\newblock {Imagenet large scale visual recognition challenge}.
\newblock In \emph{International journal of computer vision}, pages 211--252, 2015.

\bibitem[Rakin et~al.(2020)Rakin, He, and Fan]{rakin2020tbt}
Adnan~Siraj Rakin, Zhezhi He, and Deliang Fan.
\newblock {TBT: Targeted Neural Network Attack With Bit Trojan}.
\newblock In \emph{Proceedings of the IEEE/CVF Conference on Computer Vision and Pattern Recognition}, pages 13198--13207, 2020.

\bibitem[Redmon et~al.(2016)Redmon, Divvala, Girshick, and Farhadi]{joseph2016you}
Joseph Redmon, Santosh Divvala, Ross Girshick, and Ali Farhadi.
\newblock {You Only Look Once: Unified, Real-Time Object Detection}.
\newblock In \emph{Proceedings of the IEEE Conference on Computer Vision and Pattern Recognition}, pages 779--788, 2016.

\bibitem[Ren et~al.(2015)Ren, He, Girshick, and Sun]{ren2015faster}
Shaoqing Ren, Kaiming He, Ross Girshick, and Jian Sun.
\newblock {Faster R-CNN: Towards Real-Time Object Detection with Region Proposal Networks}.
\newblock In \emph{Neural Information Processing Systems}, pages 91--99, 2015.

\bibitem[Ronneberger et~al.(2015)Ronneberger, Fischer, and Brox]{olaf2015unet}
Olaf Ronneberger, Philipp Fischer, and Thomas Brox.
\newblock {U-Net: Convolutional Networks for Biomedical Image Segmentation}.
\newblock In \emph{Medical Image Computing and Computer-Assisted Intervention}, pages 234--241, 2015.

\bibitem[Sharif et~al.(2016)Sharif, Bhagavatula, Bauer, and Reiter]{sharif2016accessorize}
Mahmood Sharif, Sruti Bhagavatula, Lujo Bauer, and Michael~K. Reiter.
\newblock {Accessorize to a Crime: Real and Stealthy Attacks on State-of-the-Art Face Recognition}.
\newblock In \emph{Proceedings of the ACM SIGSAC Conference on Computer and Communications Security}, pages 1528--1540, 2016.

\bibitem[Sharma et~al.(2022)Sharma, Bian, Munz, and Narayan]{sharma2022adversarial}
Abhijith Sharma, Yijun Bian, Phil Munz, and Apurva Narayan.
\newblock {Adversarial Patch Attacks and Defences in Vision-Based Tasks: A Survey}.
\newblock In \emph{arXiv preprint arXiv:2206.08304}, 2022.

\bibitem[Szegedy et~al.(2014)Szegedy, Zaremba, Sutskever, Bruna, Erhan, Goodfellow, and Fergus]{szegedy2014intriguing}
Christian Szegedy, Wojciech Zaremba, Ilya Sutskever, Joan Bruna, Dumitru Erhan, Ian Goodfellow, and Rob Fergus.
\newblock {Intriguing properties of neural networks}.
\newblock In \emph{International Conference on Learning Representations}, 2014.

\bibitem[Tram{\`e}r et~al.(2018)Tram{\`e}r, Kurakin, Papernot, Goodfellow, Boneh, and McDaniel]{tramer2018ensemble}
Florian Tram{\`e}r, Alexey Kurakin, Nicolas Papernot, Ian Goodfellow, Dan Boneh, and Patrick McDaniel.
\newblock {Ensemble Adversarial Training: Attacks and Defenses}.
\newblock \emph{International Conference on Learning Representations}, 2018.

\bibitem[Wang et~al.(2023{\natexlab{a}})Wang, He, Wang, and Wang]{wang2023boosting}
Kunyu Wang, Xuanran He, Wenxuan Wang, and Xiaosen Wang.
\newblock {Boosting Adversarial Transferability by Block Shuffle and Rotation}.
\newblock \emph{arXiv preprint arXiv:2308.10299}, 2023{\natexlab{a}}.

\bibitem[Wang and He(2021)]{wang2021enhancing}
Xiaosen Wang and Kun He.
\newblock Enhancing the transferability of adversarial attacks through variance tuning.
\newblock In \emph{Proceedings of the IEEE Conference on Computer Vision and Pattern Recognition}, 2021.

\bibitem[Wang et~al.(2019)Wang, He, Song, Wang, and Hopcroft]{wang2019atgan}
Xiaosen Wang, Kun He, Chuanbiao Song, Liwei Wang, and John~E. Hopcroft.
\newblock {AT-GAN: A Generative Attack Model for Adversarial Transferring on Generative Adversarial Nets}.
\newblock \emph{arXiv preprint arXiv:1904.07793}, 2019.

\bibitem[Wang et~al.(2021{\natexlab{a}})Wang, He, Wang, and He]{wang2021admix}
Xiaosen Wang, Xuanran He, Jingdong Wang, and Kun He.
\newblock Admix: Enhancing the transferability of adversarial attacks.
\newblock In \emph{International Conference on Computer Vision}, pages 16138--16147, 2021{\natexlab{a}}.

\bibitem[Wang et~al.(2021{\natexlab{b}})Wang, Lin, Hu, Wang, and He]{wang2021boosting}
Xiaosen Wang, Jiadong Lin, Han Hu, Jingdong Wang, and Kun He.
\newblock {Boosting Adversarial Transferability through Enhanced Momentum}.
\newblock In \emph{British Machine Vision Conference}, 2021{\natexlab{b}}.

\bibitem[Wang et~al.(2022)Wang, Zhang, Tong, Gong, He, Li, and Liu]{wang2022triangle}
Xiaosen Wang, Zeliang Zhang, Kangheng Tong, Dihong Gong, Kun He, Zhifeng Li, and Wei Liu.
\newblock {Triangle Attack: A Query-efficient Decision-based Adversarial Attack}.
\newblock In \emph{European conference on computer vision}, 2022.

\bibitem[Wang et~al.(2023{\natexlab{b}})Wang, Tong, and He]{wang2023rethinkinga}
Xiaosen Wang, Kangheng Tong, and Kun He.
\newblock {Rethinking the Backward Propagation for Adversarial Transferability}.
\newblock In \emph{Proceedings of the Advances in Neural Information Processing Systems}, 2023{\natexlab{b}}.

\bibitem[Wang et~al.(2023{\natexlab{c}})Wang, Zhang, and Zhang]{wang2023structure}
Xiaosen Wang, Zeliang Zhang, and Jianping Zhang.
\newblock {Structure Invariant Transformation for better Adversarial Transferability}.
\newblock In \emph{Proceedings of the IEEE/CVF International Conference on Computer Vision}, pages 4607--4619, 2023{\natexlab{c}}.

\bibitem[Wei et~al.(2022)Wei, Guo, and Yu]{wei2022adversarial}
Xingxing Wei, Ying Guo, and Jie Yu.
\newblock {Adversarial Sticker: A Stealthy Attack Method in the Physical World}.
\newblock In \emph{IEEE Transactions on Pattern Analysis and Machine Intelligence}, 2022.

\bibitem[Wong et~al.(2020)Wong, Rice, and Kolter]{wong2020fast}
Eric Wong, Leslie Rice, and J.Zico Kolter.
\newblock {Fast is better than free: Revisiting adversarial training}.
\newblock In \emph{International Conference on Learning Representations}, 2020.

\bibitem[Xiang et~al.(2021)Xiang, Bhagoji, Sehwag, and Mittal]{xiang2021patchguard}
Chong Xiang, Arjun~Nitin Bhagoji, Vikash Sehwag, and Prateek Mittal.
\newblock Patchguard: A provably robust defense against adversarial patches via small receptive fields and masking.
\newblock In \emph{30th {USENIX} Security Symposium ({USENIX} Security)}, 2021.

\bibitem[Xiao et~al.(2021)Xiao, Gao, Fu, Dong, Gao, Zhang, Zhou, and Zhu]{xiao2021improving}
Zihao Xiao, Xianfeng Gao, Chilin Fu, Yinpeng Dong, Wei Gao, Xiaolu Zhang, Jun Zhou, and Jun Zhu.
\newblock {Improving Transferability of Adversarial Patches on Face Recognition with Generative Models}.
\newblock In \emph{Proceedings of the IEEE Conference on Computer Vision and Pattern Recognition}, pages 11845--11854, 2021.

\bibitem[Xie et~al.(2019)Xie, Zhang, Zhou, Bai, Wang, Ren, and Yuille]{xie2019improving}
Cihang Xie, Zhishuai Zhang, Yuyin Zhou, Song Bai, Jianyu Wang, Zhou Ren, and Alan~L Yuille.
\newblock {Improving Transferability of Adversarial Examples with Input Diversity}.
\newblock In \emph{Proceedings of the IEEE Conference on Computer Vision and Pattern Recognition}, pages 2730--2739, 2019.

\bibitem[Xu et~al.(2020)Xu, Zhang, Liu, Fan, Sun, Chen, Chen, Wang, and Lin]{xu2020adversarial}
Kaidi Xu, Gaoyuan Zhang, Sijia Liu, Quanfu Fan, Mengshu Sun, Hongge Chen, Pin-Yu Chen, Yanzhi Wang, and Xue Lin.
\newblock {Adversarial T-shirt! Evading Person Detectors in A Physical World}.
\newblock In \emph{European conference on computer vision}, pages 665--681, 2020.

\bibitem[Yakura et~al.(2020)Yakura, Akimoto, and Sakuma]{yakura2020generate}
Hiromu Yakura, Youhei Akimoto, and Jun Sakuma.
\newblock {Generate (non-software) Bugs to Fool Classifiers}.
\newblock In \emph{Proceedings of the AAAI Conference on Artificial Intelligence}, pages 1070--1078, 2020.

\bibitem[Yang et~al.(2020)Yang, Kortylewski, Xie, Cao, and Yuille]{yang2020patchattack}
Chenglin Yang, Adam Kortylewski, Cihang Xie, Yinzhi Cao, and Alan Yuille.
\newblock {PatchAttack: A Black-box Texture-based Attack with Reinforcement Learning}.
\newblock In \emph{European Conference on Computer Vision}, pages 681--698, 2020.

\bibitem[Yang et~al.(2022)Yang, Wang, and He]{yang2022robust}
Yichen Yang, Xiaosen Wang, and Kun He.
\newblock {Robust Textual Embedding against Word-level Adversarial Attacks}.
\newblock In \emph{Proceedings of the Conference on Uncertainty in Artificial Intelligence}, page 2214–2224, 2022.

\bibitem[Yu et~al.(2022)Yu, Wang, Che, and He]{yu2022texthacker}
Zhen Yu, Xiaosen Wang, Wanxiang Che, and Kun He.
\newblock {TextHacker: Learning based Hybrid Local Search Algorithm for Text Hard-label Adversarial Attack}.
\newblock In \emph{Proceedings of the Conference on Empirical Methods in Natural Language Processing (Findings)}, page 622–637, 2022.

\bibitem[Zhang et~al.(2020)Zhang, Yuan, McCoyd, and Wagner]{zhang2020clipped}
Zhanyuan Zhang, Benson Yuan, Michael McCoyd, and David Wagner.
\newblock Clipped bagnet: Defending against sticker attacks with clipped bag-of-features.
\newblock In \emph{2020 IEEE Security and Privacy Workshops (SPW)}, pages 55--61, 2020.

\bibitem[Zolfi et~al.(2021)Zolfi, Kravchik, Elovici, and Shabtai]{zolfi2021translucent}
Alon Zolfi, Moshe Kravchik, Yuval Elovici, and Asaf Shabtai.
\newblock {The Translucent Patch: A Physical and Universal Attack on Object Detectors}.
\newblock In \emph{Proceedings of the IEEE/CVF Conference on Computer Vision and Pattern Recognition}, pages 15232--15241, 2021.

\end{thebibliography}
}
\clearpage
\newpage
\section{Appendix}
\subsection{More Samples of Adversarial Patches}
\label{app:adv_patch}
\begin{figure}[b]
    \centering
    \begin{minipage}[c]{0.15\textwidth}
        % \includegraphics[width=\linewidth]{figs/RealisticPatch/GoogleAp.png}\\
        % \vspace{-0.3cm}
        % \includegraphics[width=\linewidth]{figs/RealisticPatch/GoogleAp1.png}\\
       % \vspace{-0.3cm}
       \begin{subfigure}{\textwidth}
        \includegraphics[width=\linewidth]{figs/RealisticPatch/GoogleAp.png}\\
        \vspace{-0.3cm}
        \includegraphics[width=\linewidth]{figs/RealisticPatch/GoogleAp2.png}\\
       \vspace{-0.3cm}
        \includegraphics[width=\linewidth]{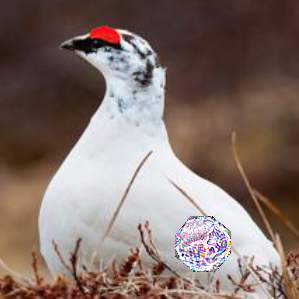}\\
       \vspace{-0.3cm}
        \includegraphics[width=\linewidth]{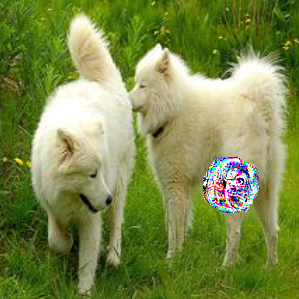}\\
       \vspace{-0.3cm}
        \includegraphics[width=\linewidth]{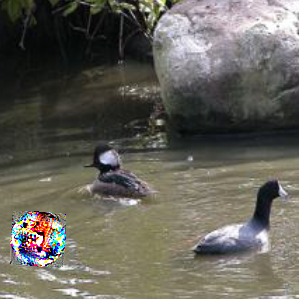}\\
       \vspace{-0.3cm}
        \includegraphics[width=\linewidth]{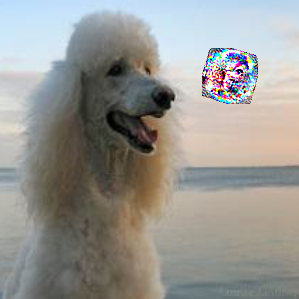}\\
       \vspace{-0.3cm}
        \caption*{GoogleAp}
        \end{subfigure}
    \end{minipage}
    %\hspace{0m}
    \begin{minipage}[c]{0.15\textwidth}
        \begin{subfigure}{\textwidth}
        \includegraphics[width=\linewidth]{figs/RealisticPatch/LAVAN.png}\\
        \vspace{-0.3cm}
        \includegraphics[width=\linewidth]{figs/RealisticPatch/LAVAN2.png}\\
        \vspace{-0.3cm}
        \includegraphics[width=\linewidth]{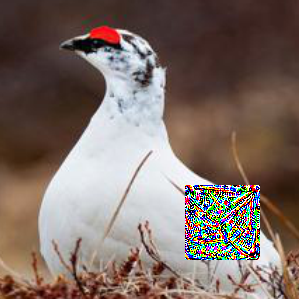}\\
        \vspace{-0.3cm}
        \includegraphics[width=\linewidth]{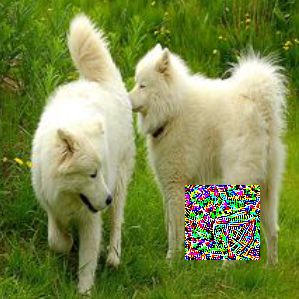}\\
        \vspace{-0.3cm}
        \includegraphics[width=\linewidth]{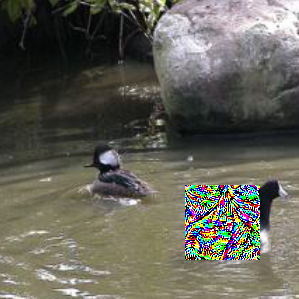}\\
        \vspace{-0.3cm}
        \includegraphics[width=\linewidth]{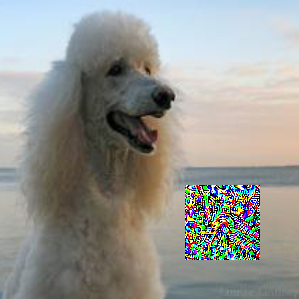}\\
       \vspace{-0.3cm}
        \caption*{LaVAN}
        %\vspace{-0.3cm}
        \end{subfigure}
    \end{minipage}%
    \hspace{0.01em}
    \begin{minipage}[c]{0.15\textwidth}
        \begin{subfigure}{\textwidth}
        \includegraphics[width=\linewidth]{figs/RealisticPatch/VRAP.png}\\
        \vspace{-0.3cm}
        \includegraphics[width=\linewidth]{figs/RealisticPatch/VRAP2.png}\\
        \vspace{-0.3cm}
        \includegraphics[width=\linewidth]{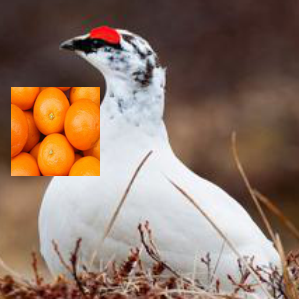}\\
        \vspace{-0.3cm}
        \includegraphics[width=\linewidth]{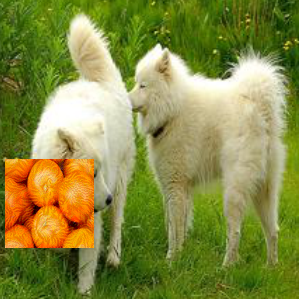}\\
        \vspace{-0.3cm}
        \includegraphics[width=\linewidth]{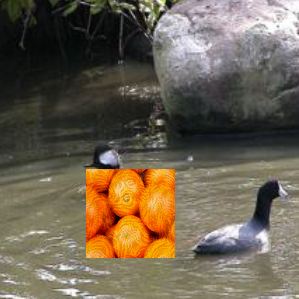}\\
        \vspace{-0.3cm}
        \includegraphics[width=\linewidth]{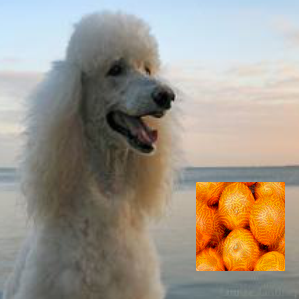}\\
        \vspace{-0.3cm}
        \caption*{\name}
        \end{subfigure}
    \end{minipage}%
    % \vspace{-0.5em}
    \caption{The adversarial patches \wrt the same input image generated by GoogleAp, LAVAN and our purposed \name, in which all the images are from ImageNet dataset.}
    \label{fig:appendix_A}
    % \vspace{-0.5em}
\end{figure}

In this section, we provide more adversarial patches generated by GoogleAp, LaVAN, and \name in Fig.~\ref{fig:appendix_A}. As we can see, the adversarial patches generated by \name are much more realistic than GoogleAp and LaVAN and can act as real scrawls or logos, which validates our motivation.
% the adversarial patches generated by GoogleAp, and LaVAN are not realistic and can easily regard as scrawl, while adversarial patches generated by \name remain realistic. These examples further validate our motivation.

\begin{figure}[tb]
    \centering
    \begin{minipage}[c]{0.15\textwidth}
       \begin{subfigure}{\textwidth}
       \includegraphics[width=\linewidth]{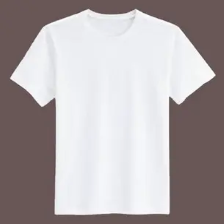}\\
        \vspace{-0.3cm}
        \includegraphics[width=\linewidth]{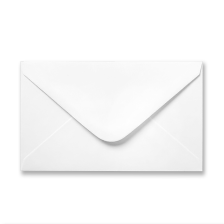}\\
       \vspace{-0.3cm}
        \includegraphics[width=\linewidth]{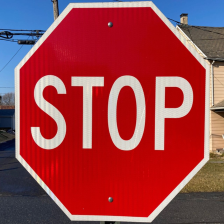}\\
       \vspace{-0.3cm}
        \includegraphics[width=\linewidth]{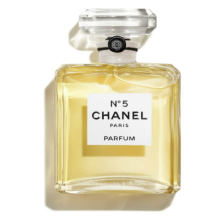}\\
       \vspace{-0.3cm}
        \includegraphics[width=\linewidth]{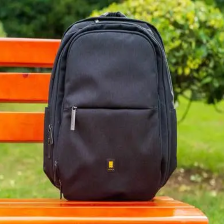}\\
       \vspace{-0.3cm}
        \includegraphics[width=\linewidth]{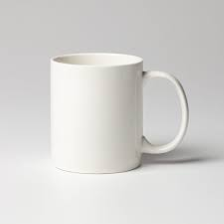}\\
       \vspace{-0.3cm}
        \caption{Raw}
        \end{subfigure}
    \end{minipage}
    \begin{minipage}[c]{0.15\textwidth}
        \begin{subfigure}{\textwidth}
        \includegraphics[width=\linewidth]{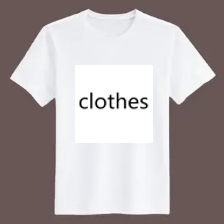}\\
        \vspace{-0.3cm}
        \includegraphics[width=\linewidth]{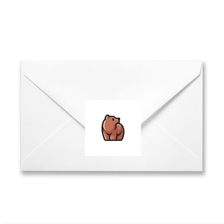}\\
        \vspace{-0.3cm}
        \includegraphics[width=\linewidth]{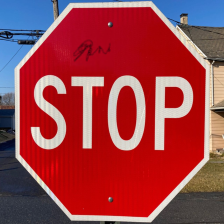}\\
        \vspace{-0.3cm}
        \includegraphics[width=\linewidth]{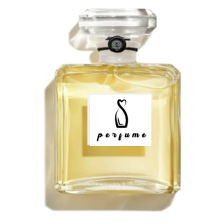}\\
        \vspace{-0.3cm}
        \includegraphics[width=\linewidth]{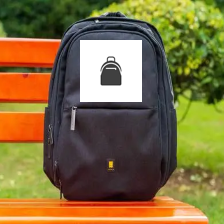}\\
        \vspace{-0.3cm}
        \includegraphics[width=\linewidth]{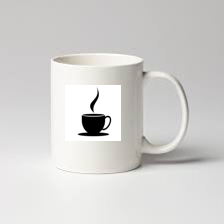}\\
        \vspace{-0.3cm}
        \caption{Original}
        \label{fig:appendix_b_o}
        \end{subfigure}
    \end{minipage}%
    \hspace{0.01em}
    \begin{minipage}[c]{0.15\textwidth}
        \begin{subfigure}{\textwidth}
        \includegraphics[width=\linewidth]{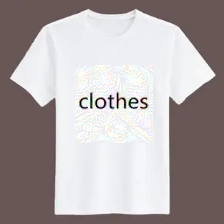}\\
        \vspace{-0.3cm}
        \includegraphics[width=\linewidth]{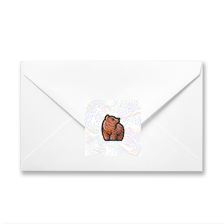}\\
        \vspace{-0.3cm}
        \includegraphics[width=\linewidth]{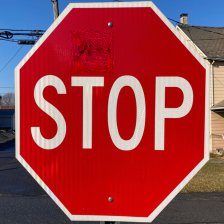}\\
        \vspace{-0.3cm}
        \includegraphics[width=\linewidth]{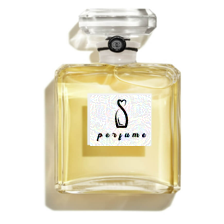}\\
        \vspace{-0.3cm}
        \includegraphics[width=\linewidth]{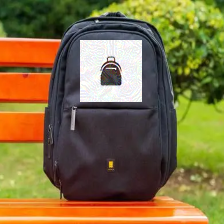}\\
        \vspace{-0.3cm}
        \includegraphics[width=\linewidth]{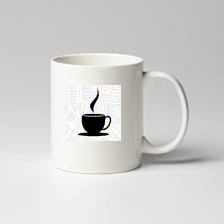}\\
        \vspace{-0.3cm}
        \caption{Adversarial}
        \label{fig:appendix_b_a}
        \end{subfigure}
    \end{minipage}%
    \caption{The raw images, images with original image patches and adversarial patches generated by \name on ResNet-18}
    \label{fig:Appendix_B}
    % \vspace{-0.5em}
\end{figure}

\subsection{More Samples of Physical Scrawls, Logos, and Adversarial Patches Crafted by \name}
\label{app:phy_scrawl}
In this section, we provide more adversarial patches generated by \name that can be regarded as scrawls or logos in the real world. As shown in Fig~\ref{fig:Appendix_B}, we craft the adversarial patches on ResNet-18 and place the adversarial patches in a natural position of victim images. In Fig.~\ref{fig:appendix_b_o}, the original patch does not mislead ResNet-18. However, after adding natural adversarial patches in Fig~\ref{fig:appendix_b_a}, ResNet-18 makes wrong decisions. The visually realistic adversarial patches validate that \name can be regarded as logos or scrawls in the real world and bring significant threats to deep models.

\begin{figure}[b]
    \centering
    \begin{minipage}[c]{0.15\textwidth}
        % \includegraphics[width=\linewidth]{figs/RealisticPatch/GoogleAp.png}\\
        % \vspace{-0.3cm}
        % \includegraphics[width=\linewidth]{figs/RealisticPatch/GoogleAp1.png}\\
       % \vspace{-0.3cm}
       \begin{subfigure}{\textwidth}
        \includegraphics[width=\linewidth]{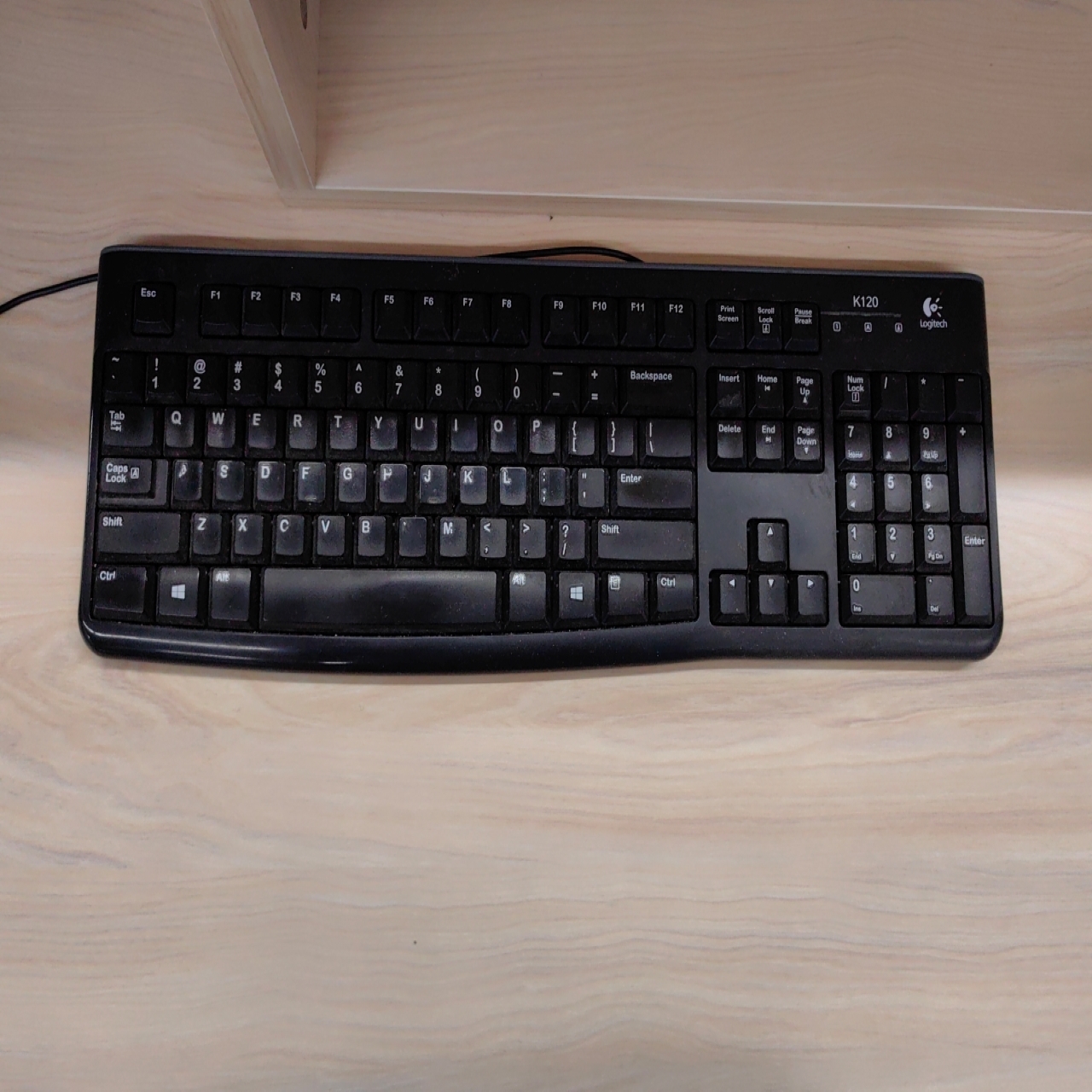}\\
        \vspace{-0.3cm}
        \includegraphics[width=\linewidth]{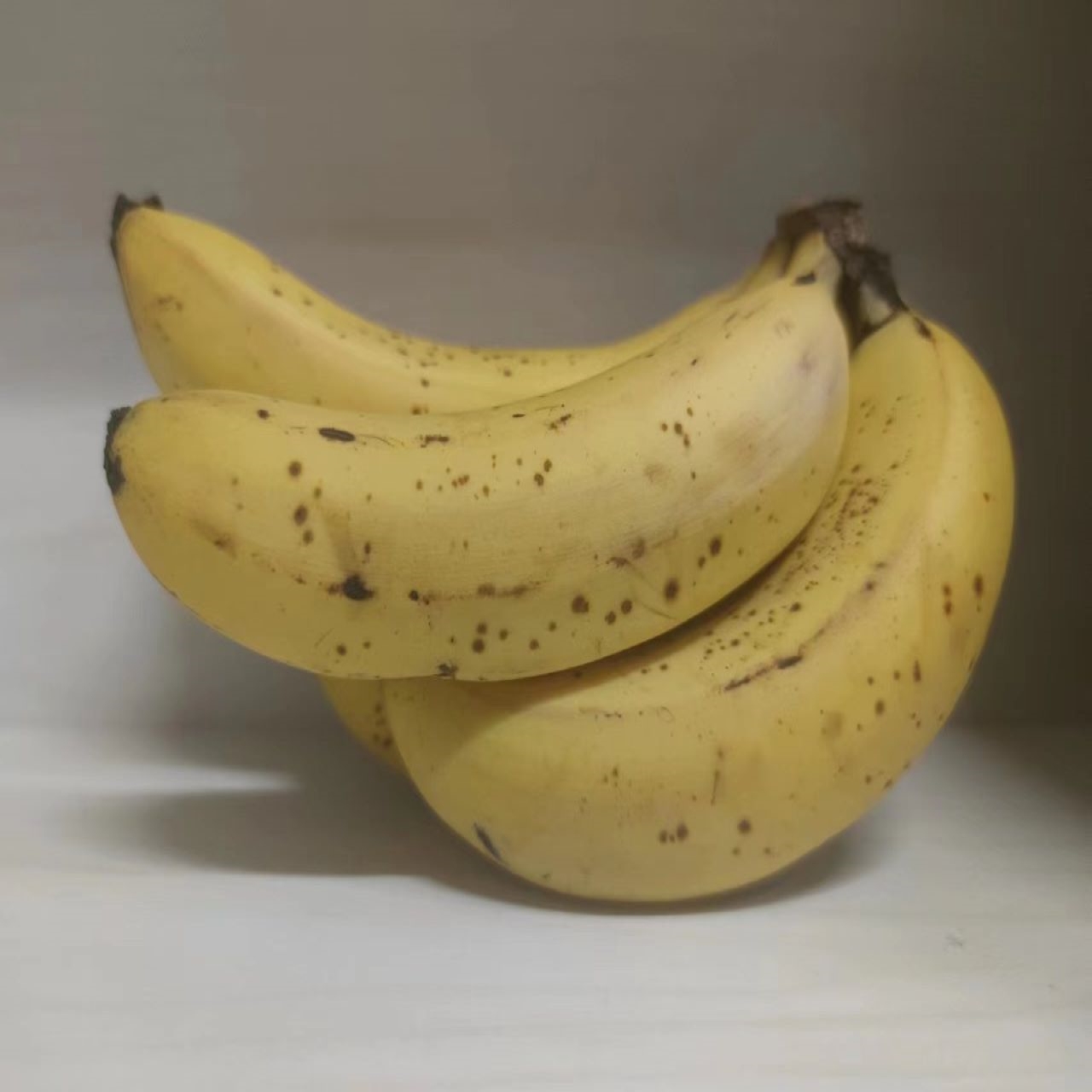}\\
       \vspace{-0.3cm}
        \includegraphics[width=\linewidth]{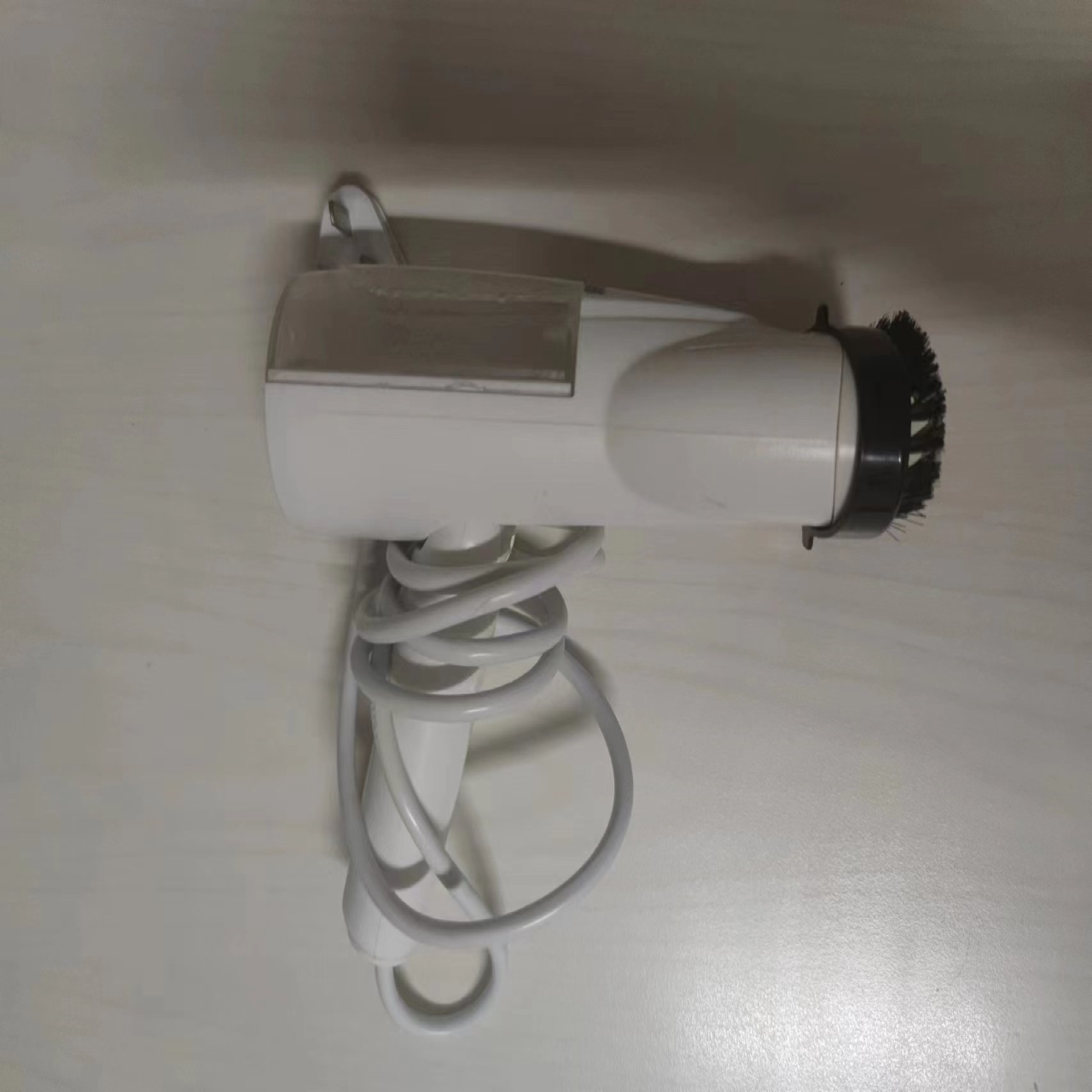}\\
       \vspace{-0.3cm}
        \includegraphics[width=\linewidth]{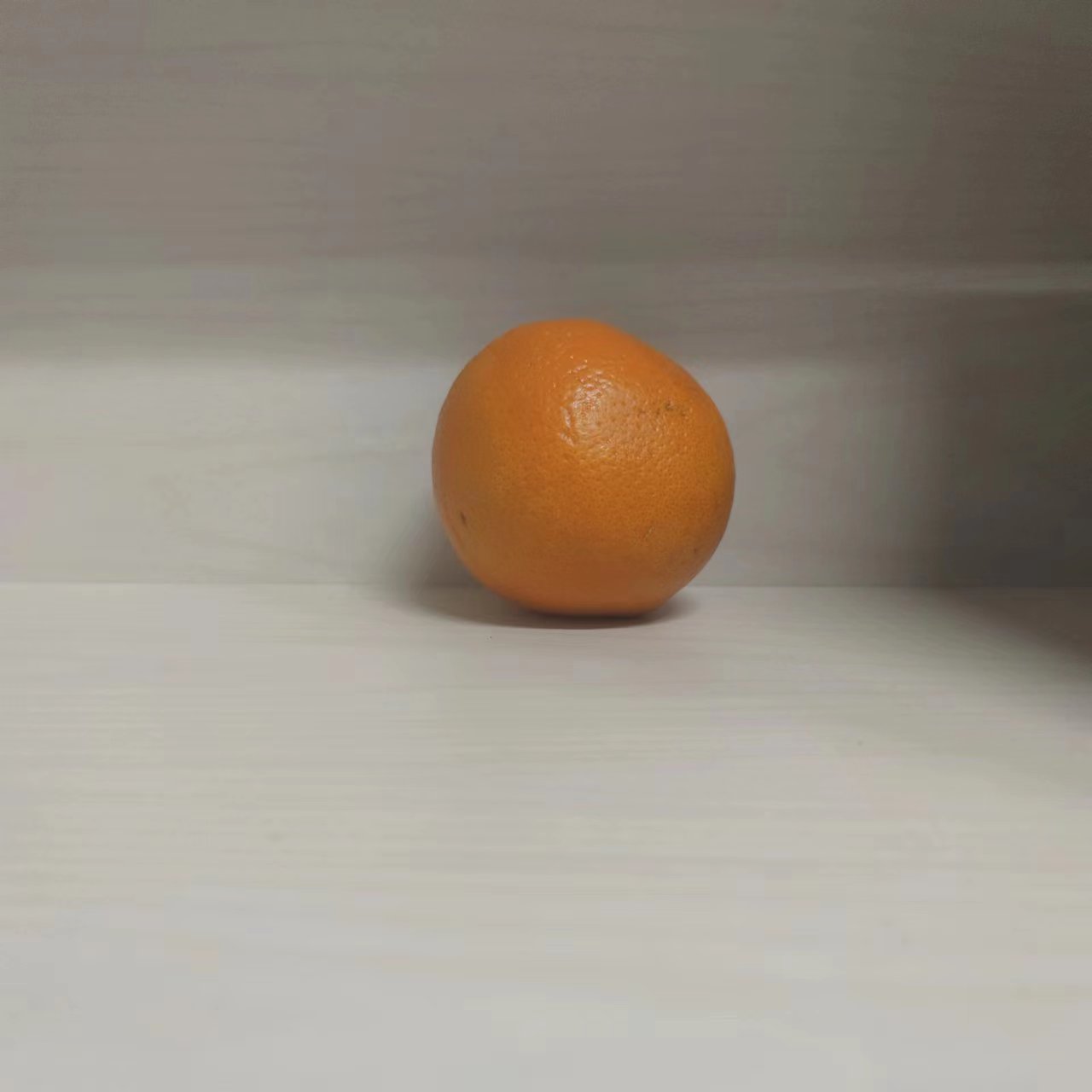}\\
       \vspace{-0.3cm}
        \includegraphics[width=\linewidth]{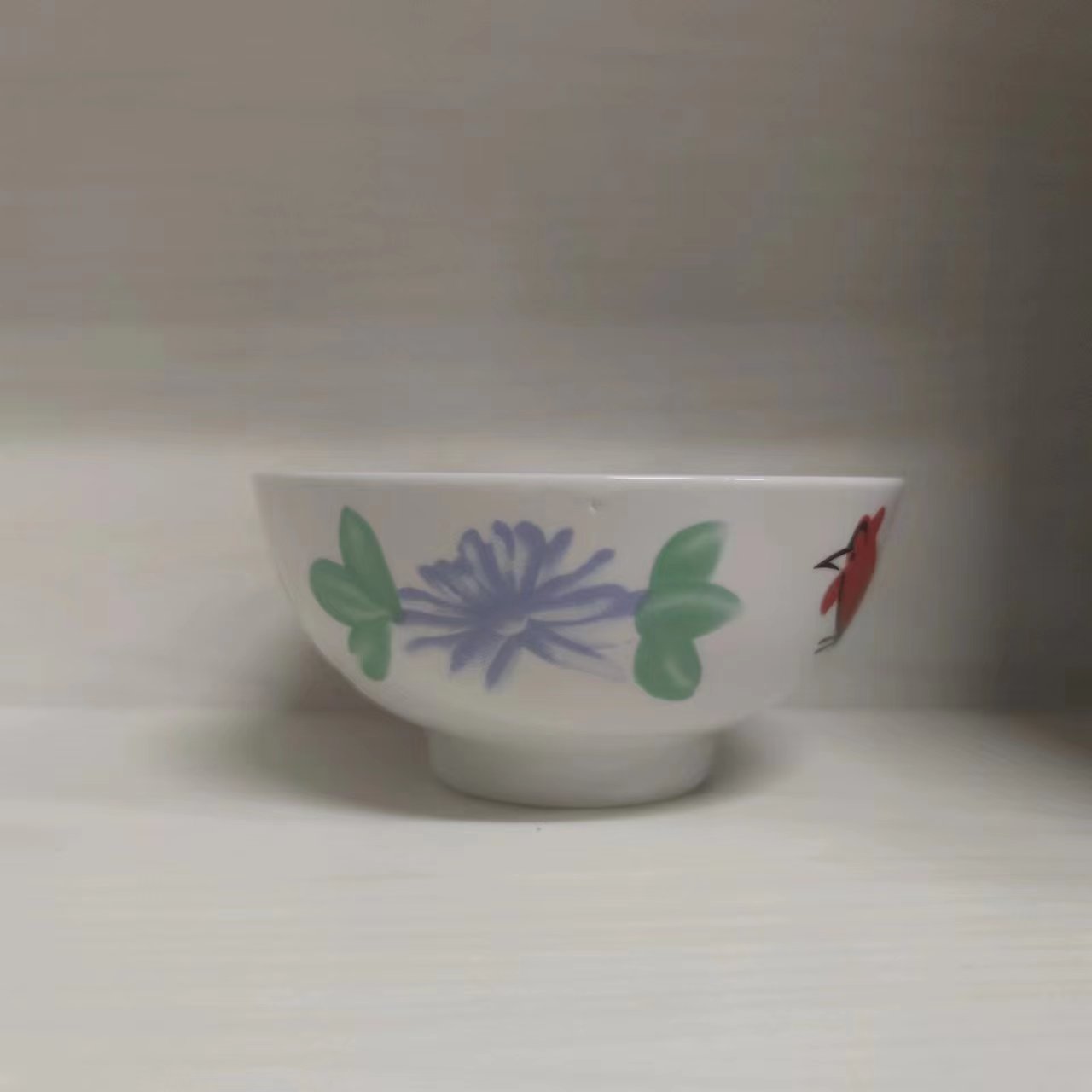}\\
       \vspace{-0.3cm}
        \includegraphics[width=\linewidth]{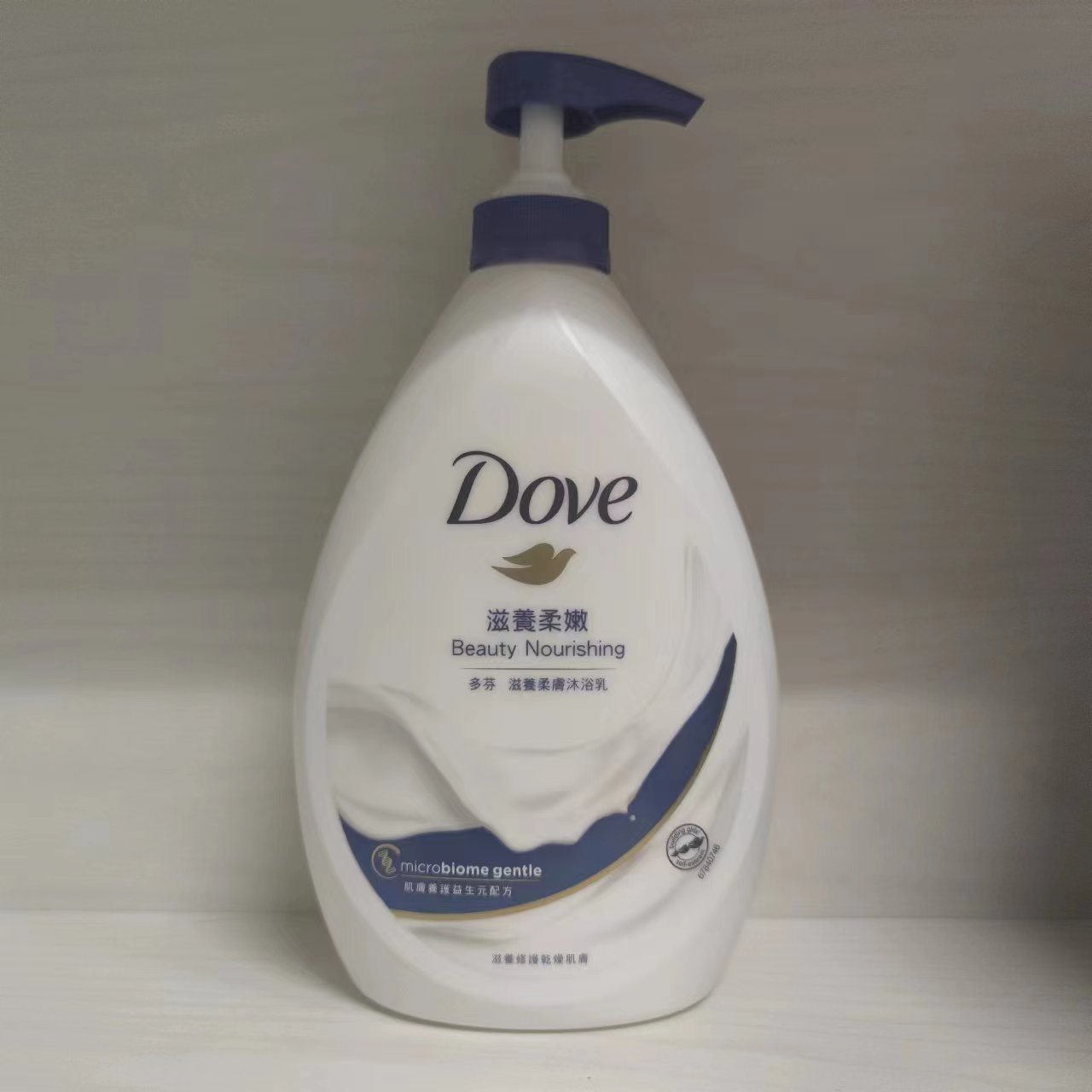}\\
       \vspace{-0.3cm}
        \caption{Raw}
        \end{subfigure}
    \end{minipage}
    \begin{minipage}[c]{0.15\textwidth}

        \begin{subfigure}{\textwidth}
        \includegraphics[width=\linewidth]{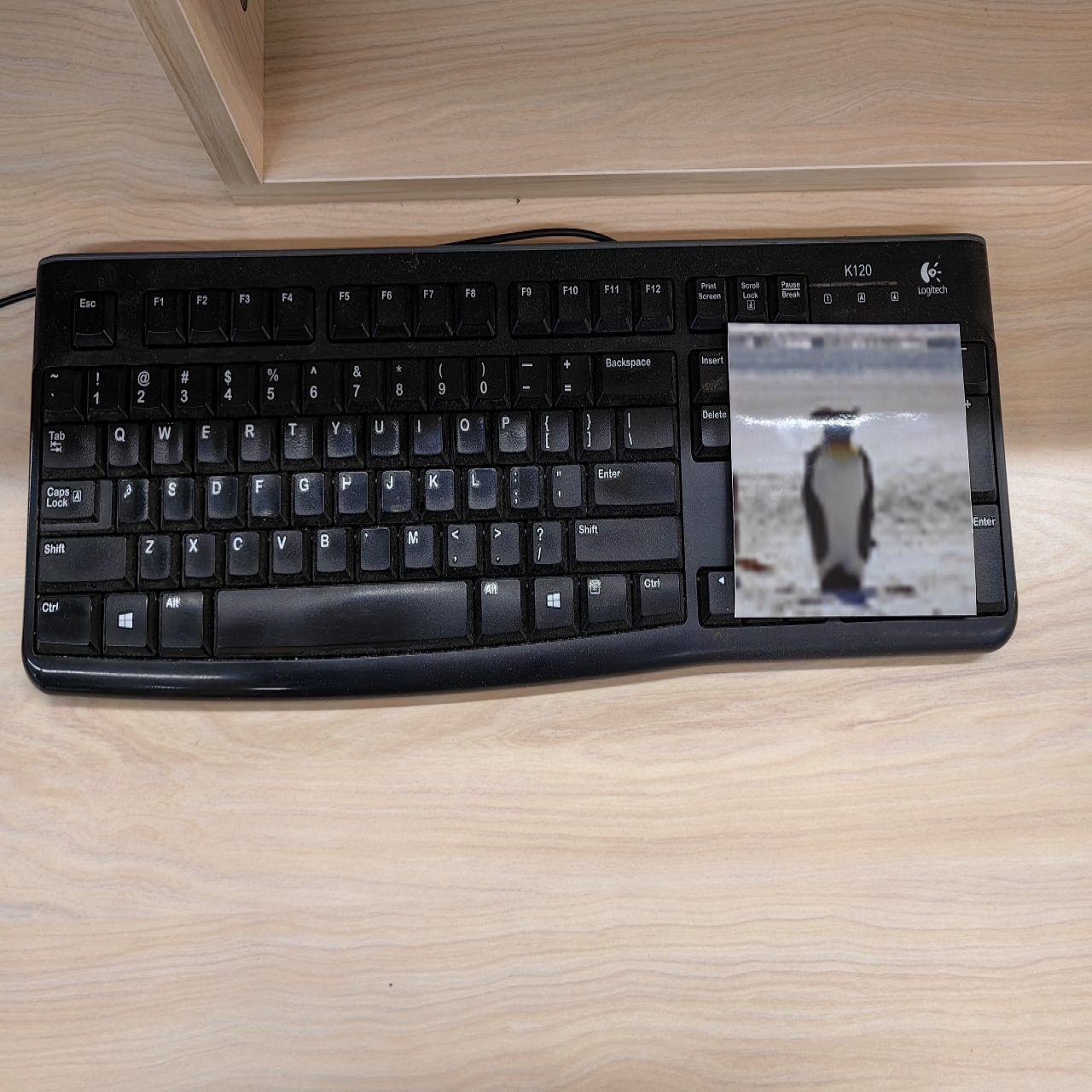}\\
        \vspace{-0.3cm}
        \includegraphics[width=\linewidth]{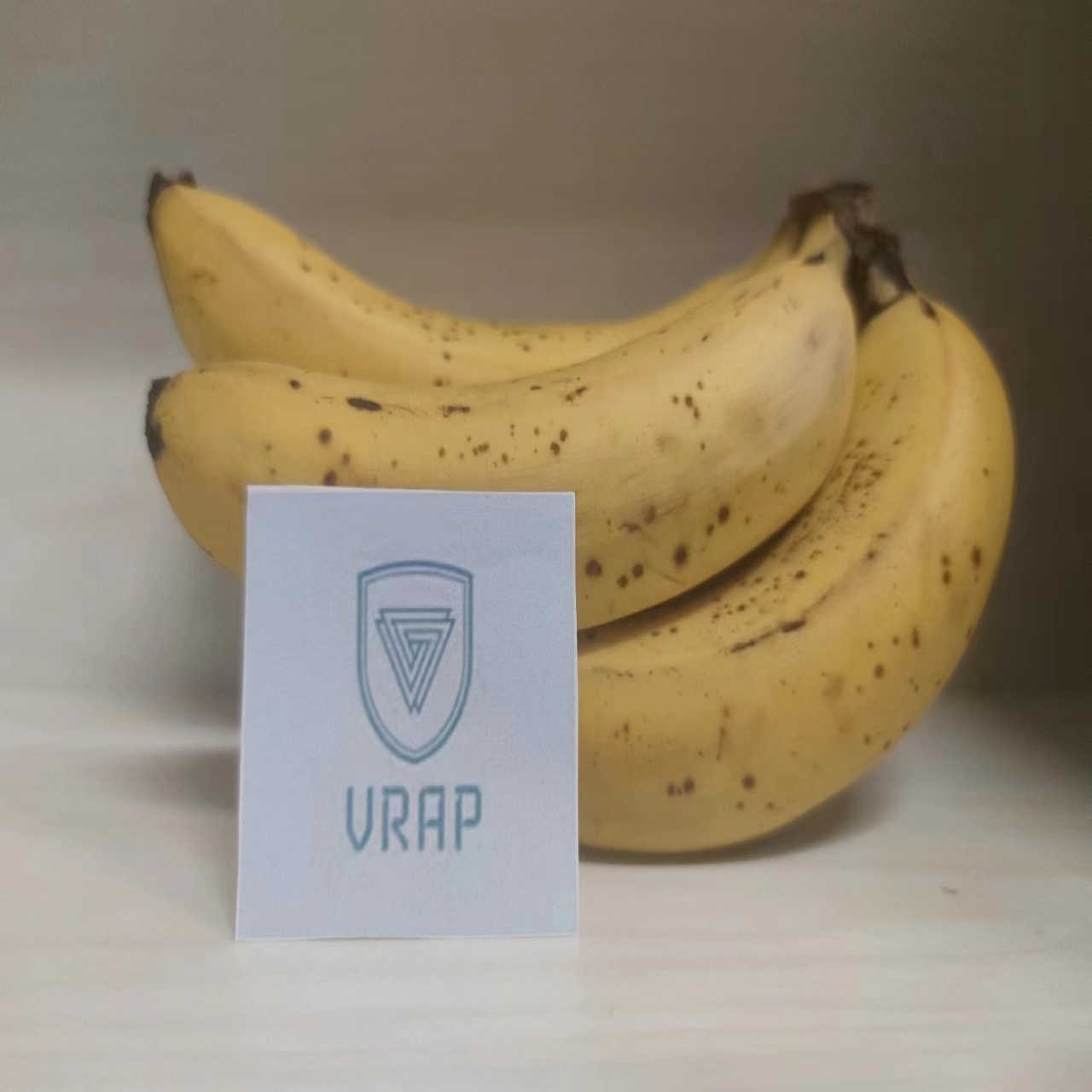}\\
        \vspace{-0.3cm}
        \includegraphics[width=\linewidth]{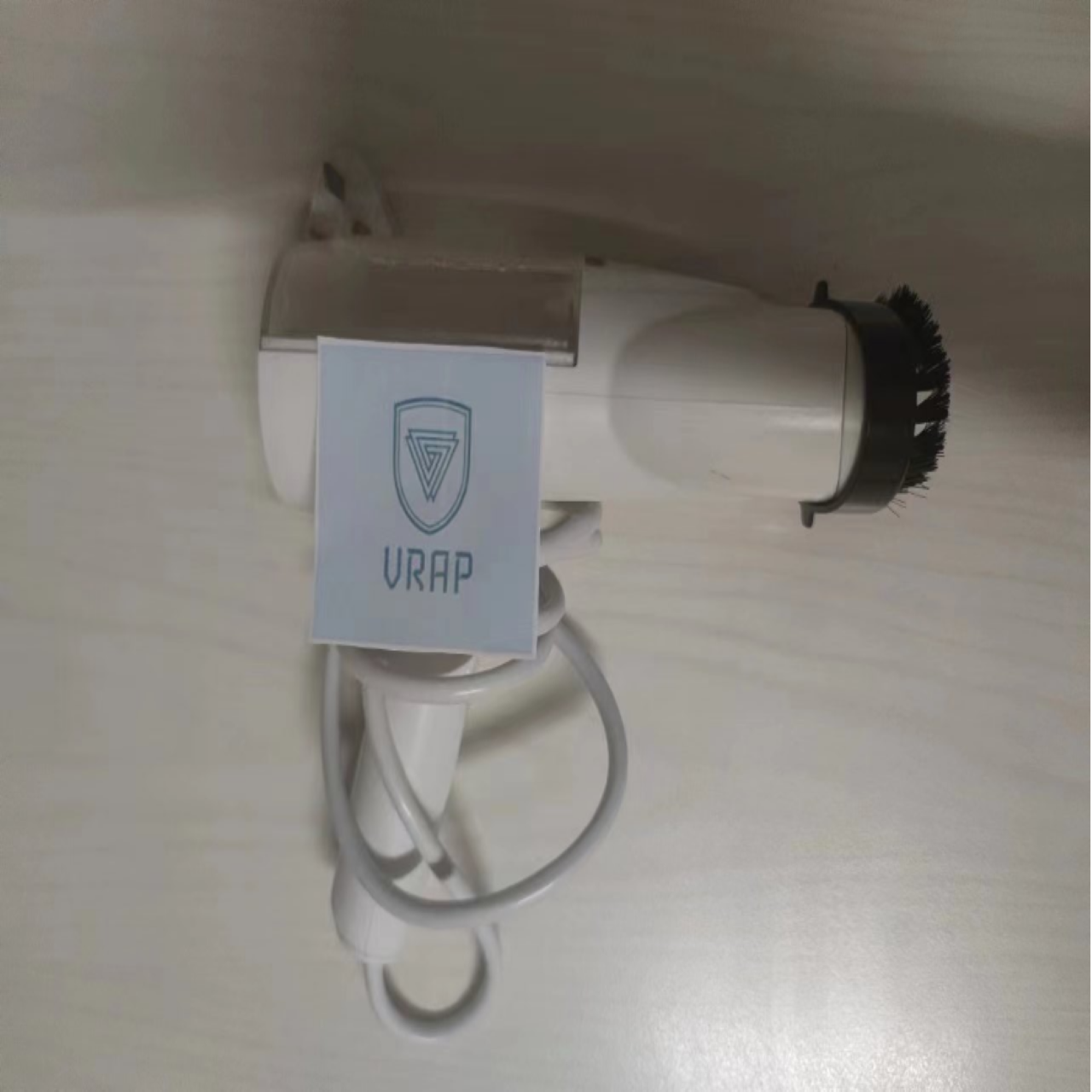}\\
        \vspace{-0.3cm}
        \includegraphics[width=\linewidth]{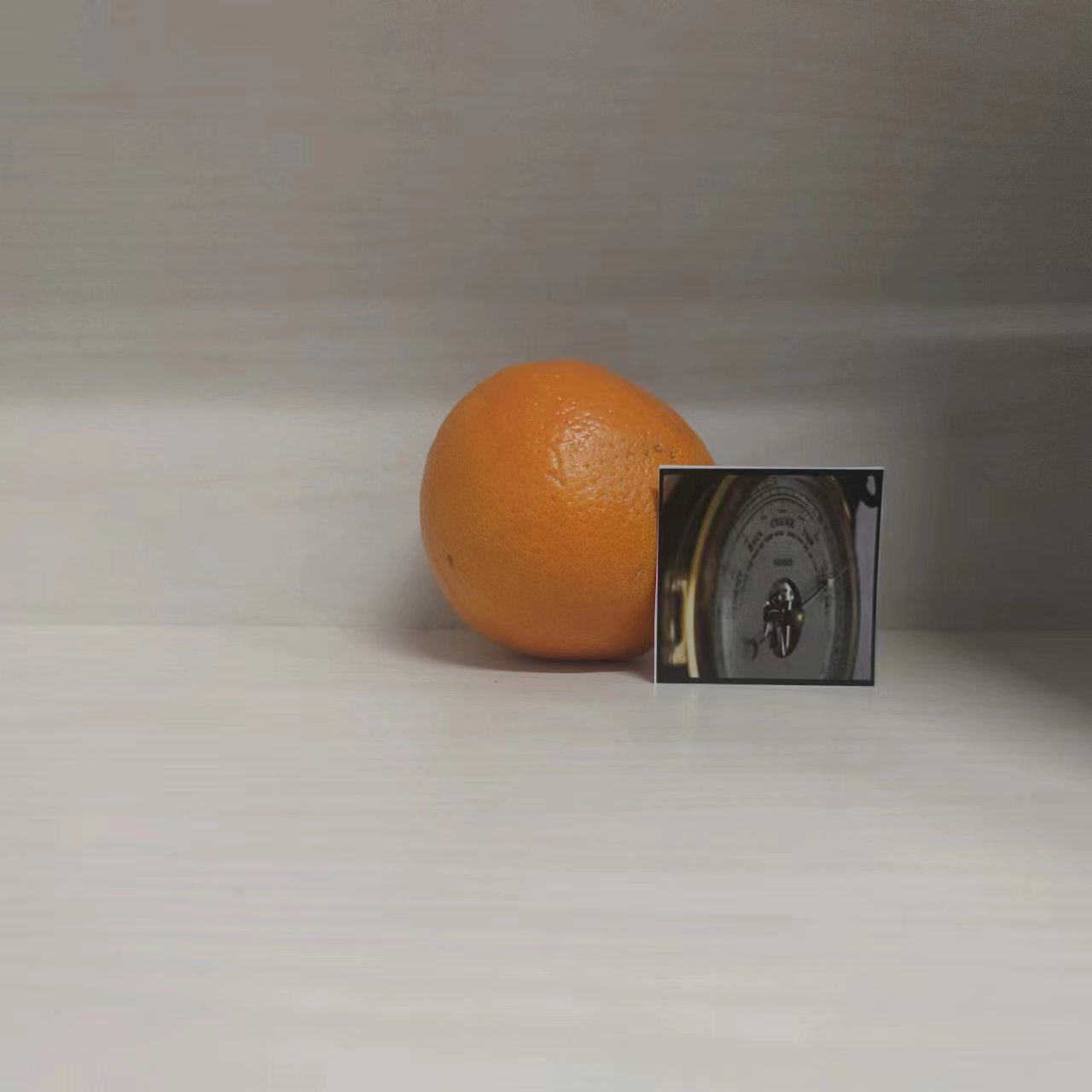}\\
        \vspace{-0.3cm}
        \includegraphics[width=\linewidth]{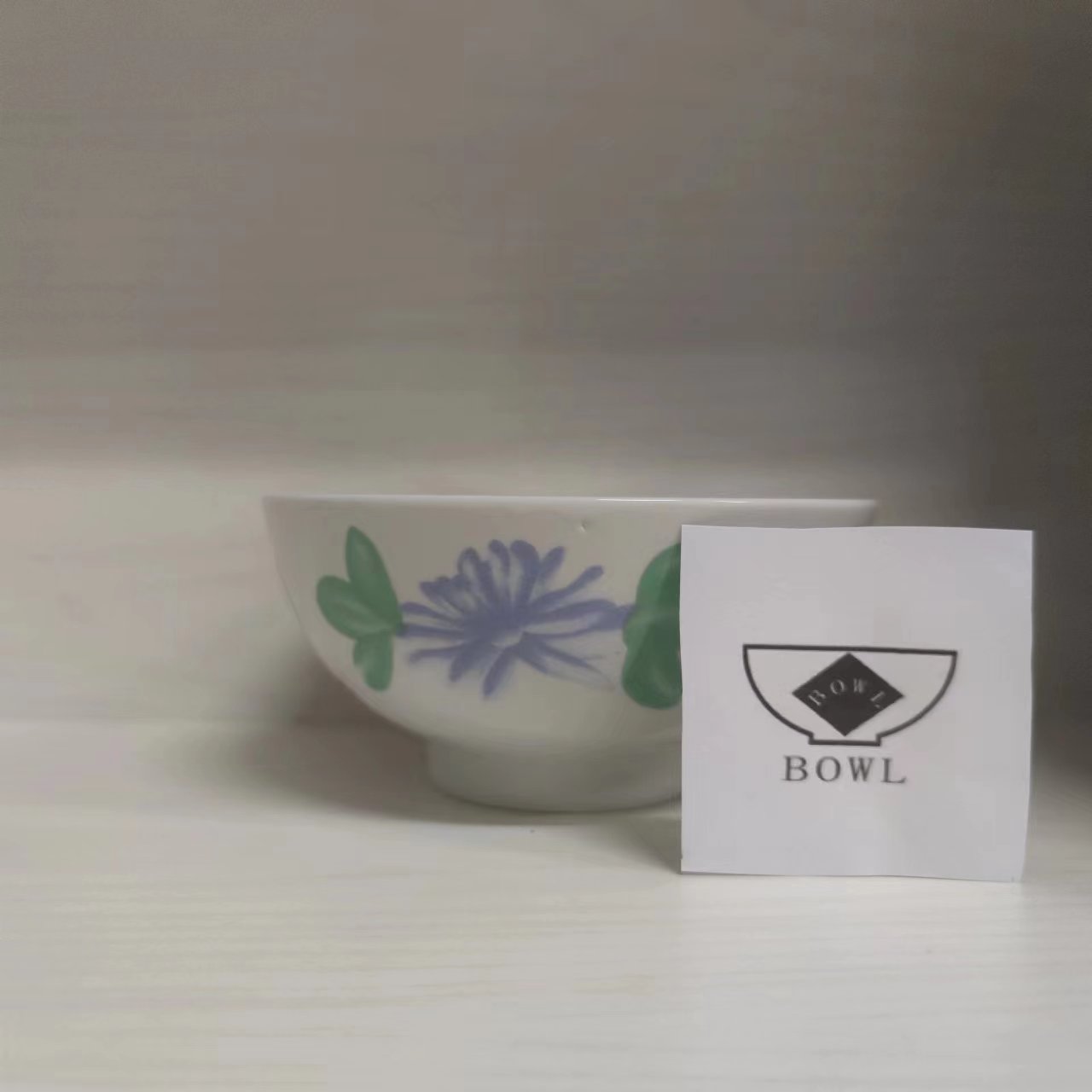}\\
        \vspace{-0.3cm}
        \includegraphics[width=\linewidth]{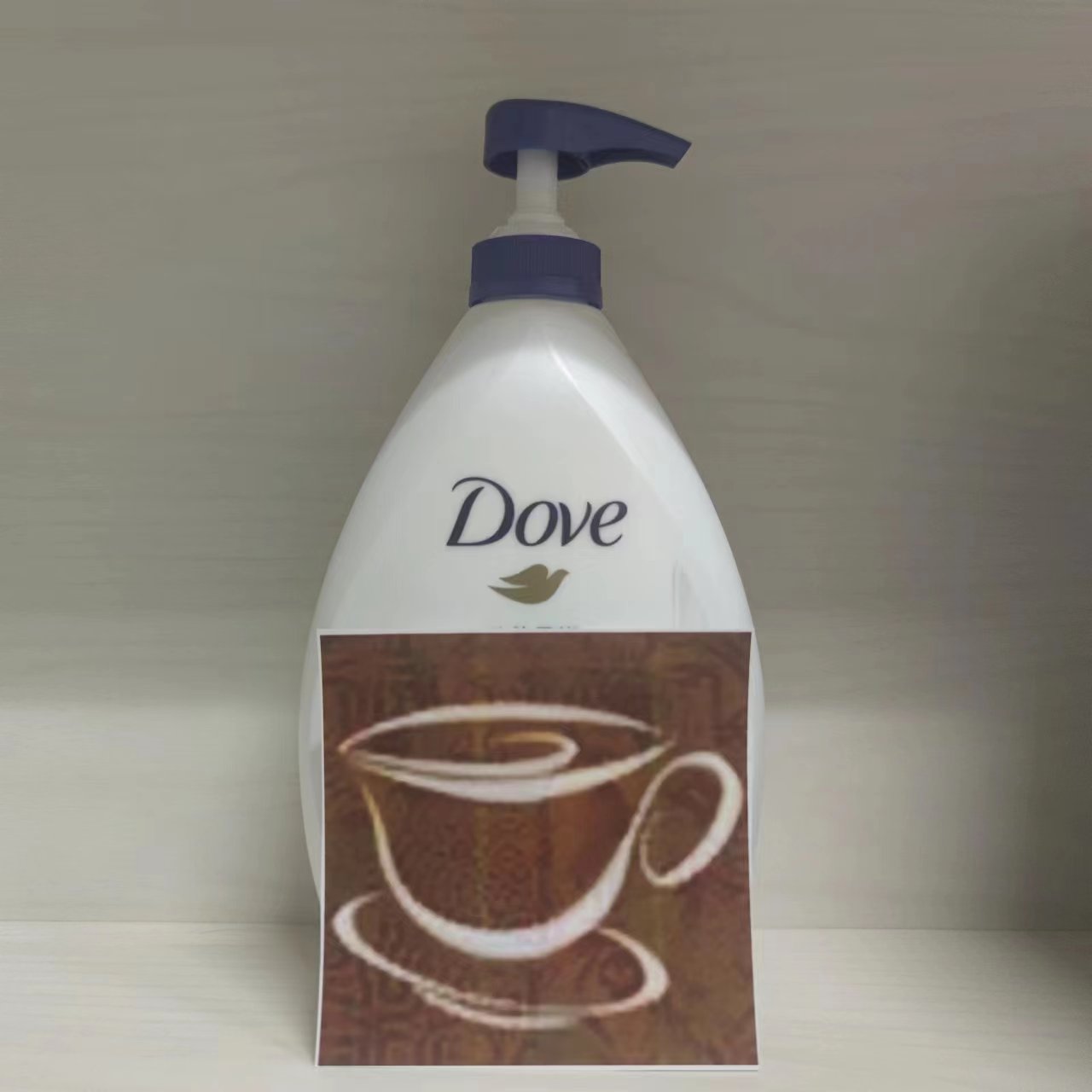}\\
        \vspace{-0.3cm}
        \caption{Original}
        %\vspace{-0.3cm}
        \label{fig:appendix_C_O}
        \end{subfigure}
    \end{minipage}%
    \hspace{0.01em}
    \begin{minipage}[c]{0.15\textwidth}

        \begin{subfigure}{\textwidth}
        \includegraphics[width=\linewidth]{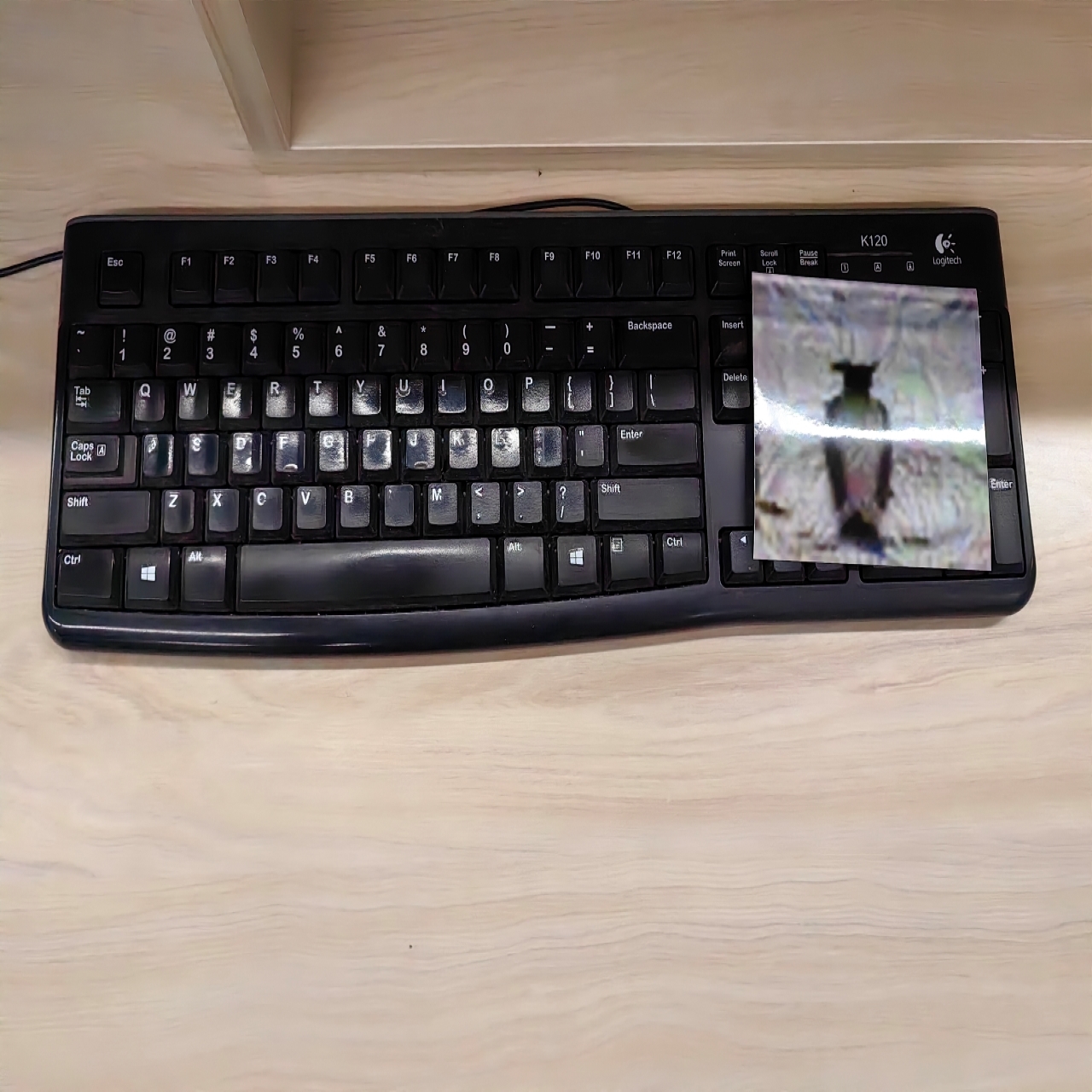}\\
        \vspace{-0.3cm}
        \includegraphics[width=\linewidth]{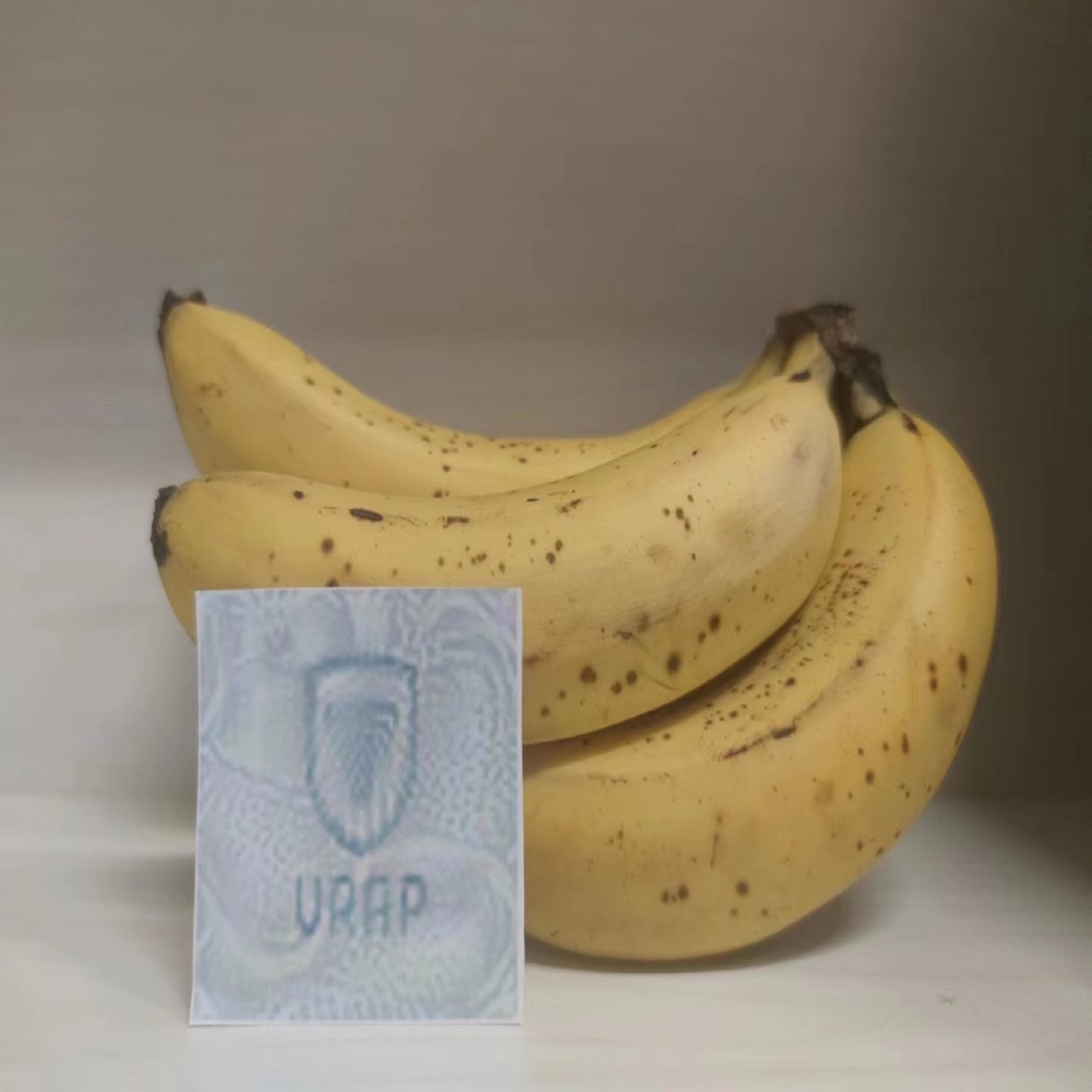}\\
        \vspace{-0.3cm}
        \includegraphics[width=\linewidth]{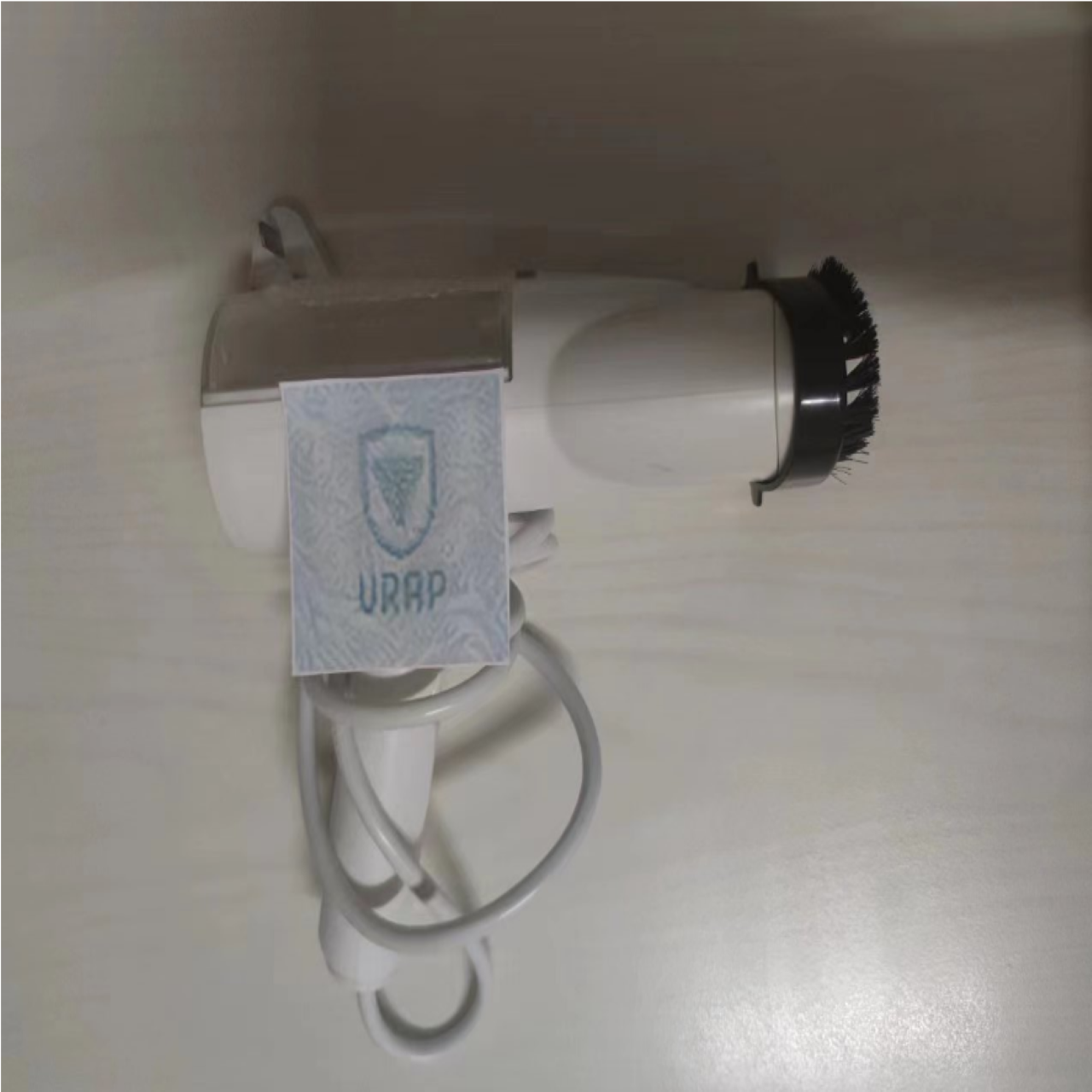}\\
        \vspace{-0.3cm}
        \includegraphics[width=\linewidth]{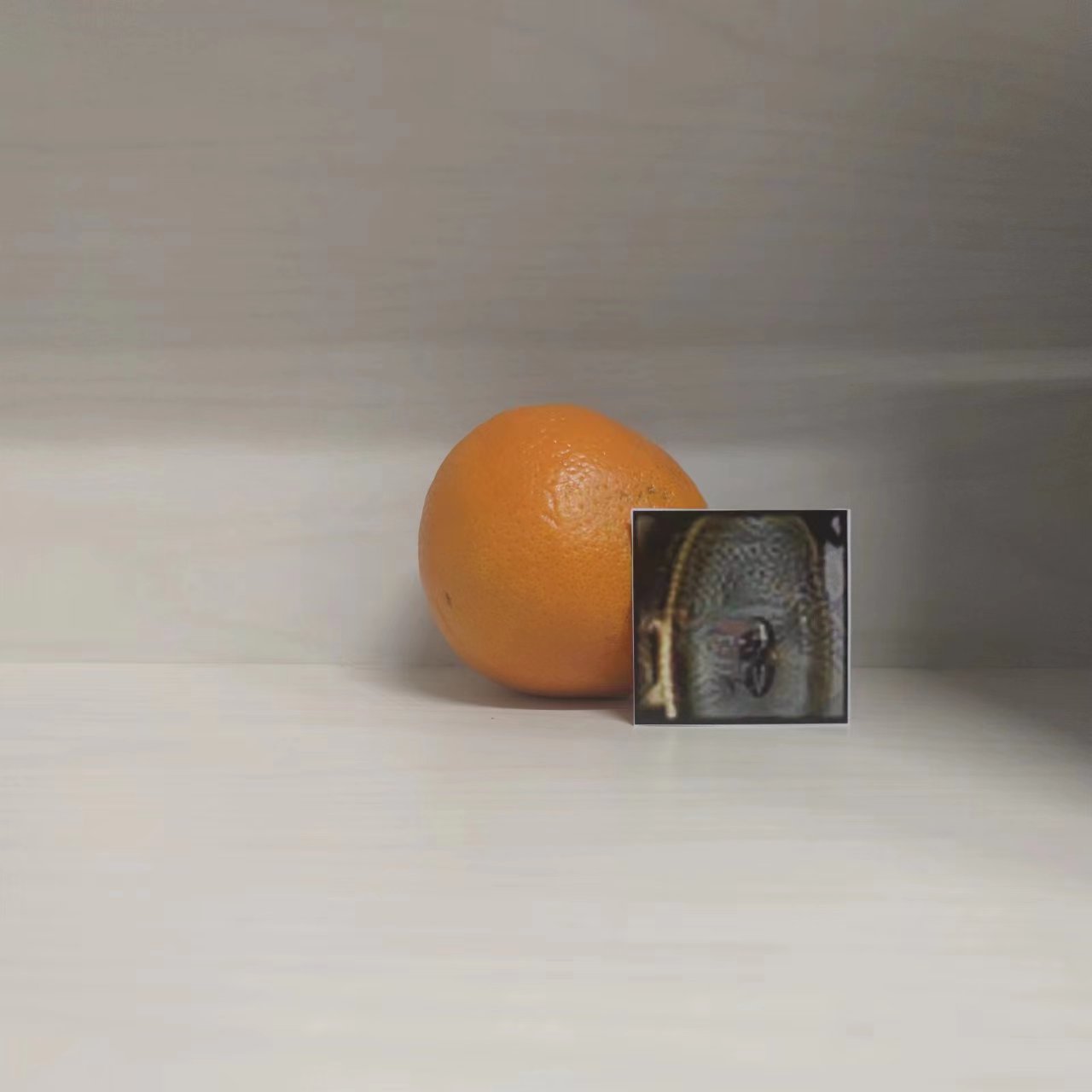}\\
        \vspace{-0.3cm}
        \includegraphics[width=\linewidth]{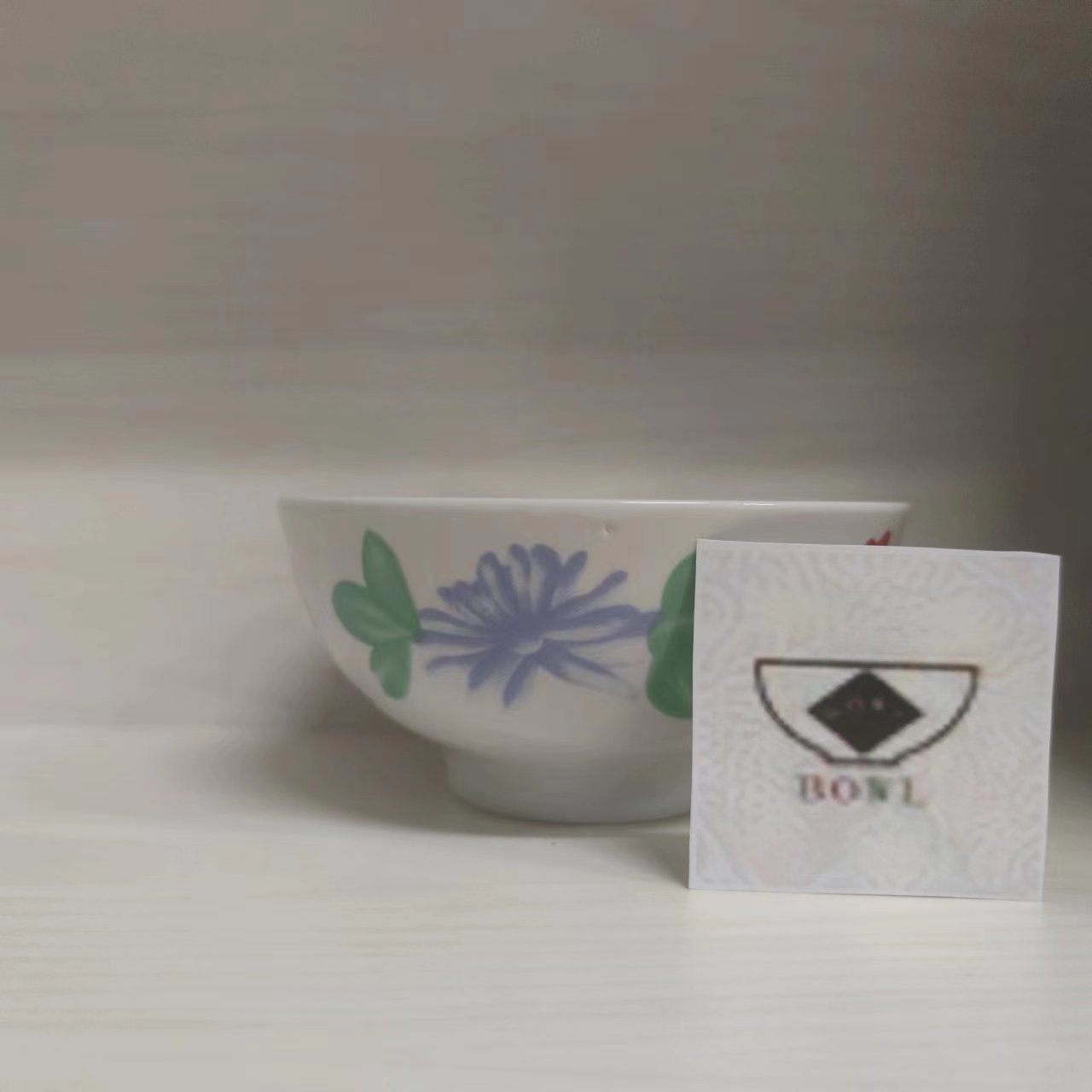}\\
        \vspace{-0.3cm}
        \includegraphics[width=\linewidth]{Appendix/figs/C/lotion2.jpg}\\
        \vspace{-0.3cm}
        \caption{Adversarial}
        \label{fig:appendix_C_A}
        \end{subfigure}
    \end{minipage}%
    % \vspace{-0.5em}
    \caption{The raw images, images with original image patches, and adversarial patches in the physical world.}
    \label{fig:Appendix_C}
    % \vspace{-0.5em}
\end{figure}
\subsection{More Samples of Physical Adversarial Patches}
\label{app:phy_patch}
In this section, we provide more adversarial patches employed in the real world. As shown in Fig~\ref{fig:Appendix_C}, the adversarial patches generated by \name remains the visual reality of the original patches. Adversarial patches in Fig~\ref{fig:appendix_C_A} successfully fool a ResNet-18 model, while the original patches in Fig~\ref{fig:appendix_C_O} pose no impact on the ResNet-18 model. This validates that our proposed \name can be employed in the real world with high visual reality.

\subsection{More Samples of Attention Heatmaps}
\label{app:atten}
\begin{figure}
    \centering
    \begin{minipage}[c]{0.15\textwidth}
        % \includegraphics[width=\linewidth]{figs/RealisticPatch/GoogleAp.png}\\
        % \vspace{-0.3cm}
        % \includegraphics[width=\linewidth]{figs/RealisticPatch/GoogleAp1.png}\\
       % \vspace{-0.3cm}
       \begin{subfigure}{\textwidth}
        \includegraphics[width=\linewidth]{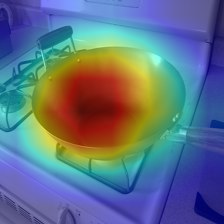}\\
        \vspace{-0.3cm}
        \includegraphics[width=\linewidth]{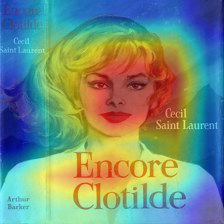}\\
       \vspace{-0.3cm}
        \includegraphics[width=\linewidth]{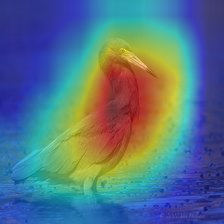}\\
       \vspace{-0.3cm}
        \includegraphics[width=\linewidth]{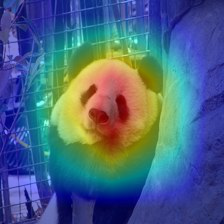}\\
       \vspace{-0.3cm}
        \includegraphics[width=\linewidth]{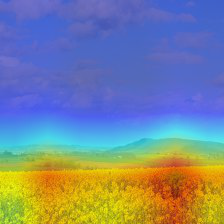}\\
       \vspace{-0.3cm}
        \includegraphics[width=\linewidth]{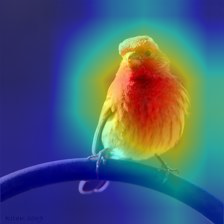}\\
       \vspace{-0.3cm}
        \caption{Raw}
        \end{subfigure}
    \end{minipage}
    \begin{minipage}[c]{0.15\textwidth}

        \begin{subfigure}{\textwidth}
        \includegraphics[width=\linewidth]{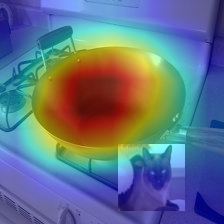}\\
        \vspace{-0.3cm}
        \includegraphics[width=\linewidth]{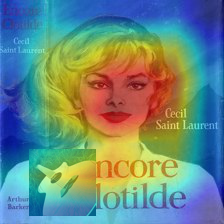}\\
        \vspace{-0.3cm}
        \includegraphics[width=\linewidth]{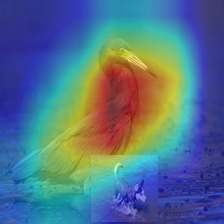}\\
        \vspace{-0.3cm}
        \includegraphics[width=\linewidth]{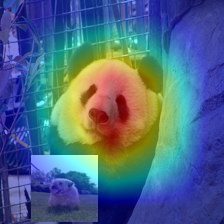}\\
        \vspace{-0.3cm}
        \includegraphics[width=\linewidth]{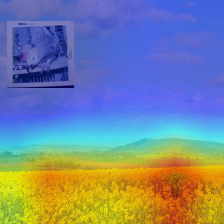}\\
        \vspace{-0.3cm}
        \includegraphics[width=\linewidth]{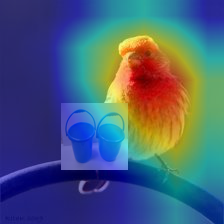}\\
        \vspace{-0.3cm}
        \caption{Original}
        %\vspace{-0.3cm}
        \end{subfigure}
    \end{minipage}%
    \hspace{0.01em}
    \begin{minipage}[c]{0.15\textwidth}

        \begin{subfigure}{\textwidth}
        \includegraphics[width=\linewidth]{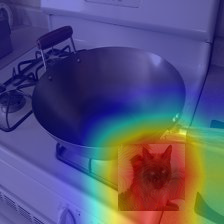}\\
        \vspace{-0.3cm}
        \includegraphics[width=\linewidth]{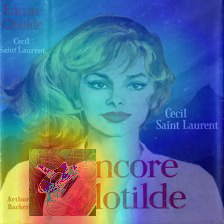}\\
        \vspace{-0.3cm}
        \includegraphics[width=\linewidth]{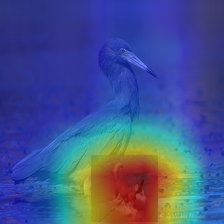}\\
        \vspace{-0.3cm}
        \includegraphics[width=\linewidth]{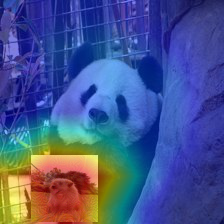}\\
        \vspace{-0.3cm}
        \includegraphics[width=\linewidth]{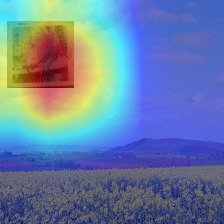}\\
        \vspace{-0.3cm}
        \includegraphics[width=\linewidth]{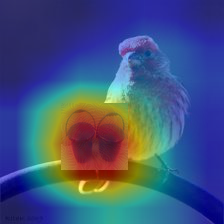}\\
        \vspace{-0.3cm}
        \caption{Adversarial}
        \end{subfigure}
    \end{minipage}%
    % \vspace{-0.5em}
    \caption{Attention heatmaps of raw image, the image with original patch and adversarial patches on ResNet-18.}
    \label{fig:Appendix_D}
    % \vspace{-0.5em}
\end{figure}

In this section, we provide more heatmaps after adding original patches and adversarial patches. As shown in Fig~\ref{fig:Appendix_D}, after adding adversarial patches, ResNet-18 is distracted by the adversarial patches, which is consistent with the results in Discussion.

\end{document}